\RequirePackage{fix-cm}
\PassOptionsToPackage{hypertexnames=false}{hyperref}
\documentclass[11pt, a4paper, twocolumn, copyright, google]{google}
\usepackage{array}
\usepackage{algorithm}
\usepackage{algpseudocode}
\usepackage{placeins}
\usepackage{tcolorbox}
\tcbuselibrary{skins,breakable}

\definecolor{LightBlue}{HTML}{E6F2EE}
\definecolor{GuardForest}{HTML}{1B5E4A}
\definecolor{GuardPine}{HTML}{216B5D}
\newcommand{\yesmark}{\ding{51}}
\newcommand{\nomark}{\ding{55}}
\newcommand{\partialmark}{{\color{gray}\ding{51}}}

\usepackage{hyperref}
\hypersetup{
    colorlinks=true,
    linkcolor=GuardForest,
    citecolor=GuardForest,
    urlcolor=GuardForest
}
\usepackage[scaled=0.92]{helvet}
\AtBeginDocument{{\sffamily}%
  \DeclareFontShape{T1}{phv}{m}{sc}{<->ssub*XCharter-TLF/m/sc}{}%
  \DeclareFontShape{T1}{phv}{bx}{sc}{<->ssub*XCharter-TLF/b/sc}{}%
  \DeclareFontShape{T1}{phv}{b}{sc}{<->ssub*XCharter-TLF/b/sc}{}}

\newtheorem{definition}{Definition}
\newtheorem{theorem}{Theorem}
\newtheorem{proposition}{Proposition}

\definecolor{gateborder}{HTML}{216B5D}
\definecolor{slateborder}{RGB}{71, 85, 105}
\definecolor{slatebg}{HTML}{FAF8F3}

\tcbset{
    promptbox/.style={
        enhanced,
        colback=slatebg,
        colframe=slateborder!55,
        arc=3pt,
        boxrule=0.7pt,
        left=6pt, right=6pt, top=5pt, bottom=5pt,
        fontupper=\small\ttfamily
    },
    resultbox/.style={
        enhanced,
        breakable,
        colback=slatebg,
        frame hidden,
        boxrule=0pt,
        borderline west={2pt}{0pt}{gateborder!85},
        sharp corners,
        left=6pt, right=5pt, top=2.5pt, bottom=2.5pt,
        before skip=4pt, after skip=4pt,
        before upper={\setlength{\abovedisplayskip}{3pt}\setlength{\belowdisplayskip}{3pt}\setlength{\abovedisplayshortskip}{1pt}\setlength{\belowdisplayshortskip}{2pt}}
    },
    questionbox/.style={
        enhanced,
        colback=LightBlue!60!white,
        frame hidden,
        boxrule=0pt,
        borderline west={2pt}{0pt}{GuardForest},
        arc=2pt,
        left=7pt, right=7pt, top=5.5pt, bottom=5.5pt,
        before skip=8pt, after skip=11pt
    }
}

\newtcolorbox[auto counter, number within=section]{promptlisting}[2][]{%
    promptbox, breakable,
    fontupper=\small\sffamily,
    fonttitle=\small\bfseries,
    title={Box~\thetcbcounter: #2},
    #1}

\IfFileExists{latexml.sty}{\usepackage{latexml}}{\newif\iflatexml\latexmlfalse}
\makeatletter
\iflatexml
  \usepackage{framed}
  \definecolor{shadecolor}{HTML}{FAF8F3}
  \theoremstyle{definition}
  \newtheorem{ltxpromptbox}{Box}[section]
  \define@key{guardbox}{label}{\def\guardbox@label{#1}}
  \renewenvironment{tcolorbox}[1][]{%
    \par\medskip\begin{shaded}\noindent\ignorespaces
  }{%
    \end{shaded}\par\medskip
  }
  \renewenvironment{promptlisting}[2][]{%
    \def\guardbox@label{}%
    \setkeys{guardbox}{#1}%
    \par\medskip\begin{framed}\begin{ltxpromptbox}[#2]%
    \ifx\guardbox@label\@empty\else\label{\guardbox@label}\fi
    \par\smallskip\small\ignorespaces
  }{%
    \end{ltxpromptbox}\end{framed}\par\medskip
  }
\fi
\makeatother

\newif\ifanon
\anonfalse

\title{GUARD: Stabilizing Recursive Self-Improvement with Regret Guidance}

\ifanon
  \author[1]{Anonymous Authors}
  \affil[1]{Paper under double-blind review}
\else
  \uselogo{}
  \author[1,*]{Minghe Zhang}
  \author[1,*]{Xuetao Kang}
  \author[1]{Yifei Ming}
  \author[1]{Yanfei Chen}
  \author[1]{Han Yu}
  \author[1]{Yufan Zhuang}
  \author[1]{Chen-Yu Lee}
  \affil[1]{Google Cloud}
  \affil[*]{Equal contribution}
  \correspondingauthor{minghezhang@google.com, xuetaok@google.com}
\fi

\begin{document}

\begin{abstract}
Autonomous AI agents pair a frozen base model with an editable \emph{harness} of prompts, skills, tool rules, and memory. Recursive self-improvement (RSI) adapts this harness from experience without weight updates, yet self-editing without external graders often backfires: reflection and scalar scores conflate \emph{what an edit newly broke} with \emph{what remains unsolved}, causing local regressions, skill forgetting, and memory bloat. We introduce \textbf{GUARD} (\emph{Grader-free Unsupervised Environment Design [UED] with Adaptive Rule Distillation}), a closed-loop framework for reference-free continual harness evolution. Our key insight is to decompose each task into pre-execution self-checks and guide harness updates via \emph{directional regret}. Each round grades trajectories against these checks, pinning mastered rules while prioritizing newly broken behaviors, and selects among parallel candidate edits by critic-free group-relative fitness. A \emph{Soft-HURT} gate then promotes an edit only when gains outweigh regressions, scoping specialized fixes to prevent over-triggering and guaranteeing non-regression on its gated buffer. At test time, one frozen harness runs in a single pass. Across benchmarks on memory, skill evolution, continual task streams, and domain reasoning, \texttt{GUARD} consistently improves performance without catastrophic forgetting or memory pollution while substantially cutting deployment token usage.
\end{abstract}

\maketitle

\section{Introduction}

Modern language model agents combine a frozen base model with a surrounding \emph{harness} of prompts, reusable skills, tool interfaces, memory, and control logic \cite{packer2023memgpt,wang2023voyager,lee2026metaharness,xia2026rrsi}. In practice, this harness decides whether the same model checks the right state before acting, calls tools with valid arguments, recovers from failed commands, and carries useful rules across sessions \cite{yang2024sweagent,tang2026wikiskill}. Much recent progress in deployed agents comes from \emph{harness engineering}, yet tuning harnesses by hand scales poorly: engineers can inspect only a small slice of failed runs, and fixed designs quickly go stale as tools, schemas, and user workflows change \cite{chen2026continual,xu2026harness2}. \emph{Recursive self-improvement} (RSI) aims to automate this loop by turning past experience into persistent edits that improve future runs \cite{zelikman2022star,shinn2023reflexion,wang2024survey,meng2026rsigym}. Because frontier models often sit behind black-box APIs with frozen weights and open-world tasks rarely provide ground-truth rewards for weight-space reinforcement learning \cite{zhao2025absolute,zhou2025selfchallenging,huang2026rzero,xia2026agent0}, treating the editable harness as a non-parametric policy around the frozen model offers a practical, inspectable way to improve agents after deployment \cite{madaan2024self,shao2026clm}.

\begin{figure}[!t]
\centering
\includegraphics[width=\columnwidth]{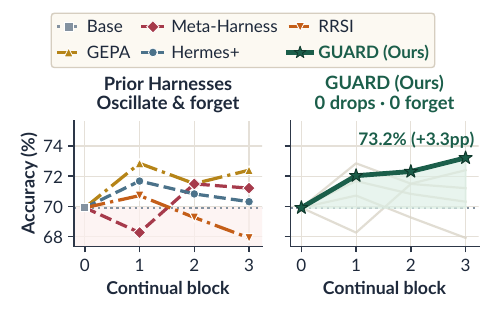}
\vspace{-6pt}
\caption{\textbf{Continual harness evolution across shifting tasks.} \emph{Left:} On \textsc{Drift Gauntlet} \cite{chen2026continual} (four continual task streams under single-model \texttt{Gemini 4}), prior persistent harnesses oscillate across adaptation blocks and drop below the base agent as new edits overwrite earlier skills. \emph{Right:} \texttt{GUARD} improves monotonically at every checkpoint with zero block drops and zero post-block forgetting.}
\label{fig:intro_teaser}
\vspace{-10pt}
\end{figure}

\begin{figure*}[t]
\centering
\includegraphics[width=\textwidth]{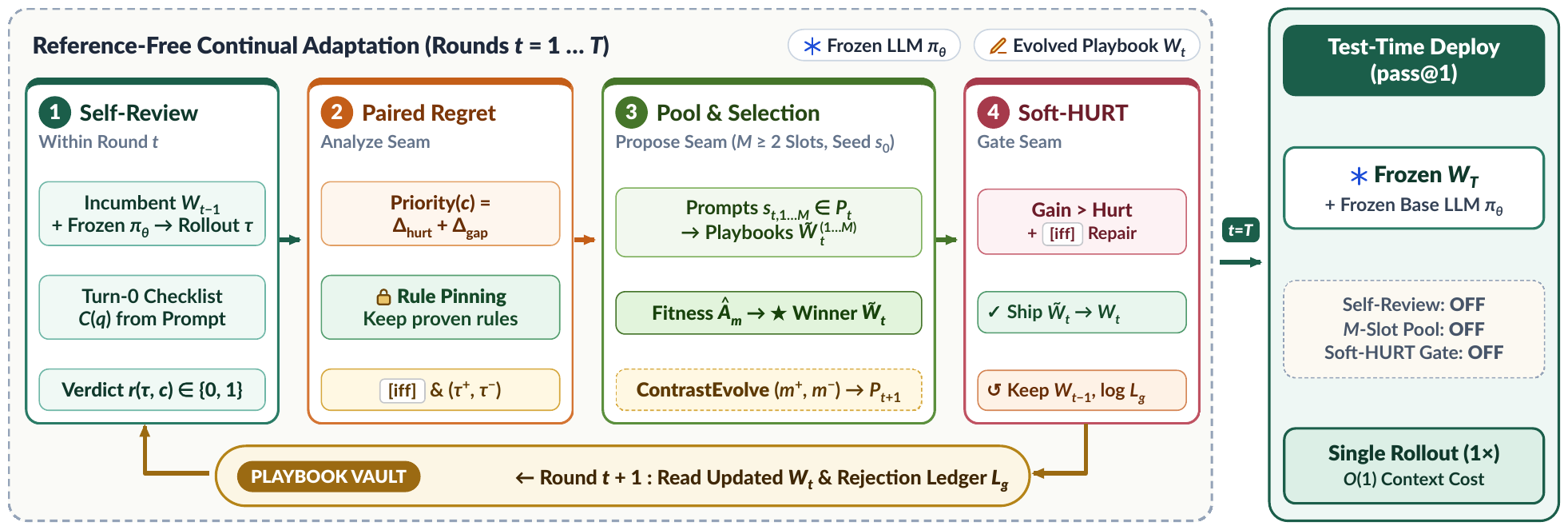}
\vspace{-6pt}
\caption{\textbf{\texttt{GUARD}: regret-guided continual recursive self-improvement of agent harnesses} (\emph{Grader-free UED with Adaptive Rule Distillation}). \textbf{Adaptation phase (\S\ref{sec:methodology}):} across rounds, the frozen model executes tasks with the current harness while a reference-free self-reviewer checks simple task requirements on each trajectory; \texttt{Analyze} separates newly broken behaviors from remaining gaps and pins working rules; \texttt{Propose} compares passing and failing runs to write targeted rules that state when they apply; and \texttt{Gate} (\emph{Soft-HURT}) keeps an update only when gains outweigh regressions while logging rejected edits so bad changes are not repeated. \textbf{Test-time evaluation:} candidate pooling, self-review, and gating are turned off, deploying the single frozen harness for standard single-run evaluation (\texttt{pass@1} on all primary benchmarks, matching the base agent's rollout budget).}
\label{fig:unified_architecture}
\vspace{-6pt}
\end{figure*}

Early automated methods search over prompts or agent code offline using labeled training splits before deployment \cite{khattab2023dspy,hu2025adas,zhang2025aflow,zhang2026dgm,lee2026metaharness}. The resulting static harnesses struggle when test-time tasks shift. Newer approaches regularize offline search for out-of-distribution generalization \cite{xia2026rrsi}, probe per-task harnesses online while saving reusable lessons \cite{xu2026harness2}, or append reflections and skills across episodes \cite{zhao2024expel,ouyang2026reasoningbank,xue2026pastbench,agrawal2026gepa}. Yet continually updating a single persistent harness across rounds without reference answers or external graders often degrades performance over time \cite{chen2026continual}. Because every edit changes subsequent rollouts, free-form reflection and scalar task scores mix two distinct failure signals: behaviors the latest edit \emph{newly broke} and capability gaps the agent \emph{still has not solved}. Mixing these signals across rounds causes three failure modes: \emph{local overfitting} (\emph{over-triggering}), where a broad rule written for recent failures breaks tasks that already worked; \emph{catastrophic forgetting}, where shifts in workflow, topic, or domain overwrite earlier skills \cite{chen2026continual}; and \emph{memory pollution}, where unverified additions accumulate stale or conflicting rules that bloat the prompt. Over multiple rounds, self-editing can leave an agent worse than its starting point (Figure~\ref{fig:intro_teaser}).

\begin{tcolorbox}[questionbox]
\itshape\textbf{Central Question.} How can an autonomous agent continually improve its persistent harness in a reference-free setting without interfering with or catastrophically forgetting previously learned skills?
\end{tcolorbox}

We address this challenge by casting harness adaptation around a frozen model as gradient-free policy optimization in text space, guided by two ideas from reinforcement learning: curriculum regret from unsupervised environment design (UED) \cite{dennis2020emergent,jiang2021prioritized,jiang2021replay,rutherford2024no} and critic-free group-relative scoring from Group Relative Policy Optimization (GRPO) \cite{shao2024deepseekmath,cai2025trainingfree}. We never update the model weights or train a value critic: the harness is the only part of the policy that changes (\S\ref{sec:generalized_rsi}). We implement this approach in \textbf{GUARD} (\emph{Grader-free UED with Adaptive Rule Distillation}; Figure~\ref{fig:unified_architecture}, \S\ref{sec:methodology}). Before executing a task, \texttt{GUARD} compiles the task prompt and tool schema into a fixed checklist of atomic binary checks (\S\ref{sec:self_review}) and grades completed runs with a two-layer (sandbox execution plus semantic trace) self-reviewer, because checking one requirement on a finished trace is simpler than satisfying every constraint during generation. Across rounds, \texttt{Analyze} (\S\ref{sec:paired_regret}) pins criteria that stay satisfied and ranks newly broken criteria ahead of unsolved gaps; \texttt{Propose} (\S\ref{sec:grpo_pool}) samples a parallel group of candidate edits, scores them against each other on the same tasks, and contrasts the best and worst candidates to rewrite its proposal prompts; and \texttt{Gate} (\S\ref{sec:rrsi_gate}) promotes the winner only when its gains exceed induced regressions (\emph{Soft-HURT}), narrowing rules that cause isolated drops with \texttt{[iff]} preconditions and logging rejected edits. At test time, candidate pooling, self-review, and gating are turned off, and the agent runs a single pass (\texttt{pass@1}) with its frozen harness.

Three choices separate \texttt{GUARD} from prior harness learners. Reflexion and \textsc{GEPA} update from free-form reflection or scalar and Pareto scores, so they cannot tell a new regression from an old gap; \texttt{GUARD} ranks checklist criteria by paired directional regret. \textsc{Meta-Harness} searches agent code against raw trace archives, and \textsc{RRSI} rewrites one candidate per round behind a score-band gate that leaves per-task drops unbounded; \texttt{GUARD} scores a parallel group of candidates on the same tasks and promotes the winner only when its gains exceed induced regressions, narrowing rules with \texttt{[iff]} preconditions instead of deleting them. We evaluate \texttt{GUARD} across four benchmark settings (\S\ref{sec:experiments}--\S\ref{sec:results}, Appendix~\ref{app:additional}): \textsc{PAST-Bench} \cite{xue2026pastbench}, \textsc{EvoHarnessBench} \cite{evoharnessbench2026}, \textsc{Drift Gauntlet} \cite{chen2026continual}, and professional legal and financial reasoning suites \cite{akyurek2025prbench,jiang2026finrate,moslonka2026ledger}. Our contributions are threefold:
\begin{itemize}[leftmargin=*,noitemsep,topsep=2pt]
    \item \textbf{Trace-grounded checklist self-review and directional regret (\texttt{Analyze}, \S\ref{sec:generalized_rsi}--\S\ref{sec:paired_regret}).} We formulate reference-free harness adaptation over a frozen model using paired directional regret on pre-execution task checklists. We prove that dense checklist grading removes the $1/\epsilon$ passive-discovery penalty on rare task invariants (Theorem~\ref{thm:targeting}), that paired priority ranks every fresh regression above every non-regressed criterion in the single-rollout binary case and pins exactly the mastered criteria (Proposition~\ref{prop:dominance}), and that gating keeps self-review scores non-decreasing on each round's gated task buffer (Theorem~\ref{thm:gate}).
    \item \textbf{GRPO-inspired candidate pooling and contrastive prompt evolution without weight updates (\texttt{Propose}, \S\ref{sec:grpo_pool}).} Inspired by group-relative reinforcement learning \cite{shao2024deepseekmath,cai2025trainingfree}, \texttt{GUARD} samples a parallel group of candidate harnesses from one domain-agnostic seed prompt, evaluates them on the same task batch, and ranks them by critic-free group-standardized fitness (a selection step, not a gradient step) while contrasting the best and worst group candidates to rewrite proposal prompts across rounds. In a controlled single-model ablation at equal budget, parallel two-candidate pooling beats sequential single-candidate search and cuts rejected gate rounds by $64\%$, while deploying a single harness in one pass at test time.
    \item \textbf{Soft-HURT promotion gating and continual generalization (\texttt{Gate}, \S\ref{sec:rrsi_gate}, \S\ref{sec:results}, Appendix~\ref{app:additional}).} Pairing Soft-HURT promotion with in-round \texttt{[iff <condition>]} precondition repair yields $19/19$ non-regressive execution-gated promotions across \textsc{EvoHarnessBench}, \textsc{Drift Gauntlet}, and the professional suites. \texttt{GUARD} is the only evaluated method with no negative backward or forward transfer on hosted \textsc{EvoHarnessBench} (FWT is zero because the gate abstained at later stages) and cuts tokens per task by $54.5\%$ versus the base agent, incurs no stream-mean drop across all $12$ block transitions and zero post-block forgetting on all four \textsc{Drift Gauntlet} streams, and ties the memory-pollution floor on \textsc{PAST-Bench}.
\end{itemize}

\section{Background and Problem Formulation}
\label{sec:generalized_rsi}
\paragraph{Agents, harnesses, and task streams.}
We model an autonomous agent $A = (\pi_\theta, W_{t,g})$ as a frozen language model policy $\pi_\theta$ paired with an inspectable \emph{harness state} $W_{t,g}$ at adaptation round $t \in \{0, 1, \dots, T\}$ ($W_{0,g}$ empty; subscript $g$ omitted when clear) \cite{lee2026metaharness,xia2026rrsi}. Tasks arrive from groups $g \in \mathcal{G}$ indexing roles, task families, or, when a continual stream hides group labels and drift boundaries \cite{chen2026continual}, the pre-execution (\emph{Turn-0}, before the solver's first action) observable tool and schema signature of query $q \in \mathcal{Q}_g$. We call $W_{t,g}$ a \emph{non-parametric} policy component because it is editable text rather than model weights: conditioning on it steers rollouts $\tau \sim \pi_\theta(\cdot \mid q, W_{t,g})$, while $\theta$ stays fixed during both adaptation and testing. In reinforcement-learning terms, an action is a harness edit, its reward is the gated self-review fitness $R_m$ (\S\ref{sec:grpo_pool}), and only $W$ is optimized. While this formulation accommodates general text or code harnesses \cite{meng2026rsigym}, we instantiate $W_{t,g}$ as structured Markdown files, termed a \emph{harness playbook}, organizing workflow instructions, entity-grouped domain rules, reusable skills (\texttt{SKILL.md}), persistent memories (\texttt{MEMORY.md}), and optional pre-submit checks, with specialized exceptions scoped by explicit \texttt{[iff <condition>]} preconditions.

\paragraph{Reference-free continual evolution and non-regressive RSI.}
Let $V^{\star}(q, \tau) \in [0, 100]$ denote the unobserved task evaluator, unavailable throughout adaptation. Observing only task prompts, tool interfaces, execution trajectories, and runtime states in interaction history $\mathcal{H}_{t,g}$, a reference-free harness \emph{evolver} $\mathcal{E}$ recursively updates the incumbent across adaptation rounds:
{\small
\begin{equation}
W_{t,g} \;=\; \mathcal{E}\big(W_{t-1,g},\, \mathcal{H}_{t,g}\big), \qquad t = 1, \dots, T \le T_{\max}.
\label{eq:harness_evolver}
\end{equation}
}%
At test time, evolution freezes and each unseen query $q \in \mathcal{Q}_g$ executes under the single promoted harness $W_{T,g}$ to maximize $\mu_g(W_{T,g}) = \mathbb{E}_{q \in \mathcal{Q}_g,\, \tau \sim \pi_\theta(\cdot \mid q, W_{T,g})}[V^{\star}(q, \tau)]$. Because $\mathcal{E}$ updates $W_{t,g}$ adaptively from finite online rollouts without $V^{\star}$, continual generalization under Eq.~\eqref{eq:harness_evolver} requires improving $\mu_g$ without local overfitting on solved tasks or catastrophic forgetting of mastered behaviors as $t$ and task distributions shift \cite{xia2026rrsi,chen2026continual}: scalar round scores mix edit-induced regressions with unresolved capability gaps and compound negative transfer across rounds.

\paragraph{What reference-free does and does not mean.}
Throughout this paper, \emph{reference-free} (and \emph{grader-free}) is an operational constraint on the adaptation loop $\mathcal{E}$: neither the solver, the self-reviewer, nor the distiller ever accesses held-out reference answers, test cases, hidden verifiers $V^{\star}$, oracle skills, or official grading rubrics. It does \emph{not} mean learning from zero supervision. Specifically, adaptation draws on four observable sources of structure: (i)~benchmark-authored training task prompt distributions, (ii)~public tool declarations and environment schemas, (iii)~the agent's own interaction trajectories and observable sandbox states (with no human expert demonstrations), and (iv)~structured criterion-level judgments from the language-model self-reviewer (\S\ref{sec:self_review}).

\section{Related Work}
\label{sec:related_work}
Table~\ref{tab:novelty_comparison} (Appendix~\ref{app:related_table}) compares \texttt{GUARD} with representative self-improvement methods and environments across the components they optimize, their closed-loop mechanisms, and their adaptation and context scalability.

\paragraph{Verbal reflection, memories, and playbooks.} Reflexion \cite{shinn2023reflexion}, ExpeL \cite{zhao2024expel}, ReasoningBank \cite{ouyang2026reasoningbank}, and AWM \cite{wang2024awm} store verbal reflections or workflows; MemGPT \cite{packer2023memgpt}, Voyager \cite{wang2023voyager}, and WikiSkill \cite{tang2026wikiskill} maintain memories, skill libraries, and append-only wikis; Dynamic Cheatsheet \cite{suzgun2025dynamic} and ACE \cite{zhang2025ace} incrementally update playbooks; and Context Language Models (CLM) \cite{shao2026clm} update model weights to compact context via Group Relative Policy Optimization (GRPO) \cite{shao2024deepseekmath}, whereas Training-Free GRPO \cite{cai2025trainingfree} summarizes group rollouts into an experience library with frozen weights. \texttt{GUARD} differs in what forms the group: it scores $M$ candidate harnesses on the same task batch rather than rollouts of one fixed context, contrasts the best and worst candidates to rewrite proposal prompts, and accepts the winner only through a regression-aware gate. AutoGuide \cite{fu2024autoguide}, AutoManual \cite{chen2024automanual}, and CLIN \cite{majumder2024clin} distill conditional guidelines, rule manuals, or causal memories from experience; \texttt{GUARD}'s \texttt{[iff]} rules are likewise conditional, but enter the harness only through regression-aware ranking and gating. Because these updates neither isolate edit-induced regressions nor gate per-task drops, they remain vulnerable to the plasticity--pollution and plasticity--stability trade-offs measured by \textsc{PAST-Bench} \cite{xue2026pastbench}, \textsc{EvoHarnessBench} \cite{evoharnessbench2026}, and \textsc{Drift Gauntlet} \cite{chen2026continual}. Accepting an update only when prior behavior does not degrade echoes regression-free model updates \cite{xie2021regression} and gradient-constrained continual learning \cite{lopezpaz2017gem}, which rely on labels or gradients that a reference-free harness loop does not have.

\paragraph{Unsupervised environment design and zero-data RL.} UED \cite{dennis2020emergent,jiang2021prioritized,jiang2021replay} prioritizes high-regret POMDP levels via antagonist regret (PAIRED \cite{dennis2020emergent}), maximal Monte Carlo (\textsc{MaxMC}) regret \cite{jiang2021prioritized,jiang2021replay,parker2022evolving}, or learnability $p(1-p)$ \cite{rutherford2024no}, whereas we transfer regret from level curation to directional criterion ranking over pre-execution task checklists \cite{zhou2023ifeval} without an antagonist. Zero-data RL self-play (Absolute Zero \cite{zhao2025absolute}, Self-Challenging Agents \cite{zhou2025selfchallenging}, R-Zero \cite{huang2026rzero}, and Agent0 \cite{xia2026agent0}) eliminates human labels via code verifiers or majority voting, yet requires weight gradients (precluding black-box APIs), falters on stateful tool mutations, and updates opaque model weights instead of inspectable conditional rules.

\paragraph{Prompt, workflow, and harness optimization.} Prompt optimizers such as DSPy \cite{khattab2023dspy}, TextGrad \cite{yuksekgonul2024textgrad}, TextReg \cite{fu2026textreg}, OPRO \cite{yang2023large}, WizardLM \cite{xu2023wizardlm}, Promptbreeder \cite{fernando2023promptbreeder}, EvoPrompt \cite{guo2024evoprompt}, and GEPA \cite{agrawal2026gepa} search instructions via scalar objectives, textual gradients, or Pareto archives; TextReg \cite{fu2026textreg} filters low-recurrence patches under labeled single-turn supervision, which can suppress valid edge-case exceptions. Workflow/code optimizers ADAS \cite{hu2025adas}, AFlow \cite{zhang2025aflow}, and Darwin G{\"o}del Machine (DGM) \cite{zhang2026dgm} mutate code via archives \cite{lehman2011novelty,mouret2015mapelites} or MCTS against validation ground truth, while AutoHarness \cite{lou2026autoharness}, Self-Harness \cite{zhang2026selfharness}, Meta-Harness \cite{lee2026metaharness}, RRSI \cite{xia2026rrsi}, and \textsc{Harness}\textsuperscript{2} \cite{xu2026harness2} evolve harnesses around frozen weights (alongside MLGym \cite{nathani2025mlgym}, PostTrainBench \cite{rank2026posttrainbench}, RSIBench-Data \cite{meng2026rsibenchdata}, and RSIGym \cite{meng2026rsigym}). Among open-world designs, \textsc{Harness}\textsuperscript{2} probes per-task harnesses online from $W_0$ at extra test-time rollout cost, RRSI's score-band gate leaves per-task drops unbounded, and persistent harnesses (ReasoningBank, GEPA, Meta-Harness) suffer severe forgetting under hidden drift \cite{chen2026continual}. \texttt{GUARD} instead evolves a single persistent harness in a reference-free loop and deploys it at $O(1)$ context in $T$ and single-pass (\texttt{pass@1}) test cost (Appendix~\ref{app:complexity}).

\section{\texttt{GUARD}: Regret-Guided Harness Evolution}
\label{sec:methodology}
During adaptation, \texttt{GUARD} instantiates the reference-free evolver $\mathcal{E}$ in Eq.~\eqref{eq:harness_evolver} as a four-stage closed loop (Figure~\ref{fig:unified_architecture}; Algorithm~\ref{alg:unified_loop}; constants in Appendix~\ref{app:instantiation}): pre-execution checklist derivation and two-layer self-review (\S\ref{sec:self_review}), paired regret ranking (\texttt{Analyze}, \S\ref{sec:paired_regret}), contrastive candidate pooling (\texttt{Propose}, \S\ref{sec:grpo_pool}), and Soft-HURT promotion (\texttt{Gate}, \S\ref{sec:rrsi_gate}), with formal guarantees in \S\ref{sec:theory}. At test time, this adaptation loop is disabled and each unseen task runs a single rollout under the frozen harness $W_{T,g}$.

\subsection{Reference-Free Self-Review via Task Checklists}
\label{sec:self_review}

\paragraph{Pre-execution task checklist $\mathcal{C}(q)$.} Extending IFEval's verifiable constraint decomposition \cite{zhou2023ifeval} and checklist-based evaluation \cite{qin2024infobench,cook2024tick,viswanathan2025checklists} to multi-turn tool workflows, the self-reviewer reads \emph{only} the task prompt, visible tool declarations, and schema once at Turn~0 (before the solver's first action) to derive and cache a fixed checklist $\mathcal{C}(q)$ of atomic binary predicates (\emph{criteria} $c \in \mathcal{C}(q)$; Appendix~\ref{app:instantiation}) across two classes: (i)~\emph{execution-verifiable} predicates (clean sandbox execution, valid entity IDs/columns, idempotent mutations) and (ii)~\emph{semantic} predicates (honoring every prompt constraint, recipient, format, and safety rule without applying superseded rules).

\paragraph{Two-layer grading and verification asymmetry.} Layer~1 checks execution predicates deterministically on an ephemeral sandbox copy; Layer~2 grades semantic predicates individually from the prompt, output, and tool/state trace, returning $r(\tau, c) \in \{0,1\}$ and a failure rationale. Over graded predicates $\mathcal{C}^{\mathrm{gr}}(\tau) \subseteq \mathcal{C}(q)$ ($|\mathcal{C}^{\mathrm{gr}}(\tau)| \ge 1$; unconfigured checks excluded; an unparseable verdict counts as failed or falls back to deterministic trace checks), trajectory $\tau$ receives criterion score $S(\tau)$ and strict score $S^{\mathrm{strict}}(\tau)$:
{\small
\begin{equation}
\begin{aligned}
S(\tau) &= 100 \cdot \frac{1}{|\mathcal{C}^{\mathrm{gr}}(\tau)|} \sum_{c \in \mathcal{C}^{\mathrm{gr}}(\tau)} r(\tau, c), \\
S^{\mathrm{strict}}(\tau) &= \begin{cases}
0 & \text{on a blocking sandbox error,}\\
100 & \text{if all graded criteria pass,}\\
S(\tau)/2 & \text{otherwise.}
\end{cases}
\end{aligned}
\label{eq:self_review_scores}
\end{equation}
}%
Pre-execution decomposition uses \emph{generation--verification asymmetry}: long-horizon generation couples $N$ constraints and suffers ``seen-but-not-carried'' omissions, while checking one atomic predicate after execution requires localized $O(1)$ attention grounded by Layer~1 sandbox states. Cross-task $\Delta_{\mathrm{hurt},t}$ and Soft-HURT gating then prevent single-instance overfitting \cite{madaan2024self}.

\subsection{Paired Regret Ranking (\texttt{Analyze} Stage)}
\label{sec:paired_regret}

\begin{definition}[Criterion obedience and paired regret]
\label{def:paired_decomposition}
For trajectories $\mathcal{T}_c(W)$ grading criterion $c$ under $W$ ($\mathcal{T}_c(W_{t-1}) = \emptyset$ skipped), criterion obedience is
{\small
\begin{equation}
\mathrm{obey}(c; W) \;=\; \frac{1}{|\mathcal{T}_c(W)|} \sum_{\tau \in \mathcal{T}_c(W)} r(\tau, c) \;\in\; [0, 1].
\label{eq:obey_def}
\end{equation}
}%
Entering round $t \ge 1$ with incumbent $W_{t-1}$, comparing the two most recent incumbents $W_{t-2}$ and $W_{t-1}$ on the same tasks gives:
{\small
\begin{equation}
\begin{aligned}
\Delta_{\mathrm{hurt},t}(c) &= \max\!\big(0,\; \mathrm{obey}(c; W_{t-2}) - \mathrm{obey}(c; W_{t-1})\big), \\
\Delta_{\mathrm{gap},t}(c) &= 1 - \mathrm{obey}(c; W_{t-1}), \\
\mathrm{Priority}_t(c) &= \Delta_{\mathrm{hurt},t}(c) + \Delta_{\mathrm{gap},t}(c) \;\in\; [0, 2],
\end{aligned}
\label{eq:paired_split}
\end{equation}
}%
where $\Delta_{\mathrm{hurt},t}(c) = 0$ if $t = 1$ or $c$ is not graded under both compared playbooks.
\end{definition}
Both terms are non-negative and parameter-free: a regressed criterion receives its incumbent gap plus its drop, outranking equally obeyed non-regressed criteria (Proposition~\ref{prop:dominance}); with one binary verdict per criterion after a promotion, $\mathrm{Priority}_t(c) \in \{2, 1, 0\}$ separates regressions ($1 \to 0$), unsolved gaps ($0 \to 0$), and mastered checks ($1 \to 1$ or $0 \to 1$). In regret terms, $\Delta_{\mathrm{gap},t}$ is the incumbent's shortfall against an oracle harness that satisfies every criterion, and $\Delta_{\mathrm{hurt},t}$ is the negative transfer introduced by the latest promotion.

\paragraph{Rejected rounds, diagnostic lists, and trajectory contrasts.} After a rejection ($W_{t-1} = W_{t-2}$), execution scoring instead takes $\Delta_{\mathrm{hurt},t}$ from $W_{t-1}$ to the rejected winner $\widetilde{W}_{t-1}$ (or its repair $\widetilde{W}^{\mathrm{rep}}_{t-1}$), so the next proposal sees which criteria the refused edit broke, while $\Delta_{\mathrm{gap},t}$ and pinning stay anchored to the incumbent; critic scoring always compares consecutive incumbents (\S\ref{sec:experiments}). The signs of the two terms split criteria in the regret bundle $\mathcal{B}_{t,g}$ into three diagnostic lists, each sorted by descending $\mathrm{Priority}_t(c)$ (ties broken by $\Delta_{\mathrm{hurt},t}(c)$): (1)~\emph{Regressed} ($\Delta_{\mathrm{hurt},t} > 0$), asking the proposer to wrap offending rules in \texttt{[iff <trigger>]} preconditions instead of discarding specialized fixes \cite{fu2026textreg}; (2)~\emph{Unsolved Gap} ($\Delta_{\mathrm{hurt},t} = 0, \Delta_{\mathrm{gap},t} > 0$), targeting persistent gaps; and (3)~\emph{Mastered} ($\mathrm{Priority}_t = 0$, i.e., $\mathrm{obey}(c; W_{t-1}) = 1 \land \Delta_{\mathrm{hurt},t} = 0$), pinning obeyed rules against overwriting. These lists serve as diagnostic annotations, not hand-coded edit operators: the proposer receives them with the incumbent, trajectory contrasts, and the ledger window, and writes candidate playbooks end-to-end. Finally, on tasks with differing scored rollouts, $\mathcal{B}_{t,g}$ pairs top and bottom trajectories $(\tau^{+}, \tau^{-})$ (ordered by $S^{\mathrm{strict}}$, ties by $S$) whenever
{\small
\begin{equation}
\begin{aligned}
&\Delta^{\mathrm{strict}}(\tau^{+},\tau^{-}) = S^{\mathrm{strict}}(\tau^{+}) - S^{\mathrm{strict}}(\tau^{-}) > 0 \\
\lor\;\; &\Delta^{\mathrm{cont}}(\tau^{+},\tau^{-}) = S(\tau^{+}) - S(\tau^{-}) \ge \delta_{\mathrm{con}},
\end{aligned}
\label{eq:contrast_condition}
\end{equation}
}%
($\delta_{\mathrm{con}} = 10$). This condition captures partial checklist progress even when blocking errors or critical caps flatten $S^{\mathrm{strict}}$.

\subsection{Group-Relative Candidate Pooling and Contrastive Evolution (\texttt{Propose} Stage)}
\label{sec:grpo_pool}
In reinforcement learning, Group Relative Policy Optimization (GRPO) \cite{shao2024deepseekmath} estimates policy advantage without a value network by sampling a group of rollouts from a shared state and standardizing their rewards within the group. Following the training-free group-relative perspective \cite{cai2025trainingfree}, the \texttt{Propose} stage applies this principle to non-parametric harness optimization without updating model weights: it samples $M \ge 2$ candidate harnesses in parallel from $(W_{t-1,g}, \mathcal{B}_{t,g})$, selects the winner by within-group standardized fitness ($\arg\max_m \hat{A}_m$), and contrasts the best and worst group candidates ($\mathsf{ContrastEvolve}$) to update the strategy-prompt pool $\mathcal{P}_{t,g} = \{s_{t,1}, \dots, s_{t,M}\}$ across rounds. Unlike GRPO, \texttt{GUARD} takes no gradient step and uses no clipping or KL penalty; $\hat{A}_m$ only ranks candidates and picks the contrast pair $(m^+, m^-)$, and the Soft-HURT gate (\S\ref{sec:rrsi_gate}) plays the role of a trust region. What the group buys is credit isolation without a learned critic: all $M$ candidates are scored on the same $(W_{t-1,g}, \mathcal{B}_{t,g}, \mathcal{Q}_g)$, so batch-level task difficulty cancels out.

\paragraph{Uniform initialization and scoring.} Round~1 initializes all $M$ parallel candidates with the same domain-agnostic prompt $s_0$ (\texttt{SHARED\_SEED\_PROMPT}, Appendix~\ref{app:seed_prompt}):
\begin{equation}
\mathcal{P}_{1,g} \;=\; \{s_{1,1}, \dots, s_{1,M}\}, \qquad s_{1,1} = \dots = s_{1,M} = s_0,
\label{eq:shared_seed}
\end{equation}
drawing diversity without operator priors from sampling $\widetilde{W}^{(m)}_{1,g} = \mathsf{Distill}(s_0, \mathcal{B}_{1,g}, W_{0,g})$. Over $N_{\mathrm{rep}} \ge 1$ replicates $\{\tau_{q,r}\}_{r=1}^{N_{\mathrm{rep}}}$ and noise floor $\delta_{\mathrm{noise}} \ge 0$, task scores and helped/hurt sets are
{\small
\begin{equation}
\begin{aligned}
S_q(W) &= \frac{1}{N_{\mathrm{rep}}} \sum_{r=1}^{N_{\mathrm{rep}}} S(\tau_{q,r}), \\
S^{\mathrm{strict}}_q(W) &= \frac{1}{N_{\mathrm{rep}}} \sum_{r=1}^{N_{\mathrm{rep}}} S^{\mathrm{strict}}(\tau_{q,r}), \\
\mathcal{Q}^{\pm}(\widetilde{W}, W) &= \big\{q \in \mathcal{Q}_g : \pm\big(S_q(\widetilde{W}) - S_q(W)\big) > \delta_{\mathrm{noise}}\big\},
\end{aligned}
\label{eq:task_sets}
\end{equation}
}%
with means $\bar{S}(W), \bar{S}^{\mathrm{strict}}(W)$ supplied by \emph{execution scoring} ($M$ group rollouts/round) or \emph{critic scoring} (one joint pass over accumulated trajectories on stateful suites). With $\mathrm{Perf}_m = \bar{S}^{\mathrm{strict}}(\widetilde{W}^{(m)}_t) + \tfrac{1}{4}\bar{S}(\widetilde{W}^{(m)}_t)$ and $\mathcal{Q}^{\pm}_m = \mathcal{Q}^{\pm}(\widetilde{W}^{(m)}_t, W_{t-1})$, scalar and standardized fitness are
{\small
\begin{equation}
\begin{aligned}
R_m &= \mathrm{Perf}_m - \beta\,|\mathcal{Q}^{-}_m| + \gamma\,|\mathcal{Q}^{+}_m|, \qquad \beta > \gamma \ge 0, \\
\hat{A}_m &= \big(R_m - \mu_R\big) \big/ \sigma_R, \\
\sigma_R &= \sqrt{\tfrac{1}{M}\textstyle\sum_{m=1}^{M}(R_m - \mu_R)^2} + 10^{-6},
\end{aligned}
\label{eq:grpo_reward}
\end{equation}
}%
($\mu_R = \frac{1}{M}\sum_m R_m$), where $\beta > \gamma$ penalizes hurts over helps and $m^{\star} = \arg\max_m \hat{A}_m = \arg\max_m R_m$ selects $\widetilde{W}_t = \widetilde{W}^{(m^{\star})}_t$ (with $M = 2$, $\hat{A}_m \approx \pm 1$ whenever $R_1 \neq R_2$, so selection keeps the higher-fitness candidate).

\paragraph{Contrastive prompt evolution, deployment, and complexity.} Selecting distinct best and worst candidates in the group,
\begin{equation}
m^{+} = \arg\max_{m \in \{1,\dots,M\}} R_m, \qquad m^{-} = \arg\min_{m \neq m^{+}} R_m
\label{eq:best_worst_select}
\end{equation}
(ties broken by $\bar{S}$, $|\mathcal{Q}^{-}_m|$, candidate index; mirroring \textsc{MaxMC} spread \cite{jiang2021replay}), $\mathsf{ContrastEvolve}$ scrubs task IDs to \texttt{<TASK\_ID>}, contrasts $(s_{t,m^{\pm}}, \widetilde{W}^{(m^{\pm})}_{t,g}, R_{m^{\pm}}, |\mathcal{Q}^{\pm}_{m^{\pm}}|)$, and outputs \texttt{learnings}, an exploitation prompt $s^{+}_{t+1}$, and a precondition-narrowing prompt $s^{-}_{t+1}$, updating
\begin{equation}
\mathcal{P}_{t+1,g} \;=\; \{s^{+}_{t+1}\}^{\lfloor M/2 \rfloor} \;\cup\; \{s^{-}_{t+1}\}^{M - \lfloor M/2 \rfloor}.
\label{eq:contrastive_pool_update}
\end{equation}
At test time, $\mathcal{P}_{t,g}$, the self-reviewer, and $\mathcal{L}_g$ are discarded: deployment runs a single ($1\times$) rollout per task (\texttt{pass@1} on all primary benchmarks) under frozen $W_{T,g}$. During adaptation, parallel $M$-candidate pooling isolates credit on identical $(W_{t-1,g}, \mathcal{B}_{t,g}, \mathcal{Q}_g)$, bootstraps Round-1 contrasts, and reduces single-sample gate rejections (\S\ref{sec:calibration_ablation}, Appendix~\ref{app:seam_ablation}). Execution scoring costs $O(TMQ)$ rollouts with $O(TQ)$ parallel rollout depth (or $O(TQ)$ online rollouts under critic scoring), while bounded playbooks $B_W$, ledger window $L_{\max}$, and capped bundles $B_{\mathcal{B}}$ keep proposer and test context $O(1)$ in $T$ (Table~\ref{tab:novelty_comparison}, Appendix~\ref{app:complexity}).

\subsection{Soft-HURT Promotion Gate and Rejection Ledger (\texttt{Gate} Stage)}
\label{sec:rrsi_gate}
\begin{algorithm}[tp]
\caption{\texttt{GUARD}: reference-free regret-guided harness RSI during adaptation (at test time, each unseen task executes a single pass under frozen $W_{T,g}$). $\mathsf{Distill}(s, \mathcal{B}, W)$ compiles a candidate harness playbook under strategy prompt $s$ from regret bundle $\mathcal{B}$ and incumbent $W$, and $\mathsf{ContrastEvolve}$ updates the best and worst strategy prompts via Eq.~\eqref{eq:contrastive_pool_update}.}
\label{alg:unified_loop}
\fontsize{8.2}{9.8}\selectfont
\begin{algorithmic}[1]
\Require groups $g \in \mathcal{G}$ with task sets $\mathcal{Q}_g$; pool size $M \ge 2$; uniform seed prompt $s_0$ (\texttt{SHARED\_SEED\_PROMPT}); round budget $T_{\max}$
\State \textbf{Output:} final incumbent $W_{T,g}$ of every group $g$, where $T \le T_{\max}$ is the last round run
\State Derive and cache pre-execution checklists $\mathcal{C}(q)$ from public prompts and tool schemas (\S\ref{sec:self_review})
\For{each group $g \in \mathcal{G}$}
    \State $W_{0,g} \gets$ empty playbook; \; $\mathcal{L}_g \gets$ empty ledger; \; $\mathcal{P}_{1,g} \gets \{s_0, \dots, s_0\}$ ($M$ parallel candidates, Eq.~\eqref{eq:shared_seed})
    \State Solve and self-review $\mathcal{Q}_g$ under $W_{0,g}$ on fixed-batch suites \Comment{measured baseline}
\EndFor
\For{$t = 1, \dots, T_{\max}$}
    \For{each group $g \in \mathcal{G}$}
        \State On stateful critic-scored suites, execute round-$t$ online episode under $W_{t-1,g}$ and self-review
        \State Compute $\Delta_{\mathrm{hurt},t}(c)$, $\Delta_{\mathrm{gap},t}(c)$, $\mathrm{Priority}_t(c)$ from $W_{t-2,g}, W_{t-1,g}$ (\S\ref{sec:paired_regret}) \Comment{Eq.~\eqref{eq:paired_split}}
        \State Partition $c$ into Regressed ($\Delta_{\mathrm{hurt},t} > 0$) / Unsolved Gap ($\Delta_{\mathrm{hurt},t} = 0 \land \Delta_{\mathrm{gap},t} > 0$) / Mastered ($\mathrm{Priority}_t = 0$, pinned)
        \State $\mathcal{B}_{t,g} \gets$ incumbent $W_{t-1,g}$, routed partitions, $(\tau^{+},\tau^{-})$ contrasts, active windowed $\mathcal{L}_g$
        \For{each candidate $m \in \{1, \dots, M\}$ with strategy prompt $s_{t,m} \in \mathcal{P}_{t,g}$}
            \State $\widetilde{W}^{(m)}_{t,g} \gets \mathsf{Distill}(s_{t,m}, \mathcal{B}_{t,g}, W_{t-1,g})$
            \State Score $\widetilde{W}^{(m)}_{t,g}$ on $\mathcal{Q}_g$ \Comment{per-candidate execution rollouts or joint critic pass}
        \EndFor
        \State $m^{\star} \gets \arg\max_m \hat{A}_m$ ($= \arg\max_m R_m$) \Comment{Eq.~\eqref{eq:grpo_reward}}
        \State $\widetilde{W}_{t,g} \gets \widetilde{W}^{(m^{\star})}_{t,g}$ \Comment{group-standardized fitness}
        \If{Soft-HURT gate (\S\ref{sec:rrsi_gate}) accepts $\widetilde{W}_{t,g}$}
            \State $W_{t,g} \gets \widetilde{W}_{t,g}$
        \ElsIf{$1 \le |\mathcal{Q}^{-}_{m^{\star}}| \le h_{\mathrm{rep}}$ and repaired $\widetilde{W}^{\mathrm{rep}}_{t,g}$ passes Soft-HURT} \Comment{\texttt{[iff]} repair}
            \State $W_{t,g} \gets \widetilde{W}^{\mathrm{rep}}_{t,g}$
        \Else
            \State $W_{t,g} \gets W_{t-1,g}$; \; record deduplicated active rejection in $\mathcal{L}_g$
        \EndIf
        \State Select $m^{+} \gets \arg\max_m R_m$ \Comment{Eq.~\eqref{eq:best_worst_select}}
        \State Select $m^{-} \gets \arg\min_{m \ne m^{+}} R_m$
        \State $\big(\texttt{learnings}, s^{+}_{t+1}, s^{-}_{t+1}\big) \gets$ \Comment{task IDs scrubbed}\newline
        \hspace*{1em}$\mathsf{ContrastEvolve}\big((s_{t,m^{\pm}}, \widetilde{W}^{(m^{\pm})}_{t,g}, R_{m^{\pm}}, |\mathcal{Q}^{\pm}_{m^{\pm}}|)\big)$
        \State $\mathcal{P}_{t+1,g} \gets \{s^{+}_{t+1}\}^{\lfloor M/2 \rfloor} \cup \{s^{-}_{t+1}\}^{M - \lfloor M/2 \rfloor}$ \Comment{Eq.~\eqref{eq:contrastive_pool_update}}
    \EndFor
    \State $T \gets t$
    \If{optional early-stop criterion fires (disabled in all reported runs)} \textbf{break} \EndIf
\EndFor
\State \Return $W_{T,g}$ for every group $g \in \mathcal{G}$
\end{algorithmic}
\end{algorithm}

\paragraph{Soft-HURT promotion gate.} Using per-task scores $S_q$ and hurt set $\mathcal{Q}^{-}(\widetilde{W}, W)$ from \eqref{eq:task_sets}, the group-averaged regression penalty beyond $\delta_{\mathrm{noise}}$ is
{\small
\begin{equation}
\mathrm{Hurt}(\widetilde{W}, W) = \frac{1}{|\mathcal{Q}_g|} \sum_{q \in \mathcal{Q}^{-}(\widetilde{W}, W)} \big(S_q(W) - S_q(\widetilde{W})\big),
\label{eq:hurt_penalty}
\end{equation}
}%
and the Soft-HURT rule promotes a candidate over $W_{t-1}$ only when net gain exceeds induced hurt:
{\small
\begin{equation}
\begin{aligned}
&\mathrm{Promote}(\widetilde{W}, W_{t-1}) \iff \\
&\quad \bar{S}(\widetilde{W}) - \bar{S}(W_{t-1}) > \mathrm{Hurt}(\widetilde{W}, W_{t-1}).
\end{aligned}
\label{eq:soft_hurt_gate}
\end{equation}
}%
Parameterized by single-task noise floor $\delta_{\mathrm{noise}} \ge 0$, hurt-task budget $h_{\max} \ge 0$, and trade-off margin $\mu_{\min} \ge 0$, Soft-HURT admits $\widetilde{W}_t$ when $|\mathcal{Q}^{-}(\widetilde{W}_t, W_{t-1})| \le h_{\max}$, $|\mathcal{Q}^{+}(\widetilde{W}_t, W_{t-1})| \ge \mathbb{I}[|\mathcal{Q}^{-}(\widetilde{W}_t, W_{t-1})| > 0]$, and
{\small
\begin{equation}
\begin{aligned}
&\bar{S}(\widetilde{W}_t) - \bar{S}(W_{t-1}) > \\
&\quad \max\!\big(\mathrm{Hurt}(\widetilde{W}_t, W_{t-1}),\, \mu_{\min}\,\mathbb{I}[|\mathcal{Q}^{-}(\widetilde{W}_t, W_{t-1})| > 0]\big).
\end{aligned}
\label{eq:soft_hurt_param}
\end{equation}
}%
When $\widetilde{W}_t$ fails solely due to isolated regressions ($1 \le |\mathcal{Q}^{-}(\widetilde{W}_t, W_{t-1})| \le h_{\mathrm{rep}}$), \texttt{GUARD} triggers at most one in-round \texttt{[iff]} scope-guard repair $\widetilde{W}^{\mathrm{rep}}_t$ and promotes $\widetilde{W}^{\mathrm{rep}}_t$ if its re-scored trajectories satisfy the same Soft-HURT rule ($\mu_{\min} = 0$ on repaired candidates; suite parameters in Appendix~\ref{app:instantiation}). Any other candidate is rejected, so every promoted update obeys \eqref{eq:soft_hurt_gate}.

\paragraph{Rejection ledger and stopping.} Rejected gate decisions log deduplicated active entries to $\mathcal{L}_g$ (an on-disk JSONL log; candidate-level runners store gate metrics and scrubbed rule excerpts with read-back cap $L_{\max} \in \{12, 40\}$, while the professional-suite runner stores active criterion-removal excerpts; Appendix~\ref{app:instantiation}), rendered into $\mathcal{B}_{t+1,g}$ at $O(1)$ prompt cost in $T$ so refused edits are not repeated verbatim while \texttt{[iff]} refinements remain eligible. Rejected candidates are never re-executed ($\Delta_{\mathrm{hurt},t+1}$ reuses their gate scores), and optional early stopping is disabled in all reported runs.

\subsection{Theoretical Guarantees}
\label{sec:theory}
These results characterize the self-review objective on gated task buffers; held-out accuracy is evaluated empirically in \S\ref{sec:results}. Proofs of Theorem~\ref{thm:gate}, Theorem~\ref{thm:targeting}, and Proposition~\ref{prop:dominance} appear in Appendix~\ref{app:proofs}. For Theorem~\ref{thm:gate}, fix group $g \in \mathcal{G}$ with fixed $\mathcal{Q}_g$: conditional on prior history, replicate errors $\{S(\tau_{q,r}) - \mu_q(W)\}_{q \in \mathcal{Q}_g,\, r \le N_{\mathrm{rep}}}$ ($N_{\mathrm{rep}} \ge 1$ under execution rollouts) are independent across $q, r$ and $\sigma$-sub-Gaussian (mean $\bar{S}(W)$, expectation $\mu(W) = \frac{1}{|\mathcal{Q}_g|}\sum_{q \in \mathcal{Q}_g}\mu_q(W)$). Over $T_{\max}$ rounds, at most $n_{\mathrm{eval}} = 1 + T_{\max}(M + \mathbb{I}_{\mathrm{rep}})$ candidate harnesses are evaluated per group ($\mathbb{I}_{\mathrm{rep}} \in \{0,1\}$ for in-round \texttt{[iff]} repair) against recorded incumbent scores without re-running incumbents; let $\Delta_t = \bar{S}(W_t) - \bar{S}(W_{t-1})$.

\begin{tcolorbox}[resultbox]
\begin{theorem}[Non-Regressive Gating and Concentration]
\label{thm:gate}
Under the gate of \eqref{eq:soft_hurt_gate} and the replicate setup above:
\begin{enumerate}[leftmargin=*, label=(\roman*), itemsep=1pt, topsep=2pt, parsep=0pt]
    \item \textbf{No downward drift:} $\Delta_t > 0$ if round $t$ promotes and $\Delta_t = 0$ otherwise, so
    \begin{equation}
    \bar{S}(W_T) = \bar{S}(W_0) + \sum_{t=1}^{T} \Delta_t \;\ge\; \bar{S}(W_0).
    \label{eq:thm_gate_mono}
    \end{equation}
    \item \textbf{Regressions are paid for:} on every promoted round, with $\mathcal{Q}^{-}_t = \mathcal{Q}^{-}(W_t, W_{t-1})$ the tasks dropping by more than $\delta_{\mathrm{noise}}$,
    \begin{equation}
    \sum_{q \in \mathcal{Q}^{-}_t} \big(S_q(W_{t-1}) - S_q(W_t)\big) \;<\; |\mathcal{Q}_g|\, \Delta_t,
    \label{eq:thm_gate_paid}
    \end{equation}
    and the left-hand side is zero whenever $\mathcal{Q}^{-}_t = \emptyset$ (in particular when $h_{\max} = 0$).
    \item \textbf{Concentration:} for any $\alpha \in (0,1)$, with probability at least $1 - \alpha$, simultaneously for every group $g \in \mathcal{G}$ (with $\log$ the natural logarithm),
    {\small
    \begin{equation}
    \begin{aligned}
    \mu(W_T) &\ge \mu(W_0) + \sum_{t=1}^{T}\!\Delta_t \\
    &\quad - 2\sigma\!\sqrt{\frac{2\log(2n_{\mathrm{eval}}|\mathcal{G}|/\alpha)}{|\mathcal{Q}_g| N_{\mathrm{rep}}}}.
    \end{aligned}
    \label{eq:thm_gate_conc}
    \end{equation}
    }%
    \item \textbf{Factored speedup:} if group $g$ promotes in each round independently with probability $a_g$, per-group gating advances a fraction $\frac{1}{|\mathcal{G}|}\sum_g a_g$ of groups per round vs.\ $\prod_g a_g$ under a global gate ($a^{1-|\mathcal{G}|}\times$ for $a_g = a$).
\end{enumerate}
\end{theorem}
\end{tcolorbox}

The union bound in (iii) covers all $n_{\mathrm{eval}}$ evaluated harnesses per group (paying only $\log(M + \mathbb{I}_{\mathrm{rep}})$ inside $n_{\mathrm{eval}}$ for within-group selection and repair) and applies to execution rollouts ($N_{\mathrm{rep}} \ge 1$; main-text suites use $N_{\mathrm{rep}} = 1$ and Appendix~\ref{app:additional} uses $N_{\mathrm{rep}} = 2$) instead of single critic calls. Here $\mu$ is the expected self-review score on $\mathcal{Q}_g$, and Eq.~\eqref{eq:thm_gate_conc} serves as an asymptotic sample-complexity scaling law ($|\mathcal{Q}_g|N_{\mathrm{rep}} = \Omega(\sigma^2 \log(n_{\mathrm{eval}}|\mathcal{G}|/\alpha)/\Delta^2)$ samples keep the selection optimism of recorded incumbent scores, at most twice the deviation term in \eqref{eq:thm_gate_conc}, below a target gain $\Delta$) rather than a non-vacuous finite-sample certificate at our benchmark task counts. When incumbents are re-scored on growing episode histories (\textsc{PAST-Bench}) or online task blocks shift across stages (\textsc{EvoHarnessBench}, \textsc{Drift Gauntlet}), (i)--(ii) hold within each round's gated task buffer rather than telescoping unconditionally across distribution shifts, which motivates pre-execution interface-signature routing and replay-buffer anchor tasks (\S\ref{sec:drift_gauntlet}).

\begin{tcolorbox}[resultbox]
\begin{theorem}[Targeted Repair vs.\ Passive Discovery]
\label{thm:targeting}
Consider $J$ invariants exercised by $\mathcal{Q}_g$ and violated by $W_0$ (an invariant is a task-independent rule that the playbook must encode). Compare two observability regimes: a \emph{passive end-task} learner that observes an invariant only when a coarse task failure surfaces it with probability $p_j \le \epsilon$, versus \emph{dense pre-execution checklist} grading where every violated invariant is captured by a derived criterion evaluated each round. Let $T_{\mathrm{passive}}$ be the rounds until passive surfacing observes all $J$ invariants, $T_{\mathrm{target}}$ the rounds until targeted repair resolves all $J$ invariants, and $H_J = \sum_{\ell=1}^{J} 1/\ell$ ($H_0 = 0$).
\begin{enumerate}[leftmargin=*, label=(\roman*), itemsep=1pt, topsep=2pt, parsep=0pt]
    \item \textbf{Passive lower bound:} if, independently across invariants and rounds, invariant $j$ surfaces in each round with probability $p_j \le \epsilon$ for $\epsilon \in (0, 1)$, then
    {\small
    \begin{equation}
    \begin{aligned}
    \mathbb{E}[T_{\mathrm{passive}}] &\ge \frac{H_J}{-\log(1-\epsilon)} \ge \frac{1-\epsilon}{\epsilon} H_J \\
    &= \Omega\!\left(\frac{\log J}{\epsilon}\right)\!,
    \end{aligned}
    \label{eq:thm_passive}
    \end{equation}
    }%
    where the $\Omega((\log J)/\epsilon)$ asymptotic form holds for $\epsilon \le 1/2$.
    \item \textbf{Targeted upper bound:} if in each round (a) every unresolved invariant fails a derived criterion ($\mathrm{Priority}_t(c) > 0$), (b) the winner addresses at least $\min(b_{\min}, u)$ of the $u$ still unresolved invariants, for an integer $b_{\min} \ge 1$, (c) each is resolved and admitted independently with probability $\ge \rho > 0$, and (d) resolved invariants stay resolved, then
    \begin{equation}
    \mathbb{E}[T_{\mathrm{target}}] \;\le\; \frac{J}{\rho\, b_{\min}} + 1 + \frac{H_{b_{\min}-1}}{\rho}.
    \label{eq:thm_target}
    \end{equation}
\end{enumerate}
\end{theorem}
\end{tcolorbox}

Dense per-criterion grading (assumption~(a)) removes the $1/\epsilon$ dependence regardless of the specific priority formula: any ranking assigning positive weight to failing criteria, including $1 - \mathrm{obey}$, satisfies~(a), and under all-or-nothing candidate admission with joint bundle pass probability $\rho_{\mathrm{bundle}} > 0$, $\mathbb{E}[T_{\mathrm{target}}] \le \lceil J / b_{\min} \rceil / \rho_{\mathrm{bundle}} \le \frac{J}{\rho_{\mathrm{bundle}} b_{\min}} + \frac{\mathbb{I}[b_{\min} > 1]}{\rho_{\mathrm{bundle}}}$. What uniquely distinguishes paired priority is regression handling, formalized next.

\begin{tcolorbox}[resultbox]
\begin{proposition}[Strict Regression Dominance under Multi-Rollout Grading]
\label{prop:dominance}
Let obedience \eqref{eq:obey_def} average the verdicts of $K_{\mathrm{gr}} \ge 1$ graded rollouts, so that it may be fractional. When round $t-1$ promoted, write $o_c = \mathrm{obey}(c; W_{t-1})$ and $\delta_c = \mathrm{obey}(c; W_{t-2}) - o_c$ ($\delta_c := 0$ if $t = 1$ or $\mathcal{T}_c(W_{t-2}) = \emptyset$), so $\mathrm{Priority}_t(c) = (1 - o_c) + \max(0, \delta_c)$ by \eqref{eq:paired_split}. Let $c$ be a regressed criterion ($\delta_c > 0$) and $c'$ a non-regressed one ($\delta_{c'} \le 0$).
\begin{enumerate}[leftmargin=*, label=(\roman*), itemsep=1pt, topsep=2pt, parsep=0pt]
    \item \textbf{Dominance at equal or higher obedience:} if $o_{c'} \ge o_c$, then $\mathrm{Priority}_t(c) - \mathrm{Priority}_t(c') \ge \delta_c > 0$, however small the regression (for example, one additional failing rollout).
    \item \textbf{Exact trade-off:} $\mathrm{Priority}_t(c) > \mathrm{Priority}_t(c')$ if and only if $o_c - o_{c'} < \delta_c$. In the single-rollout binary case ($K_{\mathrm{gr}} = 1$), a fresh regression ($1 \to 0$, priority $2$) therefore outranks every non-regressed criterion (priority at most $1$).
    \item \textbf{Exact incumbent pinning set:} under \eqref{eq:paired_split} in any round (promoted or rejected), $\mathrm{Priority}_t(c) = 0$ holds if and only if $\mathrm{obey}(c; W_{t-1}) = 1$ and $\Delta_{\mathrm{hurt},t}(c) = 0$; in particular, no regressed criterion and no criterion failed by incumbent $W_{t-1}$ is ever pinned.
\end{enumerate}
\end{proposition}
\end{tcolorbox}

Frequency ranking $1 - \mathrm{obey}(c; W_{t-1})$ satisfies~(iii) but fails~(i), tying newly broken criteria with static gaps of equal obedience; conversely, signed difference $\mathrm{obey}(c; W_{t-2}) - \mathrm{obey}(c; W_{t-1})$ satisfies~(i) but fails~(iii), assigning $0$ to unchanged failures ($0 \to 0$) alongside mastered invariants ($1 \to 1$). Paired priority satisfies both parameter-free. For $K_{\mathrm{gr}} \ge 3$ rollouts (or $|\mathcal{T}_c| \ge 3$ tasks), a partial regression ($1 \to 2/3$, priority $2/3$) scores below a $0 \to 0$ gap ($1$) by~(ii), and tying occurs at $K_{\mathrm{gr}} = 2$ ($1 \to 0.5$, priority $1$) or post-rejection when $\mathrm{obey}(c; W_{t-1}) = 1$ ($\Delta_{\mathrm{hurt},t} = 1, \Delta_{\mathrm{gap},t} = 0$). This trade-off is intended: a criterion that the incumbent always fails outranks one that regressed on a minority of rollouts, while the Regressed list still surfaces every regression to the proposer (\S\ref{sec:paired_regret}).

\section{Experimental Setup}
\label{sec:experiments}

To evaluate continual harness RSI against local overfitting, memory pollution, and catastrophic forgetting across escalating non-stationarity, we test \texttt{GUARD} on three primary public benchmarks: (1)~\textsc{PAST-Bench} \cite{xue2026pastbench} (cross-session memory retention, precision, and SOP updates), (2)~\textsc{EvoHarnessBench} (\texttt{EnterpriseOps-Gym}) \cite{evoharnessbench2026} (stage-wise continual transfer across evolving Model Context Protocol [MCP] tools and skills), and (3)~\textsc{Drift Gauntlet} \cite{chen2026continual} (continual learning without forgetting across hidden-boundary workflow, topic, domain, and compound drift), complemented by two rubric-dense professional suites and controlled component ablations (\texttt{Analyze}, \texttt{Propose}, and \texttt{Gate}) in Appendix~\ref{app:additional}. All three primary benchmarks compare the unified six-arm suite (Appendix~\ref{app:baselines}): \texttt{No-Evolve} ($W_0$), native memory/skill persistence (\textsc{Hermes+} \cite{xue2026pastbench} / \textsc{MemToolAgent} \cite{evoharnessbench2026}), \textsc{RRSI} \cite{xia2026rrsi}, \textsc{GEPA} \cite{agrawal2026gepa}, \textsc{Meta-Harness} \cite{lee2026metaharness}, and \texttt{GUARD}. To enforce strict reference-free hygiene on public datasets, our runner strips all held-out evaluator fields (reference answers, test cases, verifiers, oracle skills, and rubrics), aborts if any survives, and snapshots initial sandboxes so adaptation sees only public task inputs and execution states; after adaptation, every arm freezes its promoted harness $W_T$ and reports standard single-rollout (\texttt{pass@1}) held-out accuracy across all three primary benchmarks ($N_{\mathrm{rep}} = 2$ frozen \texttt{DEV} trials on the Appendix~\ref{app:additional} professional suites), with zero test-time candidate pooling, self-review, or retries.

\paragraph{Primary single-model \texttt{Gemini 4} RSI and applicability to teacher--student knowledge distillation.}
Our primary continual-RSI experiments use a single frozen \texttt{Gemini 4} model across every role (solver, self-reviewer, distiller, and evolver, with zero stronger teacher): this powers the \textsc{EvoHarnessBench} Official Hosted SDK (Table~\ref{tab:evoharness_main}), all four hidden-boundary continual streams on \textsc{Drift Gauntlet} (Table~\ref{tab:drift_gauntlet_main}), and the controlled dual-axis \texttt{Propose}-stage ablation (Table~\ref{tab:propose_seam_ablation}), alongside two full 12-family model-identical \textsc{PAST-Bench} suites (pure \texttt{claude-opus-5-5} and pure \texttt{gemini-3.8-flash}; \S\ref{sec:pastbench}, Appendix~\ref{app:baselines}). Complementing this primary single-model regime, we deliberately include a cross-model teacher--student configuration on \textsc{PAST-Bench} (Table~\ref{tab:pastbench}), the five-domain local gym (Tables~\ref{tab:evoharness_domains} and~\ref{tab:evoharness_seam_ablation}), and the Appendix~\ref{app:additional} professional suites, pairing a frozen \texttt{gemini-3.8-flash} student solver with a \texttt{claude-opus-5-5} teacher reviewer and distiller shared identically across all evolving baselines. Evaluating both regimes serves a deliberate purpose: the single-model \texttt{Gemini 4} suites isolate strictly self-referential continual improvement without external teacher supervision, while the cross-model suites show that \texttt{GUARD} also applies to teacher--student knowledge-distillation harness pipelines, distilling stronger-model reasoning into compact student skills and memories with low memory pollution and no regressed domain.

\paragraph{\textsc{PAST-Bench} and critic-scored selection.} Across 12 families spanning memory (\texttt{SM01}--\texttt{02}, \texttt{EP01}), procedural reuse (\texttt{PC01}, \texttt{PC03}--\texttt{04}), proactive gathering (\texttt{PG01}--\texttt{02}), and updates (\texttt{PC02}, \texttt{SM04}, \texttt{SM06}--\texttt{07}; 1 round per teach/update episode), official grading reports \emph{Task} (plus procedural SOP \emph{Proc.}), \emph{write precision} (\emph{W.Prec.}), \emph{Recall}, \emph{update correctness} (\emph{Upd.}), \emph{retention} (\emph{Ret.}), \emph{pollution} (\emph{Poll.}), \emph{mechanism} (\emph{Mech.}, mean of W.Prec., Recall, Upd., Ret., and $1 - \text{Poll.}$), and $\text{Evolve}$ (12-family mean of $\Delta_{\text{task},f} \times \text{Mech.}_f$). To avoid re-executing stateful multi-session histories, \texttt{GUARD} uses critic scoring (\S\ref{sec:grpo_pool}), evaluating the incumbent and candidate states in a single pass over accumulated session trajectories instead of re-running prior episodes (Appendix~\ref{app:baselines}); in place of per-task pre-execution checklists, the critic scores four fixed suite-level criteria (history grounding, SOP fidelity, \texttt{[iff]} scope, and superseded-rule replacement; Box~\ref{box:turn0_reviewer_prompts}(c)) that overlap with the benchmark's write-path pillars, while baselines keep their native reflection or search prompts. Because a single critic pass scores both the incumbent and the candidates without re-execution, it drives both selection and gating ($24/24$ round winners admitted across the $12$ families; execution-gated promotion fidelity is evaluated separately across \textsc{EvoHarnessBench}, \textsc{Drift Gauntlet}, and the professional suites in \S\ref{sec:calibration_ablation}), while held-out official grading and paired per-family comparisons (Table~\ref{tab:pastbench_paired}) evaluate downstream write-path quality.

\begin{table*}[t]
\centering
\footnotesize
\setlength{\tabcolsep}{5.2pt}
\caption{\textbf{\textsc{PAST-Bench} results across 12 families (matched backbone, unified six-arm suite; metrics in \S\ref{sec:experiments}).} All columns except Poll.\ are higher-is-better; bold marks the best persistent arm. \texttt{GUARD} leads Proc., Mech., Upd., W.Prec., and Evolve, stays within $0.001$ of the Ret.\ ceiling, and ties the $0.167$ Poll.\ floor, with positive mean gains on Mech., Upd., and W.Prec.\ against every persistent baseline and $0/48$ family losses on W.Prec.\ or Poll.\ across all four baseline comparisons (paired differences and sign tests in Table~\ref{tab:pastbench_paired}).}
\label{tab:pastbench}
\begin{tabular}{lccccccccc}
\toprule
Arm & Task & Proc. & Mech. & Upd. & W.Prec. & Recall & Ret. & Poll.$\downarrow$ & Evolve \\
\midrule
\texttt{No-Evolve} ($W_0$) & 0.372 & 0.427 & 0.454 & 0.278 & 0.000 & 0.000 & 0.990 & 0.000 & 0.000 \\
\textsc{Hermes+} \cite{xue2026pastbench} & 0.852 & 0.808 & 0.840 & 0.791 & 0.724 & 0.854 & \textbf{1.000} & \textbf{0.167} & 0.421 \\
\textsc{RRSI} \cite{xia2026rrsi} & 0.809 & 0.714 & 0.725 & 0.901 & 0.438 & 0.811 & 0.974 & 0.500 & 0.350 \\
\textsc{GEPA} \cite{agrawal2026gepa} & \textbf{0.858} & 0.815 & 0.837 & 0.777 & 0.719 & \textbf{0.858} & 0.996 & \textbf{0.167} & 0.424 \\
\textsc{Meta-Harness} \cite{lee2026metaharness} & 0.812 & 0.677 & 0.739 & 0.903 & 0.479 & 0.814 & 0.998 & 0.500 & 0.360 \\
\midrule
\rowcolor{LightBlue} \textbf{\texttt{GUARD} (Ours)} & 0.854 & \textbf{0.830} & \textbf{0.878} & \textbf{0.924} & \textbf{0.776} & 0.856 & 0.999 & \textbf{0.167} & \textbf{0.436} \\
\bottomrule
\end{tabular}
\vspace{-6pt}
\end{table*}

\paragraph{\textsc{EvoHarnessBench} and \textsc{Drift Gauntlet} protocols.} On \textsc{EvoHarnessBench}, at stage $v_k \in \{\texttt{v1}, \dots, v_V\}$, arms adapt on $\mathcal{D}_{\text{train}}^{(v_k)}$ via execution scoring (\S\ref{sec:self_review}) and evaluate on $\mathcal{D}_{\text{test}}^{(v_1 \dots v_k)}$, forming matrix $A_{k,j}$ ($k \ge j$). Following \cite{evoharnessbench2026}, we report final \emph{Overall ACC}, strict \emph{Solved}, \emph{Cohort ACC} $A_{V,j}$, backward transfer $\text{BWT} = \sum_{j<V} n_j (A_{V,j} - A_{j,j}) / \sum_{j<V} n_j$, forward transfer $\text{FWT} = \sum_{k \ge 2} n_k (A_{k,k} - A^{\text{pre}}_{k,k}) / \sum_{k \ge 2} n_k$ (local gym uses unweighted cohort BWT and stage FWT $A_{k,k} - A^{W_0}_{k,k}$), and per-task tokens, calls, and latency on: (1) the \emph{Official Hosted SDK} (\texttt{evolving\_tools/calendar}, \texttt{v1}--\texttt{v3}, $18$ train / $43$ test tasks) under single-model \texttt{Gemini 4} for all roles; and (2) the \emph{Full 5-Domain} local gym ($236$ test / $97$ train tasks, $16$ stages across \texttt{Calendar}, \texttt{Drive}, \texttt{ITSM} \texttt{v1}--\texttt{v4}, \texttt{CSM}, \texttt{HR}), plus a controlled single-model \texttt{Gemini 4} \texttt{Propose}-stage ablation across dual-axis \texttt{ITSM} ($N{=}132$ test tasks; Appendix~\ref{app:seam_ablation}). On \textsc{Drift Gauntlet} \cite{chen2026continual}, agents evolve sequentially across $K \in \{2,3,4\}$ unannounced task blocks ($30$ online adaptation tasks/block, $30$ held-out test tasks/group, $12$ blocks total) under \emph{single-model} \texttt{Gemini 4} for every role across four continual drift scenarios spanning BFCL tool calling, BIRD text-to-SQL, and ScienceWorld interactive environments: \emph{workflow drift} (\texttt{BFCL-Miss} $\to$ \texttt{BFCL-Base}, $K{=}2$), \emph{topic drift} (\texttt{BIRD-F1} $\to$ \texttt{BIRD-Code} $\to$ \texttt{BIRD-TP}, $K{=}3$), \emph{domain drift} (\texttt{ScienceWorld} $\to$ \texttt{BFCL-Base} $\to$ \texttt{BIRD-F1}, $K{=}3$), and \emph{compound drift} (\texttt{BIRD-Code} $\to$ \texttt{BIRD-F1} $\to$ \texttt{BFCL-Miss} $\to$ \texttt{BFCL-Base}, $K{=}4$). From the $(K{+}1) \times K$ checkpoint matrix $A[t, k]$ ($t \in \{0,\dots,K\}, k \in \{1,\dots,K\}$) and seed vector $B_k = A[0, k]$, we report \cite{chen2026continual}: \emph{Final Mean} $\bar{A}_K = \frac{1}{K}\sum_{k=1}^{K} A[K, k]$, terminal gain $\Delta_K = \bar{A}_K - \bar{B}$, average gain $\bar{\Delta} = \frac{1}{K}\sum_{t=1}^{K}(\bar{A}_t - \bar{B})$, within-block \emph{plasticity} $P = \frac{1}{K}\sum_{k=1}^{K}(A[k, k] - A[k{-}1, k])$, \emph{pre-block interference} $I = \frac{1}{K-1}\sum_{k=2}^{K}\frac{1}{k-1}\sum_{t=1}^{k-1}(A[t, k] - B_k)$, worst-case \emph{post-block forgetting} $F = \frac{1}{K-1}\sum_{k=1}^{K-1}(\min_{t>k} A[t, k] - A[k, k])$, and \emph{stability} $S = (I + F)/2$. \texttt{GUARD} keys playbook sections to each task's pre-execution observable environment/schema signature and renders only the matching section, whereas baselines keep one global playbook (lexically retrieved skills for \textsc{Hermes+}); for every arm, held-out rollouts are memoized per task and rendered playbook (Appendix~\ref{app:instantiation}). Arms are compared strictly within each benchmark and model regime.

\section{Main Experimental Results}
\label{sec:results}

\subsection{\textsc{PAST-Bench}: Cross-Session Memory Write Path}
\label{sec:pastbench}
\begin{figure*}[t]
\centering
\includegraphics[width=\textwidth]{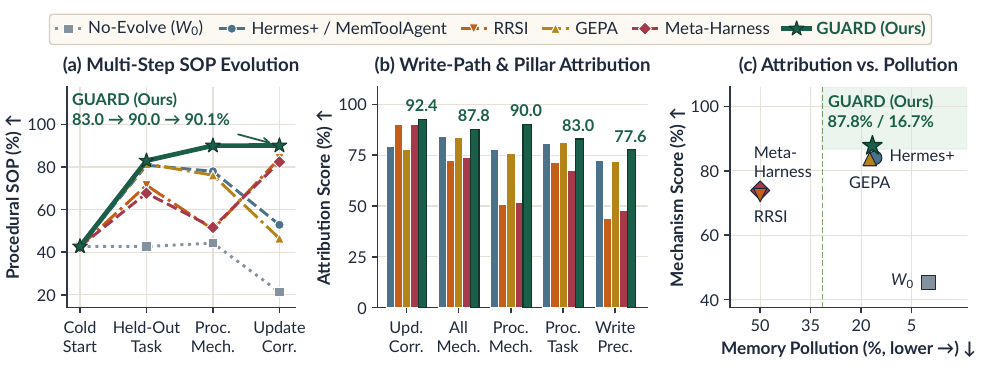}
\vspace{-6pt}
\caption{\textbf{\textsc{PAST-Bench} procedural SOP and memory write-path attribution (12 families).} (a) Multi-step SOP families (\texttt{PC01}, \texttt{PC03}, \texttt{PC04}; horizontal axis lists metrics alongside cold-start $42.7\%$, not training stages). (b) Write-path and pillar breakdown. (c) Mechanism attribution vs.\ memory pollution ($0.0\%$ pollution on $10/12$ families under \texttt{GUARD}).}
\label{fig:pastbench}
\vspace{-6pt}
\end{figure*}

While 5 of 12 families are single-rule tasks where every persistent arm reaches $0.984$--$1.000$, write-path attribution and multi-step procedural SOPs (\texttt{PC01}, \texttt{PC03}, \texttt{PC04}) expose a sharp plasticity--pollution trade-off (Table~\ref{tab:pastbench}, Figure~\ref{fig:pastbench}; Tables~\ref{tab:pastbench_procedural}--\ref{tab:pastbench_paired}). Unscoped accumulators (\textsc{Hermes+}, \textsc{GEPA}) reach $80.8\%$--$81.5\%$ procedural task at $0.167$ overall pollution, but retain stale clauses on updates ($52.9\%$ / $46.4\%$ procedural; $0.791$ / $0.777$ overall). Conversely, aggressive rewriters (\textsc{RRSI}, \textsc{Meta-Harness}) lift update correctness to $0.901$--$0.903$ ($82.4\%$--$84.1\%$ procedural) while tripling pollution ($0.500$ overall, $1.000$ procedural) and depressing write precision ($0.438$--$0.479$). Pairing regret ranking, rule pinning, and \texttt{[iff]} guards, \texttt{GUARD} leads procedural task ($83.0\%$), procedural mechanism ($90.0\%$ vs.\ at most $78.0\%$), and update ($90.1\%$) at $0.000$ procedural pollution; across all 12 families ($12/12 > W_0$), it leads update ($0.924$, $+13.3\text{pp}$ vs.\ \textsc{Hermes+}, $+14.8\text{pp}$ vs.\ \textsc{GEPA}), mechanism ($0.878$, $p{=}0.016$ vs.\ \textsc{RRSI}), write precision ($0.776$, $p{=}0.031$ vs.\ \textsc{RRSI}), and evolve score ($0.436$) within $0.004$ of top task score at the $0.167$ pollution floor ($0.000$ on $10/12$ families, and $0/48$ family losses on W.Prec.\ or Poll.\ across all four baselines; Table~\ref{tab:pastbench_paired}). These gains persist without any teacher--student model gap (Appendix~\ref{app:baselines}): under model-identical all-\texttt{claude-opus-5-5} execution, \texttt{GUARD} ranks first overall on task score ($0.870$ vs.\ $\le 0.866$, $+0.523$ over $W_0$), recall ($0.871$), and evolve score ($0.458$) with the sole lowest memory pollution ($0.167$ vs.\ $0.208$--$0.250$); under all-\texttt{gemini-3.8-flash}, it lifts task score from $0.371$ to $0.852$, ties for lowest pollution ($0.167$), and leads latent rule induction (\texttt{PC03}: $0.748$ vs.\ $0.510$--$0.653$ on reflective/search baselines, $+9.5$ to $+23.8\text{pp}$).

\begin{table*}[t]
\centering
\footnotesize
\setlength{\tabcolsep}{4.0pt}
\caption{\textbf{\textsc{EvoHarnessBench} Official Hosted SDK comparison under unified \texttt{Gemini 4} (\texttt{evolving\_tools/\allowbreak calendar}, stages \texttt{v1}--\texttt{v3}, 43 test tasks; \S\ref{sec:experiments}).} Cohort sizes are $n_1{=}29, n_2{=}10, n_3{=}4$ (Cohort~\texttt{v3} reflects single-task granularity); rollout hygiene notes in Appendix~\ref{app:instantiation}. The \texttt{GUARD} row is the final of five hosted configurations evaluated on this test split (mean $77.09\%$, range $75.57$--$80.80\%$; all five above every baseline). Paired over the $43$ test tasks (exact sign-flip test), the ACC gain of \texttt{GUARD} is significant against $W_0$ ($p = 0.008$; $8$ tasks better, $0$ worse), \textsc{GEPA} ($p = 0.023$), and \textsc{MemToolAgent} ($p = 0.010$), and directional against \textsc{RRSI} ($p = 0.094$) and \textsc{Meta-Harness} ($p = 0.11$; 95\% bootstrap CI $[-0.4, +14.5]\text{pp}$). Bold marks best per column ($^{=}$exact Cohort-\texttt{v2} tie).}
\label{tab:evoharness_main}
\resizebox{\textwidth}{!}{%
\begin{tabular}{@{}llccccccccc@{}}
\toprule
\textbf{Method (Unified \texttt{Gemini 4})} & \textbf{Adaptation Family} & \textbf{ACC} & \textbf{Solved} & \textbf{Cohort \texttt{v1 / v2 / v3} ACC} & \textbf{BWT Score} & \textbf{BWT Pass} & \textbf{FWT} & \textbf{Tokens / Task$\downarrow$} & \textbf{Calls$\downarrow$} & \textbf{Latency$\downarrow$} \\
\midrule
\texttt{No-Evolve} ($W_0$, Control) & Static ReAct Baseline & 71.91\% & 48.84\% & 70.46\% / 79.17\% / 64.29\% & $-6.00\text{pp}$ & $-5.13\text{pp}$ & -- & 131{,}032 & 14.47 & 221.6s \\
\textsc{MemToolAgent} \cite{evoharnessbench2026} & Memory (\#1 family) & 68.00\% & 51.16\% & 61.95\% / \textbf{84.17\%}$^{=}$ / 71.43\% & $-4.02\text{pp}$ & $+2.56\text{pp}$ & $-4.98\text{pp}$ & 82{,}426 & 15.84 & 239.2s \\
\textsc{RRSI} \cite{xia2026rrsi} & Reflective Harness Rewrite ($M{=}1$) & 74.28\% & 51.16\% & 75.06\% / 73.17\% / 71.43\% & $-0.45\text{pp}$ & $0.00\text{pp}$ & $+1.79\text{pp}$ & 84{,}519 & 13.19 & 613.1s \\
\textsc{GEPA} \cite{agrawal2026gepa} & Prompt (\#1 family) & 70.69\% & 48.84\% & 68.45\% / 71.17\% / 85.71\% & $-3.97\text{pp}$ & $-2.56\text{pp}$ & \textbf{+13.61pp} & 85{,}156 & 17.05 & \textbf{159.1s} \\
\textsc{Meta-Harness} \cite{lee2026metaharness} & Agent Code (\#1 family) & 74.52\% & 51.16\% & 73.42\% / 81.83\% / 64.29\% & \textbf{+5.83pp} & \textbf{+5.13pp} & $-13.21\text{pp}$ & 78{,}705 & 14.19 & 220.0s \\
\midrule
\rowcolor{LightBlue} \textbf{\texttt{GUARD} (Ours)} & \textbf{Regret-Guided Harness RSI} & \textbf{80.80\%} & \textbf{58.14\%} & \textbf{78.47\%} / \textbf{84.17\%}$^{=}$ / \textbf{89.29\%} & $+0.85\text{pp}$ & $+2.56\text{pp}$ & $\phantom{+}0.00\text{pp}$ & \textbf{59{,}651} ($-54.5\%$) & \textbf{11.21} ($-22.5\%$) & 233.8s \\
\bottomrule
\end{tabular}%
}
\vspace{-6pt}
\end{table*}

\subsection{\textsc{EvoHarnessBench}: Continual Evolution Across Tools and Skills}
\label{sec:evoharness}

\begin{figure*}[t]
\centering
\includegraphics[width=\textwidth]{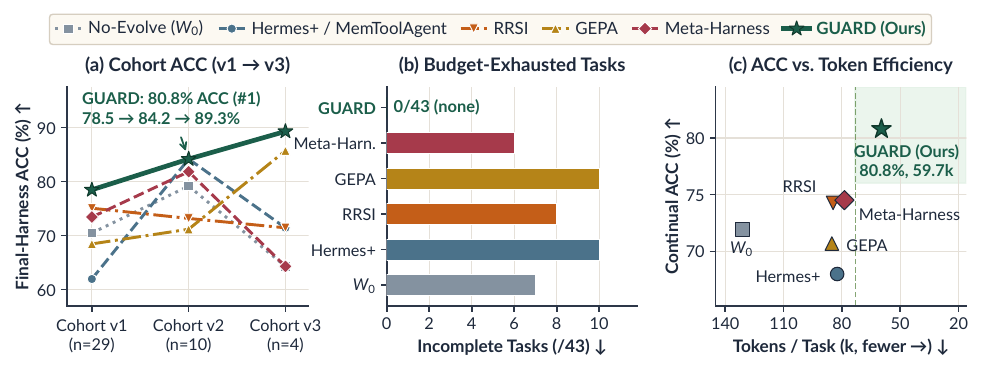}
\vspace{-6pt}
\caption{\textbf{\textsc{EvoHarnessBench} Hosted SDK evaluation under unified \texttt{Gemini 4} (\texttt{evolving\_tools/calendar}, \texttt{v1}--\texttt{v3}).} (a) Final-harness cohort progression $A_{3,j}$. (b) Tasks exhausting the eight-step budget ($0/43$ for \texttt{GUARD} vs.\ $6$--$10$ on baselines). (c) Overall accuracy vs.\ mean tokens per task.}
\label{fig:evoharness}
\vspace{-6pt}
\end{figure*}

\paragraph{Unified \texttt{Gemini 4} hosted comparison, efficiency, and 5-domain local gym.} On the hosted SDK (Table~\ref{tab:evoharness_main}, Figure~\ref{fig:evoharness}), \texttt{GUARD} achieves the highest accuracy ($80.80\%$; $+8.89\text{pp}$ over $W_0$, paired $p = 0.008$; $+6.28\text{pp}$ over \textsc{Meta-Harness}, $p = 0.11$ on $43$ tasks) and solve rate ($58.14\%$, $25/43$ vs.\ $\le 22/43$), leading Cohorts~\texttt{v1} ($78.47\%$) and \texttt{v3} ($89.29\%$) and tying Cohort~\texttt{v2} ($84.17\%$), while passing $182/221$ official verifier checks ($82.4\%$ vs.\ $156$--$170/221$ [$70.6\%$--$76.9\%$] on baselines) with the fewest zero-score tasks ($3/43$ vs.\ $5$--$7/43$). Soft-HURT promoted the Stage-\texttt{v1} playbook ($2{,}315$ chars, $83.33 \to 100$ source self-review, $0$ hurt) and abstained at Stages~\texttt{v2}--\texttt{v3}, so $\text{FWT} = 0.00\text{pp}$ holds by abstention and the entire lift stems from that Stage-\texttt{v1} playbook. \texttt{GUARD} is the only arm with no negative entry across BWT Score, BWT Pass, and FWT ($+0.85$ / $+2.56$ / $0.00\text{pp}$), whereas \textsc{RRSI} drops $-4.50\text{pp}$ on Cohort~\texttt{v2} at Stage~\texttt{v3} ($77.67\% \to 73.17\%$, BWT Score $-0.45\text{pp}$) and \textsc{GEPA} ($-3.97\text{pp}$ BWT), \textsc{Meta-Harness} ($-13.21\text{pp}$ FWT), and \textsc{MemToolAgent} ($-4.02$ / $-4.98\text{pp}$) each suffer negative transfer. Together with two-layer gating (\S\ref{sec:calibration_ablation}) and a compact playbook ($2{,}315$ vs.\ $4{,}538$--$10{,}013$ chars), \texttt{GUARD} reaches $0/43$ eight-step budget exhaustions (vs.\ $7/43$ in $W_0$ and $6$--$10/43$ on baselines), which cuts tokens/task by $54.5\%$ ($59{,}651$ vs.\ $131{,}032$) and MCP calls by $22.5\%$ while solving $9.75$ tasks per $1\text{M}$ tokens ($1.50\times$ the best baseline, $6.50$ on \textsc{Meta-Harness}, and $2.61\times$ $W_0$, $3.73$). Every one of the five hosted \texttt{GUARD} configurations evaluated in our study ($75.57\%$--$80.80\%$ ACC, mean $77.09\%$, $171$--$182/221$ checks passed, $0$--$3/43$ budget exhaustions; Appendix~\ref{app:instantiation}) beats every external baseline ($\le 74.52\%$ ACC, $\le 170/221$ checks, $6$--$10/43$ exhaustions). Across all five local-gym domains ($236$ test tasks; Table~\ref{tab:evoharness_domains}), \texttt{GUARD} improves upon or matches $W_0$ in final accuracy across domains (macro $65.81\% \to 66.46\%$, $+1.00\text{pp}$ mean FWT), lifting multi-stage skill tracks such as \texttt{ITSM} ($+1.63\text{pp}$ ACC, $+8.47\text{pp}$ Solved) and \texttt{CSM} ($+1.60\text{pp}$ ACC, with a minor trade-off on strict all-or-nothing solve rate) while safely abstaining where candidate edits do not clear Soft-HURT.

\subsection{\textsc{Drift Gauntlet}: Continual RSI Under Hidden Stream Drift}
\label{sec:drift_gauntlet}

\begin{table*}[t]
\centering
\footnotesize
\setlength{\tabcolsep}{3.8pt}
\caption{\textbf{\textsc{Drift Gauntlet} \cite{chen2026continual} continual RSI comparison under single-model \texttt{Gemini 4} across all four hidden-boundary drift streams ($12$ blocks, $360$ adaptation / $360$ held-out test rollouts per arm; metrics in \S\ref{sec:experiments}, full 24-row breakdown in Table~\ref{tab:drift_gauntlet_full}).} \emph{Block Drops} counts checkpoints $t \in \{1,\dots,K\}$ where stream-mean accuracy $\bar{A}_t < \bar{A}_{t-1}$. Bold marks the best evolving arm per column ($^{=}$exact stream tie).}
\label{tab:drift_gauntlet_main}
\resizebox{\textwidth}{!}{%
\begin{tabular}{@{}lccccccc|cccc@{}}
\toprule
& \multicolumn{7}{c|}{\textbf{Four-Stream Macro-Average}} & \multicolumn{4}{c}{\textbf{Per-Stream Final Mean ($F$)}} \\
\cmidrule(lr){2-8} \cmidrule(l){9-12}
\textbf{Method (Single-Model \texttt{Gemini 4})} & \textbf{Final Mean} & \textbf{$\Delta_K$} & \textbf{$\bar{\Delta}$} & \textbf{Plast.\ $P$} & \textbf{Forgetting $F$} & \textbf{Stab.\ $S$} & \textbf{Block Drops$\downarrow$} & \textbf{Workflow} & \textbf{Topic} & \textbf{Domain} & \textbf{Compound} \\
\midrule
\texttt{No-Evolve} ($W_0$, Control) & 69.92\% & $0.00\text{pp}$ & $0.00\text{pp}$ & $0.00\text{pp}$ & $0.00\text{pp}$ & $0.00\text{pp}$ & 0 / 12 & 65.00\% ($0.0$) & 67.78\% ($0.0$) & 76.05\% ($0.0$) & 70.84\% ($0.0$) \\
\textsc{Hermes+} / \textsc{MemToolAgent} & 70.31\% & $+0.40\text{pp}$ & $+1.09\text{pp}$ & $+2.67\text{pp}$ & $-5.59\text{pp}$ & $-2.66\text{pp}$ & 5 / 12 & 65.00\% ($-13.3$) & \textbf{70.00\%}$^{=}$ ($+1.7$) & 76.26\% ($-1.8$) & 70.00\% ($-8.9$) \\
\textsc{RRSI} \cite{xia2026rrsi} & 67.92\% & $-2.00\text{pp}$ & $-0.47\text{pp}$ & $+4.49\text{pp}$ & $-8.11\text{pp}$ & $-6.02\text{pp}$ & 8 / 12 & 60.00\% ($-20.0$) & 66.67\% ($-3.3$) & 79.17\% ($-5.8$) & 65.84\% ($-3.3$) \\
\textsc{GEPA} \cite{agrawal2026gepa} & 72.40\% & $+2.48\text{pp}$ & $+2.18\text{pp}$ & $+2.80\text{pp}$ & $-1.09\text{pp}$ & $\mathbf{-0.35\text{pp}}$ & 3 / 12 & \textbf{70.00\%}$^{=}$ ($\mathbf{+3.3}$) & 67.78\% ($-1.7$) & \textbf{79.31\%} ($-2.7$) & 72.50\% ($-3.3$) \\
\textsc{Meta-Harness} \cite{lee2026metaharness} & 71.21\% & $+1.30\text{pp}$ & $+0.15\text{pp}$ & $+3.28\text{pp}$ & $-0.07\text{pp}$ & $-2.51\text{pp}$ & 4 / 12 & 63.33\% ($0.0$) & 68.89\% ($-5.0$) & 79.29\% ($\mathbf{+1.4}$) & 73.33\% ($\mathbf{+3.3}$) \\
\midrule
\rowcolor{LightBlue} \textbf{\texttt{GUARD} (Ours)} & \textbf{73.22\%} & $\mathbf{+3.30\text{pp}}$ & $\mathbf{+2.71\text{pp}}$ & $\mathbf{+4.55\text{pp}}$ & $\mathbf{0.00\text{pp}}$ & $-0.93\text{pp}$ & \textbf{0 / 12} & \textbf{70.00\%}$^{=}$ ($0.0$) & \textbf{70.00\%}$^{=}$ ($0.0$) & 78.69\% ($0.0$) & \textbf{74.17\%} ($0.0$) \\
\bottomrule
\end{tabular}%
}
\vspace{-6pt}
\end{table*}

\begin{figure*}[t]
\centering
\includegraphics[width=\textwidth]{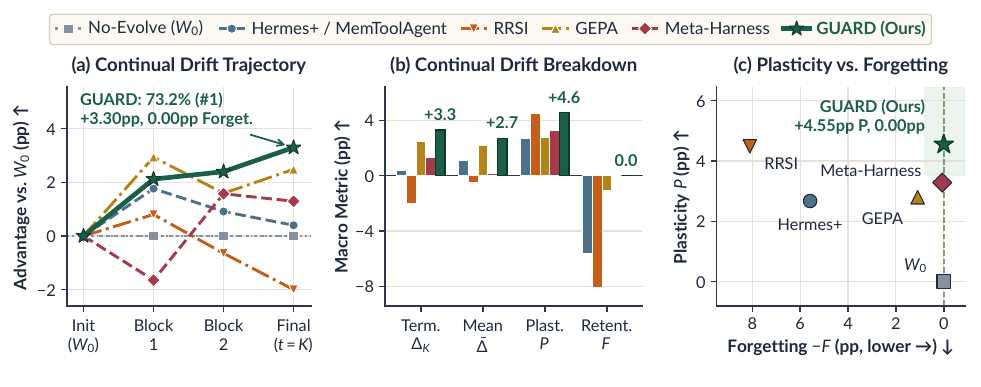}
\vspace{-6pt}
\caption{\textbf{\textsc{Drift Gauntlet} \cite{chen2026continual} continual RSI under single-model \texttt{Gemini 4} (four drift streams).} (a) Stage-normalized macro gain trajectory across continual checkpoints ($t{=}0$, Block~1, Block~2, Final): \texttt{GUARD} is the sole method improving monotonically ($0.00 \to +2.12 \to +2.40 \to +3.30\text{pp}$). (b) Macro continual metrics breakdown ($\Delta_K$, $\bar{\Delta}$, plasticity $P$, post-block retention $F$, where $0.00\text{pp}$ denotes zero forgetting). (c) Plasticity $P$ vs.\ post-block forgetting loss $-F$: \texttt{GUARD} lies above the baseline plasticity--stability trend, combining top plasticity ($+4.55\text{pp}$) with $0.00\text{pp}$ post-block forgetting.}
\label{fig:drift_gauntlet}
\vspace{-6pt}
\end{figure*}

\paragraph{Sustained multi-block progression and the plasticity--stability trade-off.}
While \textsc{EvoHarnessBench} shifts tool schemas within one domain at a time, \textsc{Drift Gauntlet} \cite{chen2026continual} stress-tests whether self-evolving harnesses can accumulate capabilities across unannounced workflow, topic, domain, and compound shifts without boundary resets. Table~\ref{tab:drift_gauntlet_main} and Figure~\ref{fig:drift_gauntlet} (per-group trajectories in Figure~\ref{fig:drift_gauntlet_group_dynamics}) show two main results under single-model \texttt{Gemini 4}. \texttt{GUARD} is the only arm whose stream-mean accuracy never decreases: across all $12$ continual block transitions, \texttt{GUARD} incurs $0/12$ stream-mean drops ($65.00 \to 70.00 \to 70.00\%$ on \emph{workflow}; $67.78 \to 67.78 \to 68.89 \to 70.00\%$ on \emph{topic}; $76.05 \to 78.69 \to 78.69 \to 78.69\%$ on \emph{domain}; $70.84 \to 71.67 \to 71.67 \to 74.17 \to 74.17\%$ on \emph{compound}). Its stage-normalized macro gain rises at every checkpoint ($0.00 \to +2.12 \to +2.40 \to +3.30\text{pp}$), ranking first on Final Mean ($73.22\%$, $0.82\text{pp}$ above \textsc{GEPA}, about three task-equivalents of $360$ held-out rollouts), $\Delta_K$ ($+3.30\text{pp}$, positive on all $4/4$ streams), and $\bar{\Delta}$ ($+2.71\text{pp}$). Every evolving baseline suffers $3$--$8$ block drops out of $12$: \textsc{Hermes+} peaks at Block~1 before decaying ($+1.76 \to +0.91 \to +0.40\text{pp}$), \textsc{RRSI} falls below $W_0$ ($+0.80 \to -0.65 \to -2.00\text{pp}$), \textsc{GEPA} dips $-1.35\text{pp}$ at Block~2 ($+2.94 \to +1.59 \to +2.48\text{pp}$), and \textsc{Meta-Harness} drops $-1.65\text{pp}$ below $W_0$ at Block~1 ($-20.00\text{pp}$ pre-block interference on \emph{workflow drift}). Consistent with \cite{chen2026continual}, plasticity $P$ and stability $S$ are strongly negatively correlated across the $16$ $(\text{baseline}, \text{stream})$ runs (Pearson $r = -0.82$, reported descriptively because $P$ and $S$ share checkpoint cells): high-plasticity \textsc{RRSI} ($P = +4.49\text{pp}$) suffers severe post-block forgetting ($F = -8.11\text{pp}$, $-20.00\text{pp}$ on opposing \texttt{BFCL-Miss} $\to$ \texttt{BFCL-Base}), whereas \textsc{GEPA}'s monolithic $4{,}200$-char prompt rewrite compresses prior-block character share on \emph{compound drift} (\texttt{BIRD-Code} $100\% \to 28.5\%$; \texttt{BFCL-Miss} $49.4\% \to 22.4\%$ once \texttt{BFCL-Base} arrives, dropping \texttt{BFCL-Miss} accuracy $73.33\% \to 66.67\%$). By keying sections to pre-execution observable tool/schema signatures ($0$ prior-signature characters overwritten) and gating updates against same-signature Vitter reservoir replay tasks (promoting $9/12$ blocks, including in-round \texttt{[iff]} repair on \texttt{ScienceWorld} to reach $82.73\%$, and rejecting $3/12$ opposing-drift overwrites where hurt equaled lift), \texttt{GUARD} achieves both top plasticity ($+4.55\text{pp}$) and $0.00\text{pp}$ post-block forgetting across all four streams, with only mild pre-block caution before opposing tool conventions are observed. On \emph{workflow} and \emph{compound} drift, where opposing blocks share the \texttt{bfcl} signature, Soft-HURT directly rejects both regressive overwrites (worked example in Box~\ref{box:drift_example}). In a symmetric routing control (Appendix~\ref{app:instantiation}), granting the same pre-execution signature routing to \textsc{GEPA} ($71.55\%$, $F = -2.78\text{pp}$) and \textsc{Meta-Harness} ($71.08\%$) leaves \texttt{GUARD} ahead across all four streams ($73.22\%$), while a single-global-playbook \texttt{GUARD} ablation without signature routing still achieves $72.35\%$ Final Mean, the highest global trajectory gain ($\bar{\Delta} = +2.25\text{pp}$), and zero post-block forgetting ($F = 0.00\text{pp}$).

\subsection{Self-Reviewer Calibration, Gate Fidelity, and Component Attribution}
\label{sec:calibration_ablation}

\begin{table}[!htbp]
\centering
\scriptsize
\setlength{\tabcolsep}{2.6pt}
\caption{\textbf{Summary of offline self-reviewer calibration and Soft-HURT gate fidelity} (zero oracle labels exposed during adaptation; NPV = negative predictive value; full 9-row audit in Table~\ref{tab:calibration_gate_fidelity_full}).}
\label{tab:calibration_gate_fidelity}
\begin{tabularx}{\columnwidth}{@{}>{\raggedright\arraybackslash}Xrr@{}}
\toprule
\textbf{Calibration / Gate Check (Sample $N$)} & \textbf{Ref.\ / 1-L} & \textbf{\texttt{GUARD}} \\
\midrule
Initial vs.\ final hosted ACC / Solved ($43$) & $75.98 / 48.84\%$ & $\mathbf{80.80 / 58.14\%}$ \\
Official ACC: L1-fail vs.\ L1-pass ($4{,}059$ test) & $45.97\%$ & $\mathbf{73.90\%}$ ($82.6\%$ NPV) \\
\textsc{PAST} playbook rank concord.\ ($169$ pairs) & $50.0\%$ & $\mathbf{82.2\%}$ ($75.6\%$ strict) \\
\midrule
Exec.-gated promotions $\Delta_{\text{test}} \ge 0$$^{\dagger}$ & --- & $\mathbf{19/19}$ ($15$ ${>}0$, $4$ ties) \\
Same gate w/o \texttt{Analyze} ($16$ local stages) & $4$ prom., $2/5$ $\downarrow$ & $\mathbf{4}$ prom., $\mathbf{0/5}$ $\downarrow$ \\
Bypassing Soft-HURT on \texttt{Calendar v3} & $-6.98\text{pp}$ Solv. & $\mathbf{0.00\text{pp}}$ (rejected) \\
\bottomrule
\end{tabularx}
\vspace{1pt}
\begin{minipage}{\columnwidth}
{\scriptsize $^{\dagger}$$\Delta_{\text{test}}$: held-out score of a promoted harness minus $W_0$ on the active cohort/group of the promoted stage/block (at $H_{v_3}$ on hosted \texttt{Calendar}, where $W_3{=}W_1$; \texttt{CSM v2}'s delta over \texttt{CSM v1} is not separately tabulated). Execution-gated suites give $19/19$ $\Delta_{\text{test}} \ge 0$ ($15/19$ ${>}0$): $9/9$ \textsc{Drift Gauntlet} blocks ($6$ ${>}0$ [$+3.33$ to $+16.67\text{pp}$], $3$ ties; Table~\ref{tab:drift_gauntlet_full}), $5/5$ \textsc{EvoHarnessBench} stages (hosted \texttt{v1} $+8.01$, \texttt{CSM v1/v2} $+4.17$/$+3.84$, \texttt{HR v1} $+1.90\text{pp}$, \texttt{ITSM v1} $0.00\text{pp}$ tie), and $5/5$ prof.\ role promotions (\textsc{PRBench} \texttt{R1 Finance/Legal} $+28.57$/$+5.77\text{pp}$; \textsc{Fin-Rate} \texttt{R1--R3} $+6.25$/$+12.50$/$+12.50\text{pp}$). Critic-scored \textsc{PAST-Bench} ($24/24$ admitted) lifts 12-family Task $0.372 \to 0.854$ (all $12/12 > W_0$).\par}
\end{minipage}
\vspace{-6pt}
\end{table}

Table~\ref{tab:calibration_gate_fidelity} (full audit across $4{,}173$ \textsc{EvoHarnessBench}/\textsc{PAST-Bench} rollouts plus \textsc{Drift Gauntlet} block decisions in Table~\ref{tab:calibration_gate_fidelity_full}) checks whether $S(\tau)$ and Soft-HURT track hidden verifiers. Layer-1 sandbox checks curb LLM-only optimism: on hosted \textsc{EvoHarnessBench}, a 1-layer reviewer passes ($\ge 75\%$) $22.2\%$ ($12/54$) of execution-broken traces vs.\ $0.0\%$ ($0/23$, sanity check by Eq.~\eqref{eq:self_review_scores}) under 2-layer review. The final hosted configuration (two-layer review, cumulative-source gating, and a completeness-first distiller instruction; Appendix~\ref{app:instantiation}) improves on the initial one-layer run by $+4.82\text{pp}$ ACC ($75.98\% \to 80.80\%$) and $+9.30\text{pp}$ Solved ($48.84\% \to 58.14\%$), with budget exhaustions dropping from $2/43$ to $0/43$. Across $4{,}059$ \textsc{EvoHarnessBench} test rollouts ($1{,}920$ hosted, $2{,}139$ local; all arms), Layer~1 separates official ACC by $+27.92\text{pp}$ ($73.90\%$ vs.\ $45.97\%$; $82.6\%$ NPV); on \textsc{PAST-Bench} ($N{=}114$), Layer~2 has near-zero mean bias ($-0.012$, $\text{MAE}=0.131$, $r{=}0.369$) and $82.2\%$ within-family concordance ($75.6\%$ non-tied, $r{=}0.432$). Under active-cohort $\Delta_{\text{test}}$ vs.\ $W_0$, all $19/19$ execution-gated Soft-HURT promotions across \textsc{Drift Gauntlet} ($9/9$), \textsc{EvoHarnessBench} ($5/5$), and the professional suites ($5/5$) satisfy $\Delta_{\text{test}} \ge 0$ ($15/19$ strictly ${>}0$ and $4/19$ $0.00\text{pp}$ ties; Table~\ref{tab:calibration_gate_fidelity} note), whereas bypassing Soft-HURT on a rejected local \texttt{Calendar v3} candidate lowers held-out accuracy and solve rate. In the five-domain teacher--student \texttt{Analyze}/\texttt{Gate} component ablation ($236$ tasks; Table~\ref{tab:evoharness_seam_ablation}), removing \texttt{Analyze} ($1 - \mathrm{obey}$, no pinning or contrasts) drops macro ACC by $-7.33\text{pp}$ ($66.46\% \to 59.13\%$, promoting $4$ stages that regress multi-stage skill domains), and removing Soft-HURT drops macro ACC by $-10.35\text{pp}$ ($66.46\% \to 56.11\%$, regressing $3/5$ domains vs.\ $0/5$). Under single-model \texttt{Gemini 4} on dual-axis \textsc{EvoHarnessBench} \texttt{ITSM} ($N{=}132$ test tasks, $N_{\mathrm{rep}}{=}3$; Table~\ref{tab:propose_seam_ablation}), parallel $M{=}2$ candidate pooling ($81.08\%$ ACC; BWT $+0.61\%$) beats matched-budget sequential $M{=}1, 2T$ search ($78.85\%$, $+2.23\text{pp}$, $p{=}0.0267$; $64\%$ fewer rejected gate rounds) and single-candidate $M{=}1, T$ ($77.20\%$, $+3.88\text{pp}$, $p{=}0.0029$).

\section{Conclusion}
\label{sec:conclusion}
Continually adapting an agent's harness around a frozen backbone model without external graders requires preventing local overfitting, catastrophic forgetting, and memory pollution. Drawing on curriculum regret from UED and group-relative scoring from reinforcement learning (GRPO), \textbf{GUARD} (\emph{Grader-free UED with Adaptive Rule Distillation}) treats the text harness as a non-parametric policy component and updates it from task-derived checklists: \texttt{Analyze} separates edit-induced regressions ($\Delta_{\mathrm{hurt}}$) from unresolved gaps ($\Delta_{\mathrm{gap}}$) while pinning mastered criteria, \texttt{Propose} selects among parallel candidate edits via critic-free group-standardized fitness and contrasts the best and worst candidates to rewrite proposal prompts, and \texttt{Gate} (\emph{Soft-HURT}) promotes only net-positive updates with formal non-regression guarantees on the self-review objective (Theorems~\ref{thm:gate}--\ref{thm:targeting}, Proposition~\ref{prop:dominance}). At test time, it deploys a single frozen harness in one pass. Across four benchmark settings, \texttt{GUARD} improves upon or matches $W_0$ in held-out accuracy, achieves $19/19$ non-regressive execution-gated promotions, and sustains monotonic \texttt{pass@1} gains with zero post-block forgetting under hidden stream drift while cutting hosted tokens per task by $54.5\%$.

\FloatBarrier
\bibliographystyle{plain}

\clearpage
\onecolumn

\appendix
\counterwithin{table}{section}
\counterwithin{figure}{section}
\renewcommand{\thetable}{\Alph{section}.\arabic{table}}
\renewcommand{\thefigure}{\Alph{section}.\arabic{figure}}
\renewcommand{\theHtable}{\Alph{section}.\arabic{table}}
\renewcommand{\theHfigure}{\Alph{section}.\arabic{figure}}
\setcounter{table}{0}
\setcounter{figure}{0}
\begin{center}
{\LARGE \textbf{Supplementary Material \& Technical Appendix}}\\[0.5em]
{\Large \textbf{GUARD: Stabilizing Recursive Self-Improvement with Regret Guidance}}\\[1.0em]
\ifanon
  {\large Anonymous Authors}\\[0.4em]
  {\normalsize Paper under double-blind review}
\else
  {\large Minghe Zhang$^{*}$ \quad Xuetao Kang$^{*}$ \quad Yifei Ming \quad Yanfei Chen \quad Han Yu}\\[0.25em]
  {\large Yufan Zhuang \quad Chen-Yu Lee}\\[0.4em]
  {\normalsize Google Cloud}\\[0.2em]
  {\footnotesize \texttt{\char`\{minghezhang,xuetaok,yifeim,yanfeichen,hanyyu,yufanzhuang,chenyulee\char`\}@google.com}}\\[0.2em]
  {\footnotesize $^{*}$Equal contribution.}
\fi
\end{center}
\vspace{1.0em}
\hrule
\vspace{1.5em}

\section{Proofs}
\label{app:proofs}

\subsection{Proof of Theorem~\ref{thm:gate} (Non-Regressive Gating and Concentration)}
\label{app:proof_gate}
\begin{proof}
\textit{Part (i).} If round $t$ promotes under the parameterized Soft-HURT gate \eqref{eq:soft_hurt_gate}--\eqref{eq:soft_hurt_param}, whether directly ($W_t = \widetilde{W}_t$) or after in-round \texttt{[iff]} repair ($W_t = \widetilde{W}^{\mathrm{rep}}_t$ on its re-scored trajectories), the promoted playbook satisfies
\begin{equation*}
\Delta_t \;>\; \mathrm{Hurt}(W_t, W_{t-1}) \;\ge\; 0,
\end{equation*}
where $\mathrm{Hurt}(W_t, W_{t-1}) = 0$ whenever $\mathcal{Q}^{-}_t = \emptyset$. If round $t$ rejects, $W_t = W_{t-1}$ and $\Delta_t = 0$. The incumbent score used by the gate is the score recorded when the incumbent was itself evaluated, so the increments telescope and give \eqref{eq:thm_gate_mono} with non-negative summands. A band of width $\delta_{\mathrm{band}} > 0$ relative to the incumbent instead admits any candidate with $\Delta_t \ge -\delta_{\mathrm{band}}$, so $T$ consecutive admissions at $\Delta_t = -\delta_{\mathrm{band}}$ accumulate a loss of $T\delta_{\mathrm{band}}$; a band relative to the best score so far bounds the aggregate loss by $\delta_{\mathrm{band}}$ but leaves per-task regressions unbounded.

\textit{Part (ii).} By \eqref{eq:hurt_penalty},
\begin{equation*}
|\mathcal{Q}_g| \cdot \mathrm{Hurt}(W_t,W_{t-1}) \;=\; \sum_{q \in \mathcal{Q}^{-}_t} \big(S_q(W_{t-1}) - S_q(W_t)\big),
\end{equation*}
and promotion requires $\mathrm{Hurt} < \Delta_t$, which is \eqref{eq:thm_gate_paid}; whenever $\mathcal{Q}^{-}_t = \emptyset$ (in particular when $h_{\max} = 0$), the left-hand side vanishes.

\textit{Part (iii).} Conditional on everything observed before a candidate harness evaluation, its $|\mathcal{Q}_g| N_{\mathrm{rep}}$ errors $\{S(\tau_{q,r}) - \mu_q(W)\}_{q \in \mathcal{Q}_g,\, r \le N_{\mathrm{rep}}}$ are independent across tasks and replicates and $\sigma$-sub-Gaussian, so $\bar{S}(W) - \mu(W)$ is $\sigma/\sqrt{|\mathcal{Q}_g| N_{\mathrm{rep}}}$-sub-Gaussian and
\begin{equation*}
\Pr\big(|\bar{S}(W) - \mu(W)| > \xi\big) \;\le\; 2\exp\!\left(-\frac{|\mathcal{Q}_g| N_{\mathrm{rep}}\, \xi^2}{2\sigma^2}\right).
\end{equation*}
Over $T_{\max}$ rounds, each group evaluates $W_0$, $M$ pool candidates per round, and at most $\mathbb{I}_{\mathrm{rep}} \in \{0,1\}$ repaired candidate per round, giving at most $n_{\mathrm{eval}} = 1 + T_{\max}(M + \mathbb{I}_{\mathrm{rep}})$ harness evaluations per group and $n_{\mathrm{eval}}|\mathcal{G}|$ across all groups. Choosing $\xi = \sigma\sqrt{2\log(2n_{\mathrm{eval}}|\mathcal{G}|/\alpha)/(|\mathcal{Q}_g| N_{\mathrm{rep}})}$ makes each tail probability at most $\alpha/(n_{\mathrm{eval}}|\mathcal{G}|)$. A union bound over the at most $n_{\mathrm{eval}}|\mathcal{G}|$ evaluations gives $|\bar{S} - \mu| \le \xi$ for all of them with probability at least $1 - \alpha$; the union bound remains valid although later candidates and repairs depend adaptively on earlier noise, because each tail bound holds conditionally on the past. On this event,
\begin{equation*}
\mu(W_T) \;\ge\; \bar{S}(W_T) - \xi \;=\; \bar{S}(W_0) + \sum_{t=1}^{T} \Delta_t - \xi \;\ge\; \mu(W_0) + \sum_{t=1}^{T}\Delta_t - 2\xi,
\end{equation*}
which is \eqref{eq:thm_gate_conc}.

\textit{Part (iv).} With per-group gating, group $g$ advances in a round with probability $a_g$, so the expected fraction of advancing groups is $\frac{1}{|\mathcal{G}|}\sum_g a_g$. A single global gate advances only if every group's candidate passes, which under independence has probability $\prod_g a_g$. For equal $a_g = a$ the ratio is $a / a^{|\mathcal{G}|} = a^{1-|\mathcal{G}|}$ (e.g., $8\times$ for $|\mathcal{G}| = 4$ and $a = 0.5$).
\end{proof}

\subsection{Proof of Theorem~\ref{thm:targeting} (Targeted Repair vs.\ Passive Discovery)}
\label{app:proof_targeting}
\begin{proof}
\textit{Part (i).} The passive learner must observe every invariant, so $T_{\mathrm{passive}} \ge \max_j G_j$, where $G_j \sim \mathrm{Geom}(p_j)$, supported on $\{1, 2, \dots\}$, is the first round in which invariant $j$ surfaces and the $G_j$ are independent. Because $p_j \le \epsilon < 1$, each $G_j$ stochastically dominates $G'_j \sim \mathrm{Geom}(\epsilon)$, and independence carries the dominance to the maxima. Couple $G'_j = \lceil E_j \rceil \ge E_j$ with $E_1, \dots, E_J$ i.i.d.\ exponential of rate $\nu = -\log(1-\epsilon)$; the coupling is exact since
\begin{equation*}
\Pr\big(\lceil E_j \rceil > \ell\big) = e^{-\nu \ell} = (1-\epsilon)^{\ell} .
\end{equation*}
By the R\'enyi representation of exponential order statistics, $\max_j E_j \stackrel{d}{=} \sum_{\ell=1}^{J} X_\ell/(\ell\nu)$ with $X_\ell$ i.i.d.\ $\mathrm{Exp}(1)$, so for all $\epsilon \in (0, 1)$,
\begin{equation*}
\mathbb{E}[T_{\mathrm{passive}}] \;\ge\; \mathbb{E}\big[\max_j E_j\big] \;=\; \frac{H_J}{-\log(1-\epsilon)} \;\ge\; \frac{1-\epsilon}{\epsilon}\,H_J,
\end{equation*}
using $-\log(1-\epsilon) \le \epsilon/(1-\epsilon)$ on $(0, 1)$ (giving $\Omega((\log J)/\epsilon)$ when $\epsilon \le 1/2$). For example, $J = 10$ and $\epsilon = 0.05$ give $\mathbb{E}[T_{\mathrm{passive}}] \ge 57.1$ rounds.

\textit{Part (ii).} Let $u_t$ be the number of invariants still unresolved after round $t$ ($u_0 = J$), and let $\mathcal{F}_t$ be the history of the first $t$ rounds. By (a) and (d), resolved invariants never displace unresolved ones in the ranking and stay resolved.

\emph{Phase 1} ($u_{t-1} \ge b_{\min}$). By (b) and (c), on the event $\{t_1 > t - 1\} \in \mathcal{F}_{t-1}$ where $t_1 = \min\{t : u_t < b_{\min}\}$, at least $b_{\min}$ unresolved invariants are targeted and
\begin{equation*}
\mathbb{E}\big[u_{t-1} - u_t \,\big|\, \mathcal{F}_{t-1}\big] \;\ge\; \rho\, b_{\min}.
\end{equation*}
The stopping time $t_1$ is integrable because each active round resolves at least one invariant with probability at least $\rho$. Defining $Z_t = (J - u_t) - \rho\, b_{\min} t$, the stopped process $Z_{t \wedge t_1}$ is a bounded-increment submartingale; optional stopping yields
\begin{equation*}
\rho\, b_{\min}\, \mathbb{E}[t_1] \;\le\; J - \mathbb{E}[u_{t_1}] \;\le\; J .
\end{equation*}

\emph{Phase 2} ($u_{t_1} < b_{\min}$). Conditional on $\mathcal{F}_{t_1}$, if $u_{t_1} = 0$ (in particular whenever $b_{\min} = 1$), zero further rounds are needed; if $1 \le u_{t_1} \le b_{\min} - 1$, every remaining invariant is targeted in every subsequent round and resolved independently with probability at least $\rho$, so the conditional remaining time is stochastically dominated by the maximum of $u_{t_1} \le b_{\min} - 1$ independent $\mathrm{Geom}(\rho)$ variables. With the coupling of Part (i) at rate $\nu' = -\log(1-\rho) \ge \rho$, this maximum is at most $1 + \max_j E'_j$, whose conditional expectation is at most $1 + H_{b_{\min}-1}/\nu' \le 1 + H_{b_{\min}-1}/\rho$.

Taking unconditional expectations and adding the two phases gives \eqref{eq:thm_target}, which does not involve $\epsilon$. (Under all-or-nothing bundle admission with joint pass probability $\rho_{\mathrm{bundle}} > 0$ per round, each accepted round resolves $\min(b_{\min}, u_{t-1})$ invariants, requiring at most $\lceil J / b_{\min} \rceil$ accepted bundles of mean waiting time $1/\rho_{\mathrm{bundle}}$ each.)
\end{proof}

\subsection{Proof of Proposition~\ref{prop:dominance} (Strict Regression Dominance under Multi-Rollout Grading)}
\label{app:proof_dominance}
\begin{proof}
When round $t-1$ promoted, \eqref{eq:paired_split} gives $\Delta_{\mathrm{gap},t}(c) = 1 - o_c$ and $\Delta_{\mathrm{hurt},t}(c) = \max(0, \delta_c)$ for any numbers of rollouts graded under $W_{t-2}$ and $W_{t-1}$, so $\mathrm{Priority}_t(c) = (1 - o_c) + \max(0, \delta_c)$. For a regressed criterion $c$ ($\delta_c > 0$) and a non-regressed criterion $c'$ ($\delta_{c'} \le 0$, so $\Delta_{\mathrm{hurt},t}(c') = 0$),
\begin{equation*}
\mathrm{Priority}_t(c) - \mathrm{Priority}_t(c') \;=\; \big(1 - o_c + \delta_c\big) - \big(1 - o_{c'}\big) \;=\; \delta_c - (o_c - o_{c'}).
\end{equation*}

\textit{Part (i).} If $o_{c'} \ge o_c$, the right-hand side is at least $\delta_c > 0$.

\textit{Part (ii).} The right-hand side is positive if and only if $o_c - o_{c'} < \delta_c$. In the single-rollout binary case ($K_{\mathrm{gr}} = 1$), a fresh regression has $o_c = 0$ and $\delta_c = 1$, so $\mathrm{Priority}_t(c) = 2$, whereas every non-regressed criterion has $\mathrm{Priority}_t(c') = 1 - o_{c'} \le 1$.

\textit{Part (iii).} In any round (promoted or rejected), both $\Delta_{\mathrm{hurt},t}(c)$ and $\Delta_{\mathrm{gap},t}(c) = 1 - \mathrm{obey}(c; W_{t-1})$ in \eqref{eq:paired_split} are non-negative, so $\mathrm{Priority}_t(c) = 0$ holds if and only if $\mathrm{obey}(c; W_{t-1}) = 1$ and $\Delta_{\mathrm{hurt},t}(c) = 0$. Because each verdict $r(\tau, c)$ is binary, $\mathrm{obey}(c; W_{t-1}) = 1$ holds exactly when every trajectory in $\mathcal{T}_c(W_{t-1})$ obeys $c$; hence no criterion failed by $W_{t-1}$ and no regressed criterion ($\Delta_{\mathrm{hurt},t}(c) > 0$) is ever pinned.
\end{proof}

\section{Harness Instantiation Used in Our Experiments}
\label{app:instantiation}
This appendix lists the concrete constants and baseline details for \S\ref{sec:generalized_rsi} and \S\ref{sec:methodology}--\S\ref{sec:experiments}.
\begin{itemize}[leftmargin=*,noitemsep,topsep=2pt]
    \item \textbf{Soft-HURT parameters, pool reward, and ledger window:} Single-task noise floor $\delta_{\mathrm{noise}} = 15$ points on $[0,100]$ (fixed across all suites without per-domain tuning); default trade-off budget $h_{\max} = 1$ hurt task and margin $\mu_{\min} = 1.5$ points (on the professional suites, $h_{\max} = \mathrm{round}(0.18\,|\mathcal{Q}_g|)$, $\mu_{\min} = 3.0$, and at least two helped tasks; on multi-stage continual tool/skill and drift suites, un-repaired candidates use conservative threshold $h_{\max} = 0$ and trigger in-round \texttt{[iff]} repair up to $h_{\mathrm{rep}} \in \{1,2,3\}$ with re-executed Soft-HURT evaluation at $h_{\max} = 1, \mu_{\min} = 0$); active deduplicated rejection-ledger read-back window $L_{\max} = 12$ entries per group on hosted \textsc{EvoHarnessBench}, $L_{\max} = 40$ in the reference candidate-level runner (\textsc{Drift Gauntlet}, local \textsc{EvoHarnessBench}, and \textsc{PAST-Bench}), and all active deduplicated criterion-removal entries in the professional-suite runner; $\mathrm{Perf}_m = \bar{S}^{\mathrm{strict}}(\widetilde{W}^{(m)}_t) + \tfrac14\, \bar{S}(\widetilde{W}^{(m)}_t)$ (estimated by the critic under critic scoring), $\beta = 3.5$, $\gamma = 1.5$, stabilizer $10^{-6}$, contrast threshold $\delta_{\mathrm{con}} = 10$ points; pool size $M = 2$ across all main-text, local-gym, and \textsc{Fin-Rate \& Ledger} runs and $M = 4$ on \textsc{PRBench-Legal \& Harvey LAB} (reaching $66.67\%$ Official; an $M = 2$ prior-free run reaches $65.76\%$, and an exploratory $M = 4$ hosted run reaches $76.42\%$), initialized from $s_0$ across all $M$ parallel candidates in Round~1 and evolved via $m^{+}$ vs.\ $m^{-}$ ($m^{-} \ne m^{+}$) in rounds $t \ge 2$ (Appendix~\ref{app:seed_prompt}).
    \item \textbf{Suite-level Soft-HURT validation buffers:} Hosted and local \textsc{EvoHarnessBench} gate candidates on cumulative visible stage tasks ($5/5$ stage promotions admitted cleanly under Soft-HURT: $1$ hosted, $4$ local); at each new stage, the local runner also re-checks the carried incumbent against $W_0$ with the same Soft-HURT rule on the expanded training split and returns to $W_0$ if it fails, which happened once (\texttt{HR}, Stage~\texttt{v2}). \textsc{Drift Gauntlet} gates against a $24$-task stratified validation buffer combining current-block UED frontier tasks with up to $10$ Vitter Algorithm~R reservoir replay samples sharing the pre-execution observable tool/schema signature \texttt{observable\_domain\_sig} (promoting $9/12$ blocks, including in-round \texttt{[iff]} repair on \texttt{ScienceWorld} under \emph{domain drift}, and rejecting $3/12$ blocks where hurt equaled lift). \textsc{PAST-Bench} applies Soft-HURT ($\delta_{\mathrm{noise}} = 15, h_{\max} = 1, \mu_{\min} = 1.5$) to per-task scores from the joint critic pass ($24/24$ round winners admitted across $12$ families; \S\ref{sec:experiments}). The two professional suites gate strictly on \texttt{TRAIN} pre-execution checklist self-reviews ($\delta_{\mathrm{noise}} = 15$, in-round \texttt{[iff]} repair; \textsc{PRBench-Legal} gates $m^{\star} = \arg\max_m \hat{A}_m$ as in Algorithm~\ref{alg:unified_loop}, while \textsc{Fin-Rate \& Ledger} evaluates across the candidate pool), while post-hoc \texttt{DEV} uses the sealed official rubric (Table~\ref{tab:ued_map}).
    \item \textbf{Rounds and replicates:} Although the reference runner supports early stopping ($<0.5$-point macro gain with $<2$ net-helped tasks, no group promoted, or $T_{\max} = 5$), all reported experiments fix the round count: 1 per teach/update episode on \textsc{PAST-Bench}; 1 per stage $\texttt{adapt}(v_k)$ on \textsc{EvoHarnessBench} (\texttt{v1}--\texttt{v3}, and \texttt{v1}--\texttt{v4} on \texttt{ITSM}); 1 per online task block $k \in \{1,\dots,K\}$ on \textsc{Drift Gauntlet} ($K \in \{2,3,4\}$); and 3 adaptation rounds (\texttt{R1}--\texttt{R3} from baseline $\texttt{R0} = W_0$) on the two professional suites. Main-text and local-gym suites use $N_{\mathrm{rep}} = 1$ and the two professional suites use $N_{\mathrm{rep}} = 2$ evaluation replicates per task.
    \item \textbf{Benchmark rollout notes:} Under unified \texttt{Gemini 4} ($8$-step agent budget/task on hosted \textsc{EvoHarnessBench}), Layer~1 verifies a final answer and required state mutations while Layer~2 grades task-derived criteria on cumulative stage tasks ($12$, $16$, and $18$ at \texttt{v1}--\texttt{v3}), and an identical single-re-execution pass re-ran relay/tunnel infrastructure outages across \emph{every} arm under the unchanged harness state before aggregating Table~\ref{tab:evoharness_main}, leaving $0$ errored records in all six final 43-task rows ($6$ infra-errored records occurred only in intermediate cell $(\text{Cohort }\texttt{v1}, H_{v_2})$, which enters no reported column). All five hosted \texttt{GUARD} configurations evaluated in our study (the reported $M{=}2$ main arm at $80.80\%$ in Table~\ref{tab:evoharness_main}, its original-distiller control at $75.57\%$, the initial one-layer run at $75.98\%$ in Table~\ref{tab:calibration_gate_fidelity}, a fixed-persona $M{=}2$ variant at $76.69\%$, and an $M{=}4$ operator pool at $76.42\%$) beat the top external baseline ($74.52\%$, \textsc{Meta-Harness}). On \textsc{Drift Gauntlet} \cite{chen2026continual} (single-model \texttt{Gemini 4}, seed $42$), every task exposes a pre-execution observable environment/schema signature (\texttt{bfcl}, \texttt{sciworld}, or \texttt{bird:<db\_id>}) without any hidden block index or group label; each signature's section is bounded to $3{,}800$ characters and placed at both the start and end of the system prompt. Because the three \textsc{BIRD} groups use different databases and \emph{domain drift} spans three environments, signatures separate the groups of \emph{topic} and \emph{domain} drift, whereas \texttt{BFCL-Miss} and \texttt{BFCL-Base} share \texttt{bfcl} and test within-signature opposing shifts via the $10$-sample replay buffer. Held-out evaluation draws one rollout per task (solver temperature $0.2$) and memoizes it by task and rendered playbook for every arm, avoiding the artificial $-0.59\text{pp}$ apparent forgetting induced when a constant $W_0$ is repeatedly re-sampled at temperature $0.2$ ($3$-pass $W_0$ mean $69.86\% \pm 0.18\text{pp}$). In a symmetric $2 \times 3$ routing control across all four streams, equipping \textsc{GEPA} and \textsc{Meta-Harness} with the same pre-execution signature routing lowers their macro means to $71.55\%$ ($F = -2.78\text{pp}$, with $-13.33\text{pp}$ forgetting on \emph{workflow drift}) and $71.08\%$, whereas running \texttt{GUARD} with a single global playbook (no signature routing) still reaches $72.35\%$ ($\Delta_K = +2.44\text{pp}$, $\bar{\Delta} = +2.25\text{pp}$, $F = 0.00\text{pp}$). This shows that \texttt{Analyze}, \texttt{Propose}, and Soft-HURT account for $74\%$ of \texttt{GUARD}'s total gain over $W_0$. In the lightweight local five-domain gym (Table~\ref{tab:evoharness_domains}), the semantic reviewer grades structured execution, entity-grounding, policy-completeness, and outcome-verification predicates directly with deterministic trace fallbacks.
\end{itemize}

\subsection{Comparison with Prior RSI Frameworks and Environments}
\label{app:related_table}
Table~\ref{tab:novelty_comparison} summarizes representative recursive self-improvement (RSI) methods and environments discussed in \S\ref{sec:related_work} across their optimization targets, closed-loop stability mechanisms, and adaptation and active-context scalability.

\begin{table}[!htbp]
\caption{\textbf{Optimization targets, closed-loop mechanisms, and scalability of recursive self-improvement (RSI) frameworks} (taxonomy adapted from RSIGym \cite{meng2026rsigym}). Checkmarks (\yesmark), crosses (\nomark), and \partialmark{} mark full presence, absence, or partial support; $^{\dagger}$ marks baselines evaluated here. \textbf{Scalability} pairs adaptation cost ($T$ rounds, $Q$ tasks, $M$ or $K$ candidates) with active context/archive growth in $T$ (\S\ref{sec:grpo_pool}, Appendix~\ref{app:complexity}).}
\label{tab:novelty_comparison}
\centering
\fontsize{7.0}{8.3}\selectfont
\setlength{\tabcolsep}{2.2pt}
\setlength{\aboverulesep}{0.14ex}
\setlength{\belowrulesep}{0.22ex}
\renewcommand{\arraystretch}{0.88}
\begin{tabularx}{\textwidth}{@{}>{\raggedright\arraybackslash}p{0.225\textwidth}cccccc>{\raggedright\arraybackslash}p{0.155\textwidth}>{\raggedright\arraybackslash}X@{}}
\toprule
\textbf{Framework / Environment} & \multicolumn{3}{c}{\textbf{Optimization Targets}} & \multicolumn{3}{c}{\textbf{Closed-Loop Features}} & \textbf{Scalability} & \textbf{System Design \& Update Mechanism} \\
\cmidrule(lr){2-4} \cmidrule(lr){5-7}
& \shortstack{Weight\\Updates} & \shortstack{Harness\\\& Skills} & \shortstack{Text\\Memory} & \shortstack{Regret\\Ranking} & \shortstack{Annealed\\Budget} & \shortstack{Non-Regr.\\Gate} & (Work / Depth; $T$-Ctx.) & \\
\midrule
Reflexion / ExpeL / AWM \cite{shinn2023reflexion,zhao2024expel,wang2024awm}
& \nomark & \nomark & \yesmark & \nomark & \nomark & \nomark & $O(TQ)$ / $O(TQ)$; $O(T)$ mem. & Episodic verbal buffers, insight extraction, \& workflow memory \\
Voyager / MemGPT / WikiSkill \cite{wang2023voyager,packer2023memgpt,tang2026wikiskill}
& \nomark & \yesmark & \yesmark & \nomark & \nomark & \nomark & $O(TQ)$ / $O(TQ)$; $O(T)$ lib. & Skill libraries \& append-only wikis; skill edits gated on validation score \\
Dynamic Cheatsheet / ACE / ReasoningBank \cite{suzgun2025dynamic,zhang2025ace,ouyang2026reasoningbank}
& \nomark & \nomark & \yesmark & \nomark & \nomark & \nomark & $O(TQ)$ / $O(TQ)$; $O(T)$ ctx. & Test-time cheatsheets, incremental context updates, \& trace memory \\
DSPy / TextGrad / EvoPrompt \cite{khattab2023dspy,yuksekgonul2024textgrad,guo2024evoprompt}
& \nomark & \nomark & \yesmark & \nomark & \nomark & \nomark & $O(TKQ)$ / $O(TQ)$; $O(1)$ ctx. & Scalar-objective, textual-gradient, or evolutionary prompt search \\
TextReg \cite{fu2026textreg}
& \nomark & \nomark & \yesmark & \nomark & \nomark & \nomark & $O(TQ)$ / $O(TQ)$; $O(T)$ bank & Regularized text updates filtering low-recurrence patches \& length (requires GT) \\
\textsc{GEPA}$^{\dagger}\textsuperscript{f}$ \cite{agrawal2026gepa}
& \nomark & \nomark & \yesmark & \nomark & \nomark & \nomark & $O(TKQ)$ / $O(TQ)$; $O(TKQ)$ front. & Reflective prompt evolution with per-instance Pareto archive (no \texttt{[iff]}) \\
\textsc{Hermes+} (\textsc{PAST-Bench})$^{\dagger}$ \cite{xue2026pastbench}
& \nomark & \yesmark & \yesmark & \nomark & \nomark & \nomark & $O(TQ)$ / $O(TQ)$; $O(T)$ mem. & Cross-session personal-agent memory \& skill persistence \\
\textsc{MemToolAgent} (\textsc{EvoHarnessBench})$^{\dagger}$ \cite{evoharnessbench2026}
& \nomark & \yesmark & \yesmark & \nomark & \nomark & \nomark & $O(TQ)$ / $O(TQ)$; $O(T)$ skills & Continual skill \& tool-memory accumulation across evolving stages \\
MLGym / PostTrainBench\textsuperscript{a} \cite{nathani2025mlgym,rank2026posttrainbench}
& \yesmark & \nomark & \nomark & \nomark & \nomark & \nomark & GPU train; $O(T)$ & Containerized ML workspaces \& autonomous post-training \\
Absolute Zero / R-Zero / Agent0\textsuperscript{d} \cite{zhao2025absolute,huang2026rzero,xia2026agent0}
& \yesmark & \nomark & \nomark & \partialmark & \nomark & \nomark & RL grads; $O(T)$ & Zero-data RL self-play via code-executor or majority-vote rewards \\
Context Language Models (CLM) \cite{shao2026clm}
& \yesmark & \nomark & \yesmark & \nomark & \nomark & \nomark & RL grads; $O(1)$ ctx. & RL-trained intra-episode context-file editing with efficiency gating \\
ADAS / AFlow / DGM\textsuperscript{e} \cite{hu2025adas,zhang2025aflow,zhang2026dgm}
& \nomark & \yesmark & \nomark & \nomark & \nomark & \nomark & $O(TKQ)$ / $O(TQ)$ GT; $O(T)$ & Archive / MCTS agent-code evolution scored on validation ground truth \\
AutoHarness\textsuperscript{b} / Self-Harness \cite{lou2026autoharness,zhang2026selfharness}
& \nomark & \yesmark & \nomark & \nomark & \nomark & \partialmark & $O(TKQ)$ / $O(TQ)$; $O(T)$ tests & Code-harness synthesis; regression-test gating in Self-Harness \\
\textsc{Meta-Harness}$^{\dagger}$ \cite{lee2026metaharness}
& \nomark & \yesmark & \yesmark & \nomark & \nomark & \nomark & $O(TKQ)$ / $O(TQ)$; $O(TQH)$ log & Agentic search over harness code \& memory using prior traces \\
RSIBench-Data / RSIGym\textsuperscript{c} \cite{meng2026rsibenchdata,meng2026rsigym}
& \yesmark & \yesmark & \nomark & \nomark & \nomark & \nomark & Shared-\$ budget & Everything-as-a-Service (EaaS) under a shared dollar budget \\
\textsc{RRSI}$^{\dagger}$ \cite{xia2026rrsi}
& \nomark & \yesmark & \yesmark & \nomark & \yesmark & \nomark & $O(TQ)$ / $O(TQ)$; $O(1)$ ctx. & Annealed edit budget, critic \& pruner, score-band gate (within $\delta_{\mathrm{band}}$ of the best score) \\
\textsc{Harness}\textsuperscript{2} \cite{xu2026harness2}
& \nomark & \yesmark & \yesmark & \nomark & \nomark & \nomark & $O(KQ)$ probe; $O(Q)$ lib. & Per-task online Propose--Probe--Compose from $W_0$ + continual improver library \\
\midrule
\rowcolor{LightBlue}
\textbf{\texttt{GUARD} (Ours)}\textsuperscript{g}
& \nomark & \yesmark & \yesmark & \yesmark & \nomark & \yesmark & \textbf{$O(TMQ)$ / $O(TQ)$; $O(1)$ ctx.} & \textbf{Non-parametric regret-guided RSI: $\Delta_{\mathrm{hurt},t}\!+\!\Delta_{\mathrm{gap},t}$ priority, \texttt{[iff]} guards, $m^{+}\!/\!m^{-}$ pool, Soft-HURT gate} \\
\bottomrule
\end{tabularx}
\vspace{2pt}
\begin{minipage}{\textwidth}
\fontsize{6.9}{8.2}\selectfont
\textsuperscript{a}Optimizes model weights/hyperparameters under fixed agent code. \textsuperscript{b}Synthesizes action-blocking code harnesses around frozen weights; Self-Harness \cite{zhang2026selfharness} gates on regression tests. \textsuperscript{c}Exposes training, rollout, evaluation, and sandboxes under a shared budget. \textsuperscript{d}Also Self-Challenging Agents \cite{zhou2025selfchallenging}; curriculum RL targets solver uncertainty or code-verified tasks instead of per-criterion $\Delta_{\mathrm{hurt}}$ across playbooks. \textsuperscript{e}Mutates Python agent workflows or coding-agent repositories (SWE-bench/Polyglot in DGM) against ground-truth validation scores. \textsuperscript{f}\textsc{GEPA} maintains an instance-Pareto archive without an incumbent non-regression promotion gate. \textsuperscript{g}\texttt{GUARD} updates zero model weights (though its abstract interface accommodates adapter/data engines); every reported run uses reference-free checklist self-review, paired criterion regret ranking $\Delta_{\mathrm{hurt},t} + \Delta_{\mathrm{gap},t}$, fixed un-annealed budgets, and Soft-HURT gating.
\end{minipage}
\vspace{-4pt}
\end{table}

\subsection{Baselines and Procedural Breakdown Across the Three Primary Benchmarks}
\label{app:baselines}
No arm reads held-out sessions, test splits, drift boundaries, or grader files (per-arm configurations and hosted playbook sizes in Table~\ref{tab:arm_configs}; procedural \textsc{PAST-Bench} breakdown in Table~\ref{tab:pastbench_procedural}). On \textsc{PAST-Bench} (all persistent arms on \textsc{Hermes+} with \texttt{gemini-3.8-flash} solver and \texttt{claude-opus-5-5} optimizer/judge), \textsc{RRSI} \cite{xia2026rrsi} reflects on visible trajectories and rewrites memory and skill files after each teach or update episode (omitting its validation-split score-band gate because families expose single-trajectory sessions without a validation split), \textsc{GEPA} \cite{agrawal2026gepa} evolves instructions via $K = 2$ reflective candidates with per-episode Pareto tracking, and \textsc{Meta-Harness} \cite{lee2026metaharness} proposes $K = 2$ candidate rewrites of both files from accumulated trajectories and keeps the higher self-scored state. To confirm that \texttt{GUARD}'s gains on \textsc{PAST-Bench} do not rely on a stronger teacher optimizer (\texttt{claude-opus-5-5} distilling for a \texttt{gemini-3.8-flash} solver), we also executed two complete 12-family, six-arm \emph{model-identical} suites in which solver, distiller, critic, and grader all use the exact same model: (1) under pure \texttt{claude-opus-5-5}, \texttt{GUARD} ($M{=}2$ prior-free) ranks first on overall Task ($0.870$ vs.\ $0.866$ \textsc{RRSI}, $0.861$ \textsc{Hermes+}, $0.861$ \textsc{Meta-Harness}, $0.858$ \textsc{GEPA}, and $0.347$ $W_0$; $+0.523$ lift over $W_0$), Recall ($0.871$), and Evolve ($0.458$ vs.\ $0.431$--$0.454$), and achieves the sole lowest memory pollution among all persistent arms ($0.167$ vs.\ $0.208$ on \textsc{Hermes+}/\textsc{GEPA} and $0.250$ on \textsc{RRSI}/\textsc{Meta-Harness}; $0.000$ procedural pollution); and (2) under pure \texttt{gemini-3.8-flash}, \texttt{GUARD} lifts Task from $0.371$ ($W_0$) to $0.852$ (with Mechanism $0.840$ vs.\ $0.829$--$0.835$ on \textsc{Meta-Harness}/\textsc{GEPA}/\textsc{RRSI}), ties for the lowest memory pollution ($0.167$ vs.\ $0.194$ on \textsc{RRSI}; $0.000$ procedural pollution), and leads multi-session latent rule induction (\texttt{PC03}: $0.748$ vs.\ $0.653$ \textsc{RRSI}, $0.645$ \textsc{Meta-Harness}, and $0.510$ \textsc{GEPA}; $+9.5$ to $+23.8\text{pp}$). On hosted \textsc{EvoHarnessBench} and \textsc{Drift Gauntlet} (both under single-model \texttt{Gemini 4} for every role), \textsc{RRSI} performs single-candidate recursive self-critic diagnosis (\texttt{Failure Mode Diagnosis}) and rewrites the harness at each stage or continual block without paired priority ranking or Soft-HURT gating, while \textsc{GEPA} ($K = 2$, $4{,}200$-char prompt cap on \textsc{Drift Gauntlet}) and \textsc{Meta-Harness} ($K = 2$) evaluate candidates on each stage/block's visible adaptation tasks.

\begin{table}[!htbp]
\centering
\scriptsize
\setlength{\tabcolsep}{3.5pt}
\caption{\textbf{Per-arm adaptation configuration across the unified six-arm suite.} All evolving arms share identical solver and optimizer/reviewer models within each benchmark (\S\ref{sec:experiments}). Final hosted chars report the Stage-\texttt{v3} playbook/memory length on hosted \textsc{EvoHarnessBench}.}
\label{tab:arm_configs}
\begin{tabularx}{\textwidth}{@{}>{\raggedright\arraybackslash}p{0.21\textwidth}c>{\raggedright\arraybackslash}p{0.15\textwidth}>{\raggedright\arraybackslash}p{0.19\textwidth}>{\raggedright\arraybackslash}Xr@{}}
\toprule
\textbf{Arm} & \textbf{Cands./Rd.} & \textbf{Adapt.\ Rollouts} & \textbf{Selection \& Gate Rule} & \textbf{Adaptation Feedback Signal} & \textbf{Hosted Chars} \\
\midrule
\texttt{No-Evolve} ($W_0$) & $0$ & $0$ & Static base harness & None (zero-shot ReAct / base harness) & $0$ \\
\textsc{Hermes+} / \textsc{MemToolAgent} & $1$ & $1\times$ online & Unconditional append/edit & Agent-initiated memory \& skill writes & $6{,}989$ \\
\textsc{RRSI} \cite{xia2026rrsi} & $1$ & $1\times$ online / stage & Unconditional rewrite (no val.\ split) & Self-critic \texttt{Failure Mode Diagnosis} + pruner & $10{,}013$ \\
\textsc{GEPA} \cite{agrawal2026gepa} & $K{=}2$ & $2\times$ (exec.) / $1\times$ (critic) & Per-instance Pareto frontier argmax & Reflective trace critique on non-dominated tasks & $5{,}123$ \\
\textsc{Meta-Harness} \cite{lee2026metaharness} & $K{=}2$ & $2\times$ (exec.) / $1\times$ (critic) & Best self-scored candidate (ungated) & Proposer--judge search over raw trace archive & $4{,}538$ \\
\midrule
\rowcolor{LightBlue} \textbf{\texttt{GUARD} (Ours)} & $\boldsymbol{M{=}2}$ & $\boldsymbol{2\times}$ \textbf{(exec.) /} $\boldsymbol{1\times}$ \textbf{(critic)} & $\boldsymbol{\arg\max \hat{A}_m}$ \textbf{+ Soft-HURT gate} & \textbf{2-layer checklist self-review (exec.\ or critic), $\Delta_{\mathrm{hurt}}\!+\!\Delta_{\mathrm{gap}}$, pinning, $m^{+}\!/\!m^{-}$ pool} & $\mathbf{2{,}315}$ \\
\bottomrule
\end{tabularx}
\end{table}

\begin{table}[!htbp]
\centering
\footnotesize
\setlength{\tabcolsep}{6.5pt}
\caption{\textbf{\textsc{PAST-Bench} procedural SOP sub-table across the three multi-step procedural families (\texttt{PC01}, \texttt{PC03}, \texttt{PC04}; Figure~\ref{fig:pastbench}(a)).} Bold marks the best persistent arm per column.}
\label{tab:pastbench_procedural}
\begin{tabular}{lcccc}
\toprule
\textbf{Arm} & \textbf{Proc.\ Task} & \textbf{Proc.\ Mech.} & \textbf{Proc.\ Upd.} & \textbf{Proc.\ Poll.$\downarrow$} \\
\midrule
\texttt{No-Evolve} ($W_0$) & 0.427 & 0.443 & 0.214 & 0.000 \\
\textsc{Hermes+} & 0.808 & 0.780 & 0.529 & \textbf{0.000} \\
\textsc{RRSI} \cite{xia2026rrsi} & 0.714 & 0.508 & 0.841 & 1.000 \\
\textsc{GEPA} \cite{agrawal2026gepa} & 0.815 & 0.761 & 0.464 & \textbf{0.000} \\
\textsc{Meta-Harness} \cite{lee2026metaharness} & 0.677 & 0.516 & 0.824 & 1.000 \\
\midrule
\rowcolor{LightBlue} \textbf{\texttt{GUARD} (Ours)} & \textbf{0.830} & \textbf{0.900} & \textbf{0.901} & \textbf{0.000} \\
\bottomrule
\end{tabular}
\end{table}

\begin{table}[!htbp]
\centering
\footnotesize
\setlength{\tabcolsep}{4.8pt}
\caption{\textbf{Paired per-family differences on \textsc{PAST-Bench}, \texttt{GUARD} minus each persistent baseline (12 families; better/tied/worse counts; mean differences computed from unrounded per-family scores before rounding to three decimals).} \texttt{GUARD} achieves positive mean gains on Mech., Upd., and W.Prec.\ against every baseline with zero W.Prec.\ or Poll.\ family losses; exact two-sided sign tests reach unadjusted $p < 0.05$ on Mech.\ ($p = 0.016$) and W.Prec.\ ($p = 0.031$) vs.\ \textsc{RRSI}, with none surviving Holm correction across all 18 comparisons.}
\label{tab:pastbench_paired}
\begin{tabular}{lccccc}
\toprule
Comparison & Task & Mech. & Upd. & W.Prec. & Poll.$\downarrow$ \\
\midrule
vs.\ \textsc{Hermes+} & $+0.002$ (4/5/3) & $+0.037$ (5/5/2) & $+0.133$ (6/5/1) & $+0.052$ (1/11/0) & $+0.000$ (0/12/0) \\
vs.\ \textsc{RRSI} & $+0.045$ (6/5/1) & $+0.153$ (7/5/0) & $+0.023$ (6/5/1) & $+0.339$ (6/6/0) & $-0.333$ (4/8/0) \\
vs.\ \textsc{GEPA} & $-0.004$ (1/6/5) & $+0.041$ (4/6/2) & $+0.148$ (4/6/2) & $+0.057$ (3/9/0) & $+0.000$ (0/12/0) \\
vs.\ \textsc{Meta-Harness} & $+0.042$ (5/5/2) & $+0.139$ (6/5/1) & $+0.021$ (5/5/2) & $+0.297$ (5/7/0) & $-0.333$ (4/8/0) \\
\bottomrule
\end{tabular}
\end{table}

\subsection{Reproducibility, Hyperparameters, and Prompt Templates}
\label{app:reproducibility}
\paragraph{Model identifiers, decoding parameters, and budgets.} Hosted \textsc{EvoHarnessBench} ($8$-step rollout cap, $B_W = 2{,}600$ characters), \textsc{Drift Gauntlet} ($B_W = 3{,}800$ characters per observable interface signature), and the controlled dual-axis \texttt{ITSM} \texttt{Propose}-stage ablation ($25$-turn cap, $B_W = 4{,}201$ characters, $32{,}768$ solver and $65{,}536$ distiller/reviewer output tokens; Table~\ref{tab:propose_seam_ablation}) all use our primary single-model \texttt{Gemini 4} configuration for every role. To evaluate \texttt{GUARD} as a drop-in stabilizer for teacher--student knowledge-distillation harnesses (\S\ref{sec:experiments}), \textsc{PAST-Bench}, the five-domain local \textsc{EvoHarnessBench} gym ($15$-turn cap), and the two professional suites pair a \texttt{gemini-3.8-flash} student solver with \texttt{claude-opus-5-5} on Vertex AI as reviewer, distiller, and $\mathsf{ContrastEvolve}$ model ($2{,}048$--$4{,}096$ output tokens for pre-execution checklist derivation, review, and $\mathsf{ContrastEvolve}$; $4{,}096$--$8{,}192$ for distillation and repair). No temperature reaches the paired teacher--student endpoints or the hosted \textsc{EvoHarnessBench} relay, whereas the \textsc{Drift Gauntlet} runner forwards explicit temperatures to \texttt{Gemini 4} (solver rollouts $0.2$; pre-execution checklist derivation and Layer-2 grading $0.0$; \texttt{GUARD} distiller candidates $0.65$, with $0.8$--$0.9$ on empty or duplicate retries; in-round \texttt{[iff]} repair $0.25$; $\mathsf{ContrastEvolve}$ $0.2$; baseline proposers $0.2$--$0.5$). $B_W$ is $2{,}500$ characters in the local gym, $2{,}400$ per skill body ($1{,}900$ per memory entry) on \textsc{PAST-Bench}, and $4{,}200$ characters of playbook body excluding the \texttt{Output Contract} ($5{,}600$ after an \texttt{[iff]} repair) in the professional suites. Boxes~\ref{box:turn0_reviewer_prompts}--\ref{box:distill_evolve_prompts} summarize the remaining prompt templates (complementing $s_0$ and $\mathsf{ContrastEvolve}$ output in Boxes~\ref{box:seed}--\ref{box:evolution_example}); we plan to release reference harness code, pre-execution checklist self-reviewers, and prompt templates, subject to approval.

\begin{promptlisting}[label={box:turn0_reviewer_prompts}]{Pre-execution checklist derivation, Layer-2 semantic grader, and PAST-Bench critic templates}
\textbf{(a) Pre-execution atomic checklist derivation} (hosted EvoHarnessBench, Drift Gauntlet, and professional suites; cached once before execution from public inputs only):\par
You are a deterministic rubric architect. Given ONLY the user prompt, visible tool declarations, and public schema (no trajectories or verifiers), decompose the request into concrete atomic binary sub-requirements (\texttt{req\_1}, \texttt{req\_2}, \ldots) covering clean execution, valid schema identifiers, idempotent mutations, and explicit prompt constraints. Return JSON \texttt{[\char`\{"id": "req\_k", "desc": "..."\char`\}]}.

\smallskip
\textbf{(b) Layer-2 per-criterion semantic self-reviewer} (an unparseable verdict counts as failed or falls back to deterministic trace checks):\par
You are a strict, reference-free Self-Reviewer with no access to hidden verifiers. Using ONLY \texttt{user\_prompt}, \texttt{schema\_summary}, \texttt{tool\_trace} / sandbox row deltas, and \texttt{final\_response}, grade each pre-registered criterion in \texttt{frozen\_sub\_requirements}. Reply strictly with JSON \texttt{\char`\{"passed": true|false, "why": "<evidence from trace and output>"\char`\}}.

\smallskip
\textbf{(c) PAST-Bench four-criterion critic} (one joint pass over the incumbent and $M=2$ candidates):\par
Score the incumbent and each candidate memory/skill state on $[0,100]$ across four criteria: (1)~grounding in session history without hallucinated preferences, (2)~exact format/SOP step fidelity, (3)~explicit scope/recipient \texttt{[iff]} boundaries avoiding episodic pollution, and (4)~complete replacement of superseded rules. Return per-family strict and continuous scores plus helped/hurt indicators.
\end{promptlisting}

\begin{promptlisting}[label={box:distill_evolve_prompts}]{$\mathsf{Distill}$ request wrapper and $\mathsf{ContrastEvolve}$ meta-prompt}
\textbf{(a) $\mathsf{Distill}(s_{t,m}, \mathcal{B}_{t,g}, W_{t-1,g})$ request wrapper} ($\le B_W$ chars, active ledger window $L_{\max}$):\par
\texttt{\#\#\# DISTILLER STRATEGY PROMPT}: \texttt{<}$s_{t,m}$\texttt{>} \quad $|\quad$ \texttt{\#\#\# INCUMBENT PLAYBOOK}: \texttt{<}$W_{t-1,g}$\texttt{>}\par
\texttt{\#\#\# REGRET \& CONTRAST BUNDLE}: (1)~\texttt{wrap\_in\_iff\_guard} ($\Delta_{\mathrm{hurt},t} > 0$), (2)~\texttt{add\_method\_and\_check} ($\Delta_{\mathrm{hurt},t} = 0, \Delta_{\mathrm{gap},t} > 0$), (3)~\texttt{pin\_mastered} ($\mathrm{Priority}_t = 0$, pinned invariants), (4)~grounded $(\tau^{+}, \tau^{-})$ trace contrasts, and (5)~active deduplicated \texttt{rejection\_ledger} entries (do not re-emit verbatim).

\smallskip
\textbf{(b) $\mathsf{ContrastEvolve}$ meta-prompt} (all task/episode IDs scrubbed to \texttt{<TASK\_ID>}):\par
Given the strategy prompt, playbook excerpt, and scrubbed metrics $(R_m, |\mathcal{Q}^{+}_m|, |\mathcal{Q}^{-}_m|)$ of the BEST candidate ($m^{+}$) and WORST candidate ($m^{-}$) in round $t$, extract causal \texttt{learnings} explaining why $m^{+}$ outperformed $m^{-}$ and rewrite both prompts into \texttt{\char`\{"learnings": [...], "updated\_\allowbreak best\_\allowbreak prompt": "...", "updated\_\allowbreak worst\_\allowbreak prompt": "..."\char`\}} so candidate~1 amplifies $m^{+}$'s scoped structure while preserving pinned rules and candidate~2 repairs $m^{-}$'s over-broad triggers into strict \texttt{[iff <precondition>]} guards.
\end{promptlisting}

\subsection{Shared Seed Prompt Example}
\label{app:seed_prompt}
In Round~1, every candidate in the pool receives the same strategy prompt $s_0$ (\texttt{SHARED\_SEED\_PROMPT}, Eq.~\eqref{eq:shared_seed}), shown in Box~\ref{box:seed}. $s_0$ is written once, before any task is seen, and names no benchmark, domain, tool, or task; in its wording, the \emph{champion} is the incumbent playbook $W_{t-1}$. We reproduce the prompts and the evolver output exactly, setting Markdown code spans in \texttt{typewriter} type and Markdown lists as lists.

\begin{promptlisting}[label={box:seed}]{Shared seed prompt $s_0$}
Diagnose the champion's dominant behavioral failure mode from its execution traces and failure report before designing anything. Then propose exactly ONE harness mechanism (prefer a skill: a SKILL.md workflow plus at most one small script) that directly removes that failure mode. Keep the mechanism narrow and guarded with explicit \texttt{[iff~<precondition>]} scope so it cannot fire on tasks the champion already solves. Validate it on a small fix slice plus a guard slice before emitting it.
\end{promptlisting}

\paragraph{One evolution step.} Box~\ref{box:evolution_example} shows the output of the contrastive evolver of \S\ref{sec:grpo_pool} after the first round of one of our runs ($M = 2$). In that run, each candidate is generated by a separate distiller call in which $s_0$ fills the strategy-prompt field, while the incumbent playbook, the regret bundle, and the rest of the request are shared; the two Round-1 requests were therefore byte-identical (the runner asserts this), and the two candidates differed only through sampling. Under execution scoring on the round's four source tasks, the incumbent averaged $\bar{S} = 75.00$ points. Candidate $m_1$ averaged $\bar{S}_1 = 93.75$ ($\bar{S}^{\mathrm{strict}}_1 = 84.375$) with one helped and no hurt tasks (pool reward $R_1 = (84.375 + \tfrac{1}{4}\cdot 93.75) - 3.5(0) + 1.5(1) = 109.31$, Eq.~\eqref{eq:grpo_reward}), and candidate $m_2$ averaged $\bar{S}_2 = 68.75$ ($\bar{S}^{\mathrm{strict}}_2 = 46.875$) with one helped and three hurt tasks ($R_2 = (46.875 + \tfrac{1}{4}\cdot 68.75) - 3.5(3) + 1.5(1) = 55.06$). The gate promoted $m_1$ under Soft-HURT with zero hurt tasks (net gain $18.75$ points), and the evolver contrasted $m^{+} = m_1$ with $m^{-} = m_2$. Its two rewritten prompts became the candidate strategy prompts of Round~2. Both keep the structure of $s_0$ (diagnose, propose one mechanism, validate on a fix slice and a guard slice): $s^{+}_{2}$ keeps the opening instruction of $s_0$ verbatim and adds the scoping and content rules that separated $m_1$ from $m_2$, while $s^{-}_{2}$ narrows the edit to the single most frequently failing tool call and lists what to stop doing.

\begin{promptlisting}[label={box:evolution_example}]{Contrastive evolver output after Round~1}
\textbf{learnings:}
\begin{itemize}[leftmargin=1.2em,itemsep=1pt,topsep=1pt,parsep=0pt]
    \item Trigger scope decides the outcome. The winner attached every rule to a concrete tool call or a literal request phrase, such as `calling create\_new\_hr\_case' or `creating it if necessary'. The loser opened with a broad `request names people, groups, services\ldots' trigger and an unconditional `[always, at final response]' rule. Those fire on nearly every task and hurt tasks the champion already solved.
    \item Unconditional end-of-task rules that tell the agent to take more actions are dangerous. `If a required write is still pending, perform it before replying' can cause spurious or duplicate writes. Final-response guidance should only interpret results, never trigger new actions.
    \item Precision in field formats matters. The winner said `numeric service id, not the name' and `leave the default unless the request states a status'. The loser said `resolved service id' and invented a draft-or-work\_in\_progress rule with extra synonyms. Vague wording leaves the agent guessing, and invented defaults override correct behavior.
    \item Complete the data chain each rule needs. The winner told the agent to resolve each matched profile to a user's first and last name before writing, because the write field needs a name. The loser listed the matches but never said how to get the field value.
    \item Order conditional lookups explicitly: look up the service first, and fetch topic details only if it is not found. Ambiguous order invites wasted or duplicate calls.
    \item Error-recovery wording should be narrow (`fix only the named field's format and retry'), not generic.
    \item Notification rules should name the payload (affected case numbers and outcome) and require that the send call actually succeeded.
\end{itemize}

\medskip
\textbf{updated\_best\_prompt} ($s^{+}_{2}$, candidate~1 in Round~2):\par
Diagnose the champion's dominant behavioral failure mode from its execution traces and failure report before designing anything. Then propose exactly ONE harness mechanism (prefer a skill: a SKILL.md workflow plus at most one small script) that directly removes that failure mode.

\smallskip
Scope rules:
\begin{itemize}[label={--},leftmargin=1.2em,itemsep=0pt,topsep=1pt,parsep=0pt]
    \item Gate every rule with an explicit \texttt{[iff~<precondition>]}.
    \item Each precondition must be either a specific tool call about to be made or a literal phrase in the request.
    \item Never use broad triggers like `request mentions entities'.
    \item Never write \texttt{[always]} rules.
    \item Put grounding steps inside the write-tool trigger they serve.
\end{itemize}
Content rules:
\begin{itemize}[label={--},leftmargin=1.2em,itemsep=0pt,topsep=1pt,parsep=0pt]
    \item For every field the failure touches, state the exact expected format (for example, display name vs numeric id) and where to obtain that value.
    \item Do not invent defaults. Say `leave the default unless the request states otherwise'.
    \item Spell out complete data chains:
    \begin{itemize}[label={--},leftmargin=1.2em,itemsep=0pt,topsep=0pt,parsep=0pt]
        \item lookup, then resolve to the exact value the write field needs;
        \item one write per item;
        \item never stop early.
    \end{itemize}
    \item Order conditional calls explicitly: check existence first, and fetch prerequisites only if the item is missing. Never create duplicates.
    \item Keep error recovery narrow: fix only the field named in the error, then retry.
    \item Success-interpretation rules must only define what counts as success or failure. They must never instruct the agent to take new actions at the end of a task.
    \item For notifications, name the payload and require that the send call actually succeeded before the agent claims it.
\end{itemize}
Validate on a small fix slice plus a guard slice of tasks the champion already solves. If any guard task changes behavior, narrow the trigger until it doesn't, then emit.

\medskip
\textbf{updated\_worst\_prompt} ($s^{-}_{2}$, candidate~2 in Round~2):\par
Do not write a broad playbook. Diagnose the champion's single most frequent failing tool call from its traces: the write call whose arguments most often fail or are wrong. Build ONE mechanism that fixes only that call: a SKILL.md plus at most one small validation script that checks the call's arguments before submission.

\smallskip
Requirements:
\begin{itemize}[label={--},leftmargin=1.2em,itemsep=0pt,topsep=1pt,parsep=0pt]
    \item The skill fires only \texttt{[iff about to call <that exact tool>]}. No global grounding rule, no \texttt{[always]} rule, and no final-response rule that triggers extra actions or `pending' writes.
    \item For each argument of that tool, list:
    \begin{itemize}[label={--},leftmargin=1.2em,itemsep=0pt,topsep=0pt,parsep=0pt]
        \item the exact format (display name vs numeric id, casing, enum values);
        \item the exact lookup tool and returned field that supplies it.
    \end{itemize}
    \item If an argument requires a value derived from another record (for example, a person's name from a profile), state the resolution step explicitly.
    \item Do not invent defaults or synonym mappings. Preserve the tool's default unless the request explicitly specifies a value.
    \item On a validation error, change only the named field and retry.
\end{itemize}
Stop doing:
\begin{itemize}[label={--},leftmargin=1.2em,itemsep=0pt,topsep=1pt,parsep=0pt]
    \item vague phrases like `resolved id';
    \item unordered lookup lists;
    \item catch-all summaries.
\end{itemize}
Validate on a fix slice where that tool failed, plus a guard slice of solved tasks. Emit only if the guard slice is unchanged.
\end{promptlisting}

\subsection{Time and Space Complexity Analysis}
\label{app:complexity}
This subsection derives the asymptotic time and space complexity of \texttt{GUARD} across adaptation and deployment, comparing it against the unified baseline suite (\texttt{No-Evolve} $W_0$, \textsc{Hermes+} / \textsc{MemToolAgent}, \textsc{RRSI}, \textsc{GEPA}, and \textsc{Meta-Harness}; Table~\ref{tab:complexity_comparison}). Let $|\mathcal{G}|$ be the number of independent groups, $Q = \sum_{g \in \mathcal{G}} |\mathcal{Q}_g|$ the total source tasks across groups, $T \le T_{\max}$ the adaptation rounds, $M \ge 2$ the candidate pool size ($K$ for multi-candidate baselines, with $M = K = 2$ in matched runs), $C = \max_q |\mathcal{C}(q)|$ the pre-execution checklist size ($4$--$8$ predicates, so $C = O(1)$), $H$ the per-task rollout step horizon, $B_W$ the bounded playbook size ($2.3\text{k}$--$4.2\text{k}$ characters, $B_W = O(1)$ in $T$), $L_{\max}$ the bounded active rejection-ledger read-back window ($\le 12$--$40$ entries per group, $L_{\max} = O(1)$ in $T$), and $B_{\mathcal{B}}$ the regret-bundle payload size passed to $\mathsf{Distill}$ (our runners cap rendered $(\tau^{+},\tau^{-})$ contrasts to $\le 6$ pairs and clip failure rationales to a fixed prompt budget so $B_{\mathcal{B}} = O(1)$ in $T$; without capping, $B_{\mathcal{B}} = O(|\mathcal{Q}_g|(C + H))$ under execution scoring, and critic scoring grows as $O(t|\mathcal{Q}_g|H)$ if accumulated trajectories are unwindowed).

\begin{table}[!htbp]
\centering
\scriptsize
\setlength{\tabcolsep}{3.6pt}
\caption{\textbf{Asymptotic adaptation, proposer-prefill, and deployment complexity across the unified six-arm suite.} Here $T$ is adaptation rounds, $Q$ source tasks ($|\mathcal{G}|$ groups), $M$ (or $K$) candidates per round ($M{=}K{=}2$ in matched runs), $H$ rollout step horizon, and $B_W, L_{\max}, B_{\mathcal{B}} = O(1)$ in $T$ the bounded playbook, active ledger-window, and capped regret-bundle sizes.}
\label{tab:complexity_comparison}
\begin{tabularx}{\textwidth}{@{}>{\raggedright\arraybackslash}p{0.205\textwidth}>{\raggedright\arraybackslash}p{0.155\textwidth}>{\raggedright\arraybackslash}p{0.155\textwidth}>{\raggedright\arraybackslash}p{0.235\textwidth}>{\raggedright\arraybackslash}X@{}}
\toprule
\textbf{Method} & \textbf{Adaptation Rollouts (Total)} & \textbf{Sequential Rollout Depth} & \textbf{Proposer Prefill Context (Round $t$)} & \textbf{Test-Time Rollouts \& Active Context} \\
\midrule
\texttt{No-Evolve} ($W_0$) & $0$ & $0$ & --- & $1\times$ rollout; $O(1)$ ctx. \\
\textsc{Hermes+} / \textsc{MemToolAgent} & $O(TQ)$ & $O(TQ)$ & $O(t)$ accumulated mem./skills & $1\times$ rollout; $O(T)$ ctx. \\
\textsc{RRSI} ($M{=}1$) & $O(TQ)$ & $O(TQ)$ & $O(B_W) = O(1)$ & $1\times$ rollout; $O(B_W) = O(1)$ ctx. \\
\textsc{GEPA} ($K$ cands.) & $O(TKQ)$ & $O(TQ)$ & $O(B_W) + O(tKQ)$ Pareto & $1\times$ rollout; $O(B_W) = O(1)$ ctx. \\
\textsc{Meta-Harness} ($K$ cands.) & $O(TKQ)$ & $O(TQ)$ & $O(tQH)$ raw trace archive & $1\times$ rollout; $O(B_W) = O(1)$ ctx. \\
\midrule
\rowcolor{LightBlue} \textbf{\texttt{GUARD} (Execution scoring)} & $\boldsymbol{O(TMQ)}$ ($M{=}K{=}2$) & $\boldsymbol{O(TQ)}$ ($M$-way par.) & $\boldsymbol{O(B_W + L_{\max} + B_{\mathcal{B}}) = O(1)}$ & $\boldsymbol{1\times}$ \textbf{rollout;} $\boldsymbol{O(B_W) = O(1)}$ \textbf{ctx.} \\
\rowcolor{LightBlue} \textbf{\texttt{GUARD} (Critic scoring)} & $\boldsymbol{O(TQ)}$ ($1\times$ online) & $\boldsymbol{O(TQ)}$ & $\boldsymbol{O(B_W + L_{\max} + B_{\mathcal{B}})}$$^{\star}$ & $\boldsymbol{1\times}$ \textbf{rollout;} $\boldsymbol{O(B_W) = O(1)}$ \textbf{ctx.} \\
\bottomrule
\end{tabularx}
\vspace{1pt}
{\scriptsize $^{\star}$$O(1)$ in $T$ when the episode window is capped; $O(t|\mathcal{Q}_g|H)$ when critic scoring inspects all unwindowed episodes.}
\end{table}

\paragraph{Stage-by-stage adaptation time complexity.}
Decomposing Algorithm~\ref{alg:unified_loop} across its stages isolates LLM rollout calls from local CPU bookkeeping:
\begin{enumerate}[leftmargin=*,noitemsep,topsep=2pt]
    \item \emph{Pre-execution checklist derivation (\S\ref{sec:self_review}):} Deriving the fixed atomic checklist $\mathcal{C}(q)$ requires a single pre-execution model pass per source task once at $t = 0$ ($O(Q)$ single-pass calls total; benchmark field stripping in \S\ref{sec:experiments} is $O(Q)$ local CPU work), cached unchanged across all $T$ rounds and $M$ parallel candidates.
    \item \emph{Solver rollouts (\S\ref{sec:grpo_pool}):} Multi-turn environment rollouts dominate wall-clock adaptation cost. Under \emph{execution scoring} on a fixed task set $\mathcal{Q}_g$ (Theorem~\ref{thm:gate}), \texttt{GUARD} executes one baseline pass under $W_0$ plus $M$ candidate passes per round, reusing recorded incumbent scores without re-running incumbents (plus at most $\mathbb{I}_{\mathrm{rep}} \in \{0,1\}$ conditional re-score when in-round \texttt{[iff]} repair fires on $1 \le |\mathcal{Q}^{-}_{m^{\star}}| \le h_{\mathrm{rep}}$, which adds at most one extra $Q$-task sweep in that round, i.e., worst-case per-round rollout depth $2Q$), totaling $(1 + T(M + \mathbb{I}_{\mathrm{rep}}))Q = O(TMQ)$ rollouts ($O(TMQH)$ environment steps). Because the $M$ candidates within round $t$ depend only on $(W_{t-1,g}, \mathcal{B}_{t,g}, \mathcal{P}_{t,g})$, they execute in parallel across workers and across the $|\mathcal{G}|$ groups, yielding $O(TQ)$ sequential rollout depth. This matches single-candidate \textsc{RRSI}/\textsc{Hermes+} in asymptotic wall-clock depth and matches $K$-candidate \textsc{GEPA}/\textsc{Meta-Harness} ($O(TKQ)$) in total work at $M = K = 2$. (On stage-expanded continual suites such as hosted \textsc{EvoHarnessBench}, where stage $t$ adds $q$ tasks under drifted tools and re-scores the incumbent and $M$ candidates on the cumulative $tq$ tasks, total rollouts scale as $\sum_{t=1}^{T}(M+1)tq = O(T^2 M q)$ with $O(T^2 q)$ sequential depth.) Under \emph{critic scoring} on stateful cross-session benchmarks (\textsc{PAST-Bench}), \texttt{GUARD} executes $O(TQ)$ ($1\times$) online episode rollouts, adding zero environment rollouts over \textsc{Hermes+}, \textsc{RRSI}, or \textsc{Meta-Harness}.
    \item \emph{Two-layer self-review (\S\ref{sec:self_review}):} Layer~1 grades execution-verifiable predicates deterministically in the local sandbox ($O(H)$ CPU per rollout, zero model calls). Layer~2 grades semantic predicates one at a time in $O(C)$ localized single-pass judgments per rollout, adding $O(TMQC)$ unbatched single-pass calls under execution scoring ($O(TQC)$ under critic scoring, with $C = O(1)$) without multi-turn tool interaction.
    \item \emph{Three closed-loop stages (\texttt{Analyze}, \texttt{Propose}, \texttt{Gate}):} In \texttt{Analyze} (\S\ref{sec:paired_regret}), computing $\mathrm{obey}(c; W)$, $\Delta_{\mathrm{hurt},t}(c)$, and $\Delta_{\mathrm{gap},t}(c)$ and sorting $QC$ criteria by $(\mathbb{I}[\Delta_{\mathrm{hurt},t}>0], \mathrm{Priority}_t(c), \Delta_{\mathrm{hurt},t}(c))$ takes $O(QC \log(QC))$ pure CPU time, and extracting same-task $(\tau^{+},\tau^{-})$ contrasts takes $O(QM)$ CPU time (\emph{zero} LLM calls). In \texttt{Propose} (\S\ref{sec:grpo_pool}), compiling $M$ candidates via $\mathsf{Distill}$ and evolving $(s^{+}_{t+1}, s^{-}_{t+1})$ via $\mathsf{ContrastEvolve}$ costs $M + 1$ single-pass distiller calls per group and round ($O(T|\mathcal{G}|M)$ total), while within-group standardized fitness normalization costs $O(T|\mathcal{G}|M)$ CPU time; because each distiller call reads bounded $(W_{t-1,g}, \mathcal{B}_{t,g})$ and at most $L_{\max}$ active ledger entries, cumulative proposer prefill tokens over $T$ rounds grow \emph{linearly} in $T$ as $O(T|\mathcal{G}|(B_W + L_{\max} + B_{\mathcal{B}}))$, whereas raw-trace archive search (\textsc{Meta-Harness}) grows per-round trace prefill as $O(tQH)$ and cumulative prefill quadratically as $\sum_{t=1}^{T} O(tQH) = O(T^2 Q H)$. In \texttt{Gate} (\S\ref{sec:rrsi_gate}), evaluating Eq.~\eqref{eq:soft_hurt_gate} and updating $\mathcal{L}_g$ takes $O(Q)$ CPU arithmetic (\emph{zero} LLM calls unless conditional \texttt{[iff]} repair triggers).
\end{enumerate}

\paragraph{Deployment (test-time) complexity and empirical runtime.}
At deployment, $\mathcal{P}_{t,g}$, the two-layer self-reviewer, and $\mathcal{L}_g$ are discarded: each held-out task $q_{\mathrm{test}}$ runs \textbf{a single ($1\times$) rollout} conditioned on $W_{T,g}$, incurring zero Best-of-$M$ sampling or reviewer calls and $O(B_W) = O(1)$ prompt-prefix overhead independent of $T$. Empirically, compact \texttt{[iff]}-guarded rules shorten test-time trajectories instead of slowing them down: on hosted \textsc{EvoHarnessBench} under unified \texttt{Gemini 4} ($43$ test tasks; Table~\ref{tab:evoharness_main}), \texttt{GUARD}'s final playbook ($2{,}315$ characters vs.\ $4{,}538$ in \textsc{Meta-Harness}, $5{,}123$ in \textsc{GEPA}, $6{,}989$ in \textsc{MemToolAgent}, and $10{,}013$ in \textsc{RRSI}), combined with two-layer gating (\S\ref{sec:calibration_ablation}), reaches $0/43$ eight-step budget exhaustions (vs.\ $6$--$10/43$ on baselines), cutting inference tokens per task to $59{,}651$ ($-54.5\%$ vs.\ $W_0$'s $131{,}032$, and $-24.2\%$ to $-30.0\%$ below evolving baselines) and keeping 43-task test evaluation wall-clock time ($167.6\text{ min}$, $233.8\text{s}$/task) within $6\%$ of $W_0$ ($158.8\text{ min}$) and $2.6\times$ faster than \textsc{RRSI} ($439.4\text{ min}$).

\paragraph{Space and active-context complexity.}
Memory scales along three distinct tiers: (i) \emph{active deployment prompt context} is $O(B_W) = O(1)$ in $T$ (and progressive \texttt{wiki\_read} loads only referenced sections on multi-file suites), avoiding the $O(T)$ prompt growth and $0.500$ memory pollution of uncompacted buffers; (ii) \emph{active proposer prefill context} is $O(B_W + L_{\max} + B_{\mathcal{B}})$ ($O(1)$ in $T$ under capped contrasts/rationales, vs.\ $O(|\mathcal{Q}_g|(C+H))$ uncapped), avoiding both the $O(TQH)$ raw-trace archive of \textsc{Meta-Harness} and the $O(TKQ)$ instance-Pareto score matrix of \textsc{GEPA}; and (iii) \emph{host working RAM and disk} hold $O(QC)$ cached pre-execution predicates plus $O(|\mathcal{G}| M B_W)$ active pool strings in RAM, while the deduplicated on-disk rejection log is windowed to $L_{\max} = O(1)$ active entries whenever read back into model context.

\section{Additional Experiments}
\label{app:additional}
This appendix evaluates \texttt{GUARD} across two public rubric-dense professional suites (\textsc{PRBench-Legal \& Harvey LAB} \cite{akyurek2025prbench} and \textsc{Fin-Rate \& Ledger} \cite{jiang2026finrate,moslonka2026ledger}; \S\ref{sec:public_domain_suites}), reports the complete 24-row per-stream continual breakdown and per-group learning dynamics on \textsc{Drift Gauntlet} \cite{chen2026continual} (\S\ref{app:drift_gauntlet_full}), and concludes with the five-domain \textsc{EvoHarnessBench} breakdown, matched \texttt{Analyze}/\texttt{Gate} leave-one-out ablation, and nine-row calibration audit (\S\ref{app:seam_ablation}).

\subsection{Setup and Active Mechanisms}
\label{app:additional_setup}
\paragraph{Scoring, splits, and models.} On the two rubric-dense professional suites, a judge grades each trajectory $\tau$ on a weighted rubric $\{c_j\}$ ($\sum_j w_j = 100$, score $v_j \in [0,1]$, critical flag $\kappa_j \in \{0,1\}$), giving \emph{Uncapped Score} $V(\tau) = \sum_j w_j v_j$ and \emph{Official Capped Score} (capped at $30$ if any critical criterion fails). Both suites evolve over three adaptation rounds (\texttt{R1}--\texttt{R3} from baseline $\texttt{R0} = W_0$) on 24 \texttt{TRAIN} tasks and evaluate post-hoc on 16 frozen \texttt{DEV} tasks ($N_{\mathrm{rep}} = 2$ single-rollout trials/task under the frozen round harness, with zero test-time candidate pooling or self-review), with all distillation, in-round \texttt{[iff]} repair, and Soft-HURT gating ($\delta_{\mathrm{noise}} = 15$) operating strictly reference-free on \texttt{TRAIN} self-reviews in hermetic workspaces with progressive \texttt{wiki\_read}. \textsc{PRBench-Legal \& Harvey LAB} \cite{akyurek2025prbench} is a public benchmark spanning statutory/contractual verification, redline reconciliation, compliance triage, and financial-legal covenants across \texttt{Legal} and \texttt{Finance} (1 \texttt{Finance} and 15 \texttt{Legal} \texttt{DEV} tasks); \textsc{Fin-Rate \& Ledger} pairs public SEC 10-K financial analytics QA from \textsc{Fin-RATE} \cite{jiang2026finrate} (\texttt{fin\_rate}) with corporate annual-report KPI extraction from \textsc{LEDGER} \cite{moslonka2026ledger} (\texttt{ledger}), stratified across \texttt{Finance}, \texttt{Legal}, \texttt{HR}, and \texttt{StratOps} ($3.125$-point Official steps) under a unified 100-point weighted verification rubric kept sealed from all adaptation loops. Across both professional suites and the five-domain \texttt{Analyze}/\texttt{Gate} ablation, \texttt{gemini-3.8-flash} serves as agent and \texttt{claude-opus-5-5} (Vertex AI) serves as distiller, self-reviewer, critic, and rubric judge, with strict within-suite model parity across all arms (Tables~\ref{tab:arms}--\ref{tab:ued_map}). Paired $W_0$ vs.\ $W_{\text{final}}$ comparisons on the professional suites use exact two-sided sign-flip permutation tests, task-level bootstrap confidence intervals, and pooled within-task replicate standard deviation $\sigma_w$.

\begin{table}[!htbp]
\centering\footnotesize
\setlength{\tabcolsep}{4pt}
\begin{tabularx}{\textwidth}{@{}l >{\raggedright\arraybackslash}X@{}}
\toprule
\textbf{Arm} & \textbf{Experience source and optimizer} \\
\midrule
\texttt{No-Evolve} ($W_0$) & Base agent executing without a learned playbook. \\
\texttt{RRSI} \cite{xia2026rrsi} & Multi-round failure-driven reflective playbook rewrite on \texttt{TRAIN} self-reviews. \\
\texttt{Meta-Harness} \cite{lee2026metaharness} & Counterfactual diagnostic harness search over full \texttt{TRAIN} execution histories. \\
\rowcolor{LightBlue} \textbf{\texttt{GUARD} (Ours)} & Paired priority $\Delta_{\mathrm{hurt}} + \Delta_{\mathrm{gap}}$, rule pinning, \texttt{[iff]} guards, prior-free contrastive prompt evolution, and Soft-HURT gating (\S\ref{sec:methodology}). \\
\bottomrule
\end{tabularx}
\caption{\textbf{Arm taxonomy for the professional-suite experiments.} All arms within each benchmark share identical agent and optimizer/judge models (\S\ref{app:additional_setup}); main-text six-arm details appear in \S\ref{sec:experiments}.}
\label{tab:arms}
\end{table}

\begin{table}[!htbp]
\centering\scriptsize
\setlength{\tabcolsep}{4pt}
\resizebox{\textwidth}{!}{%
\begin{tabular}{@{}>{\raggedright\arraybackslash}p{4.0cm}>{\raggedright\arraybackslash}p{2.9cm}>{\raggedright\arraybackslash}p{3.9cm}>{\raggedright\arraybackslash}p{3.2cm}>{\raggedright\arraybackslash}p{2.2cm}>{\raggedright\arraybackslash}p{1.9cm}@{}}
\toprule
\textbf{Experiment (rounds; splits)} & \textbf{Arm} & \textbf{\texttt{Analyze} stage} & \textbf{\texttt{Propose} stage} & \textbf{\texttt{Gate} rule} & \textbf{Reported in} \\
\midrule
\textsc{PRBench-Legal \& Harvey LAB} and \textsc{Fin-Rate \& Ledger} (3; 24 train, 16 frozen DEV$\times$2) & \texttt{No-Evolve} ($W_0$) / \texttt{RRSI} / \texttt{Meta-Harness} & none / \texttt{TRAIN} self-review / trace history & -- / reflective rewrite / harness search & -- / rewrite / trace search & Tab.~\ref{tab:public_domain_suites}, Fig.~\ref{fig:outer3_outer4_suites} \\
 & \textbf{\texttt{GUARD} (Ours)} & paired $\Delta_{\mathrm{hurt}} + \Delta_{\mathrm{gap}}$, rule pinning, $(\tau^{+},\tau^{-})$ contrasts & $M{\in}\{2,4\}$ pool, \texttt{[iff]} guards, $m^{+}\!/\!m^{-}$ evolution & \texttt{TRAIN} Soft-HURT & Tab.~\ref{tab:public_domain_suites}, Fig.~\ref{fig:outer3_outer4_suites} \\
\midrule
\textsc{EvoHarnessBench} 5-domain \texttt{Analyze}/\texttt{Gate} ablation (16 stages; 97 source, 236 test) & \texttt{No-Evolve} ($W_0$) / \texttt{w/o Analyze} / \texttt{w/o Gate} & -- / $1 - \mathrm{obey}$ (no pinning or contrasts) / full \texttt{Analyze} & -- / $M{=}2$ pool ($m^{+}\!/\!m^{-}$) / $M{=}2$ pool ($m^{+}\!/\!m^{-}$) & -- / Soft-HURT / unconditional & Tab.~\ref{tab:evoharness_seam_ablation} \\
 & \textbf{\texttt{GUARD} (Ours)} & paired $\Delta_{\mathrm{hurt}} + \Delta_{\mathrm{gap}}$, rule pinning, $(\tau^{+},\tau^{-})$ contrasts & $M{=}2$ pool, \texttt{[iff]} guards, $m^{+}\!/\!m^{-}$ evolution & Soft-HURT & Tab.~\ref{tab:evoharness_seam_ablation} \\
\bottomrule
\end{tabular}%
}
\caption{\textbf{Active closed-loop stages and reference-free gate rules across Appendix~\ref{app:additional}.} Every arm adapts strictly without reference solutions, oracle skills, or held-out rubrics; \texttt{TRAIN Soft-HURT} gates on \texttt{TRAIN} pre-execution checklist self-reviews ($\delta_{\mathrm{noise}} = 15$, in-round \texttt{[iff]} repair; \S\ref{sec:rrsi_gate}).}
\label{tab:ued_map}
\end{table}

\subsection{Public Professional Suites (\textsc{PRBench-Legal} and \textsc{Fin-Rate \& Ledger})}
\label{sec:public_domain_suites}
\begin{table}[!htbp]
\centering\footnotesize
\setlength{\tabcolsep}{4.2pt}
\resizebox{\textwidth}{!}{%
\begin{tabular}{@{}llcccccccccl@{}}
\toprule
\textbf{Suite (24 \texttt{TRAIN} $\to$ 16 Frozen \texttt{DEV})} & \textbf{Arm} & \textbf{R0} & \textbf{R1} & \textbf{R2} & \textbf{R3 (Final)} & \textbf{$\Delta$ Official vs.\ $W_0$ [95\% CI]} & \textbf{Final Uncapped} & \textbf{$\Delta$ Uncapped} & \textbf{Hurt} & \textbf{Sub-domain Breakdown (Official)} \\
\midrule
\textbf{\textsc{PRBench-Legal \& Harvey LAB}} & \texttt{No-Evolve} ($W_0$) & 59.48\% & 59.48\% & 59.48\% & 59.48\% & --- & 59.77\% & --- & --- & Finance ($n{=}1$): 48.05\% $\;|\;$ Legal: 60.24\% \\
(\texttt{gemini-3.8-flash} agent, & \texttt{RRSI} & 59.48\% & 64.91\% & 65.42\% & 63.30\% & $+3.82\text{pp}$ $[-0.21, +8.06]$ & 63.59\% & $+3.82\text{pp}$ & \textbf{0} & Finance ($n{=}1$): 72.08\% $\;|\;$ Legal: 62.72\% \\
\texttt{claude-opus-5-5} distiller/judge) & \texttt{Meta-Harness} & 59.48\% & 63.81\% & 65.49\% & 66.11\% & $+6.63\text{pp}$ $[+2.52, +11.04]$ & 66.40\% & $+6.63\text{pp}$ & \textbf{0} & Finance ($n{=}1$): \textbf{76.62\%} $\;|\;$ Legal: 65.41\% \\
 & \cellcolor{LightBlue}\textbf{\texttt{GUARD} (Ours)} & \cellcolor{LightBlue}59.48\% & \cellcolor{LightBlue}\textbf{66.67\%} & \cellcolor{LightBlue}\textbf{66.67\%} & \cellcolor{LightBlue}\textbf{66.67\%} & \cellcolor{LightBlue}$\mathbf{+7.19\text{\textbf{pp}}}$ $[+1.85, +12.85]^{*}$ & \cellcolor{LightBlue}\textbf{66.96\%} & \cellcolor{LightBlue}$\mathbf{+7.19\text{\textbf{pp}}}$ & \cellcolor{LightBlue}\textbf{0} & \cellcolor{LightBlue}Finance ($n{=}1$): \textbf{76.62\%} $\;|\;$ Legal: \textbf{66.01\%} \\
\midrule
\textbf{\textsc{Fin-Rate \& Ledger Suite}} & \texttt{No-Evolve} ($W_0$) & 37.50\% & 37.50\% & 37.50\% & 37.50\% & --- & 70.58\% & --- & --- & \texttt{fin\_rate}: 12.50\% $\;|\;$ \texttt{ledger}: 62.50\% \\
(\texttt{gemini-3.8-flash} agent, & \texttt{RRSI} & 37.50\% & \textbf{46.88\%} & 40.62\% & 40.62\% & $+3.12\text{pp}$ $[-6.25, +12.50]$ & 71.50\% & $+0.92\text{pp}$ & 1 & \texttt{fin\_rate}: \textbf{18.75\%} $\;|\;$ \texttt{ledger}: 62.50\% \\
\texttt{claude-opus-5-5} distiller/judge) & \texttt{Meta-Harness} & 37.50\% & 37.50\% & 40.62\% & 40.62\% & $+3.12\text{pp}$ $[+0.00, +9.38]$ & 73.54\% & $+2.96\text{pp}$ & \textbf{0} & \texttt{fin\_rate}: \textbf{18.75\%} $\;|\;$ \texttt{ledger}: 62.50\% \\
 & \cellcolor{LightBlue}\textbf{\texttt{GUARD} (Ours)} & \cellcolor{LightBlue}37.50\% & \cellcolor{LightBlue}43.75\% & \cellcolor{LightBlue}\textbf{50.00\%} & \cellcolor{LightBlue}\textbf{50.00\%} & \cellcolor{LightBlue}$+12.50\text{pp}$ $[+0.00, +31.25]$ & \cellcolor{LightBlue}\textbf{82.96\%} & \cellcolor{LightBlue}$\mathbf{+12.38\text{\textbf{pp}}}$ & \cellcolor{LightBlue}\textbf{0} & \cellcolor{LightBlue}\texttt{fin\_rate}: 12.50\% $\;|\;$ \texttt{ledger}: \textbf{87.50\%} \\
\bottomrule
\end{tabular}%
}
\caption{\textbf{Multi-round frozen \texttt{DEV} results (\texttt{R0} $\to$ \texttt{R3}) on the two professional suites} (24 \texttt{TRAIN} tasks, reference-free gating; 16 frozen \texttt{DEV} tasks $\times$ 2 trials). $^{*}$Exact two-sided sign-flip $p = 0.0251$ ($M = 4$ prior-free, four helped and zero hurt tasks; $M = 2$ prior-free reaches $65.76\%$, $+6.28\text{pp}$); on \textsc{Fin-Rate \& Ledger} ($M = 2$, $n = 16$), the 95\% CI $[+0.00, +31.25]$ touches $0$ (two helped \texttt{ledger} tasks, exact sign-flip $p = 0.50$).}
\label{tab:public_domain_suites}
\end{table}

\begin{figure}[!htbp]
\centering
\includegraphics[width=0.49\textwidth]{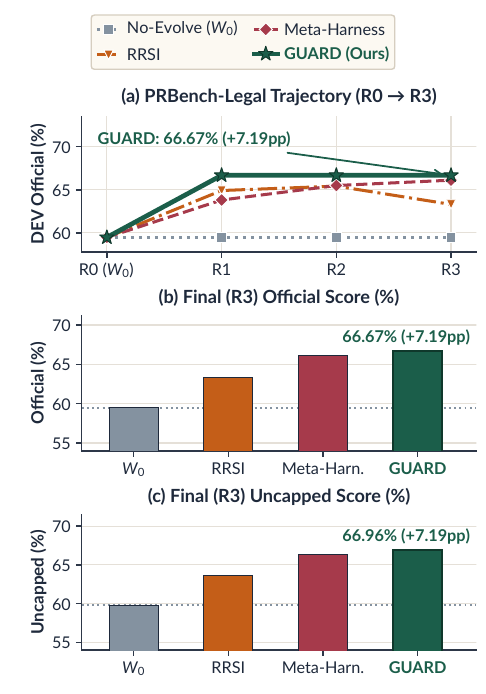}\hfill
\includegraphics[width=0.49\textwidth]{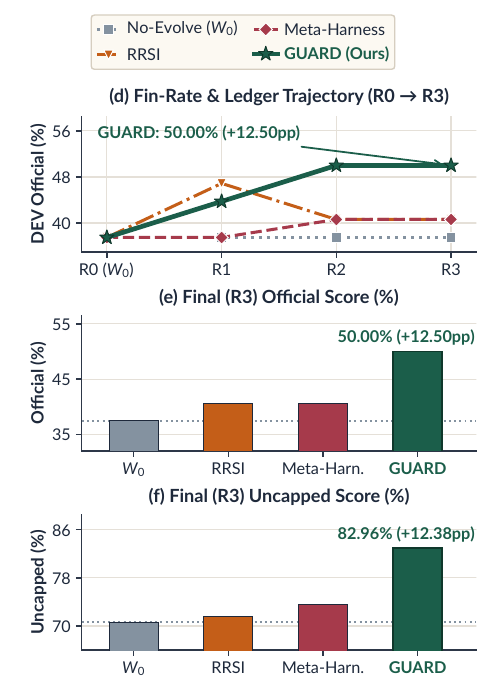}
\caption{\textbf{Frozen \texttt{DEV} trajectories (\texttt{R0} $\to$ \texttt{R3}, top), final Official (capped) scores (middle), and final Uncapped scores (bottom) on \textsc{PRBench-Legal \& Harvey LAB} (panels (a)--(c), left) and \textsc{Fin-Rate \& Ledger} (panels (d)--(f), right).} \texttt{GUARD} is non-decreasing with zero hurt tasks on both suites ($66.67\%$ Official / $66.96\%$ Uncapped on \textsc{PRBench-Legal}; $50.00\%$ Official / $82.96\%$ Uncapped with $\sigma_w = 0.00$ on \textsc{Fin-Rate \& Ledger}).}
\label{fig:outer3_outer4_suites}
\end{figure}

Table~\ref{tab:public_domain_suites} and Figure~\ref{fig:outer3_outer4_suites} show two main results under \texttt{TRAIN}-only reference-free gating. \texttt{GUARD} eliminates the late-round regressions of unguarded \texttt{RRSI}: on public \textsc{PRBench-Legal \& Harvey LAB}, \texttt{RRSI} drops $2.12\text{pp}$ in \texttt{R3} as statutory exception rules collide ($63.30\%$, $\sigma_w = 7.11$) and \texttt{Meta-Harness} reaches $66.11\%$ ($\sigma_w = 7.40$), whereas \texttt{GUARD} ($M = 4$ prior-free) reaches $66.67\%$ Official ($+7.19\text{pp}$ over $W_0$, 95\% CI $[+1.85, +12.85]$, exact sign-flip $p = 0.0251$; $66.96\%$ Uncapped) with zero hurt tasks ($4$ helped / $0$ hurt) and the lowest replicate variance among evolving arms ($\sigma_w = 5.56$). It lifts Tier-1 critical pass rates from $54.90\%$ to $62.10\%$ and Tier-2 from $63.85\%$ to $69.38\%$, advancing both \texttt{Legal} ($60.24\% \to 66.01\%$, $n = 15$) and the single \texttt{Finance} \texttt{DEV} task ($48.05\% \to 76.62\%$, $n = 1$). On \textsc{Fin-Rate \& Ledger} \cite{jiang2026finrate,moslonka2026ledger}, where \texttt{RRSI} drops $6.26\text{pp}$ after \texttt{R1} ($\sigma_w = 30.62$, 1 hurt task) and \texttt{Meta-Harness} stalls at $40.62\%$, \texttt{GUARD} reaches $50.00\%$ Official ($+12.50\text{pp}$ over $W_0$, 95\% CI $[+0.00, +31.25]$ touching $0$ at $n = 16$, 0 hurt, $\sigma_w = 0.00$) and $82.96\%$ Uncapped, lifting SEC 10-K \texttt{ledger} extraction from $62.50\%$ to $87.50\%$ ($7/8$ \texttt{DEV} tasks solved across both replicates while \texttt{fin\_rate} remains at $12.50\%$).

\subsection{Per-Stream Continual Breakdown and Dynamics on \textsc{Drift Gauntlet}}
\label{app:drift_gauntlet_full}
\begin{figure}[!htbp]
\centering
\includegraphics[width=0.98\textwidth]{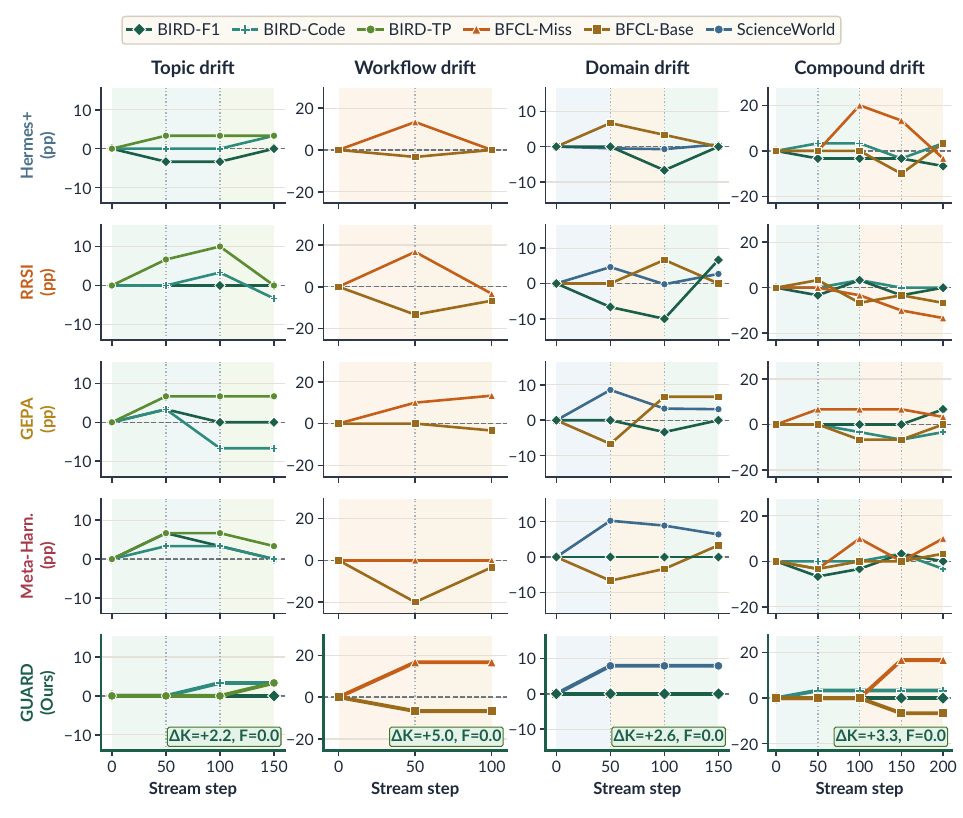}
\caption{\textbf{Per-group continual learning dynamics $A[t, k] - B_k$ across all five evolving methods and four \textsc{Drift Gauntlet} scenarios under single-model \texttt{Gemini 4}} (shaded spans mark each group's active online adaptation block; layout following Figure~1 of \cite{chen2026continual}). Baselines exhibit sharp post-block drops once subsequent blocks arrive, whereas \texttt{GUARD} (bottom row) retains $100\%$ of post-block gains ($F = 0.00\text{pp}$ on all four streams).}
\label{fig:drift_gauntlet_group_dynamics}
\end{figure}

Table~\ref{tab:drift_gauntlet_full} reports all $24$ $(\text{stream}, \text{arm})$ evaluations underlying Table~\ref{tab:drift_gauntlet_main} (\S\ref{sec:drift_gauntlet}), and Figure~\ref{fig:drift_gauntlet_group_dynamics} plots the $5 \times 4$ per-group continual advantage trajectories $A[t, k] - B_k$ (modeled on Figure~1 of Chen et~al.\ \cite{chen2026continual}). Across every scenario, \texttt{GUARD} achieves $F = 0.00\text{pp}$ worst-case post-block forgetting and zero downward block transitions in $\bar{A}_t$: on opposing \emph{workflow drift} (\texttt{BFCL-Miss} $\to$ \texttt{BFCL-Base}), Block~1 lifts \texttt{BFCL-Miss} $+16.67\text{pp}$ ($53.33\% \to 70.00\%$) and Soft-HURT rejects Block~2's over-broad rewrite (where $10$ sampled \texttt{bfcl} reservoir replay tasks detected $\text{hurt} = \text{lift} = 6.25\text{pp}$), avoiding the $-20.00\text{pp}$ and $-13.34\text{pp}$ collapses of \textsc{RRSI} and \textsc{Hermes+}; on \emph{topic drift} (\texttt{BIRD-F1} $\to$ \texttt{BIRD-Code} $\to$ \texttt{BIRD-TP}), \texttt{GUARD} promotes clean database-scoped sections across all three blocks ($I = F = S = 0.00\text{pp}$); on cross-environment \emph{domain drift}, in-round \texttt{[iff]} repair narrows Block~1 (\texttt{ScienceWorld} $74.80\% \to 82.73\%$, $+7.93\text{pp}$) while Soft-HURT rejects Block~2; and on four-block \emph{compound drift}, \texttt{GUARD} leads Final Mean ($74.17\%$), $\Delta_K$ ($+3.33\text{pp}$), $\bar{\Delta}$ ($+2.08\text{pp}$), and plasticity ($+5.00\text{pp}$) with $F = 0.00\text{pp}$.

\begin{table}[!htbp]
\centering
\scriptsize
\setlength{\tabcolsep}{3.6pt}
\caption{\textbf{Complete 24-row per-stream continual RSI results on \textsc{Drift Gauntlet} \cite{chen2026continual} under single-model \texttt{Gemini 4} ($30$ adaptation tasks/block, $30$ held-out test tasks/group; summarized in Table~\ref{tab:drift_gauntlet_main}).} $A[K, \cdot]$ lists final checkpoint accuracy across the stream's $K$ task groups in arrival order. Bold marks the best evolving arm within each stream.}
\label{tab:drift_gauntlet_full}
\resizebox{\textwidth}{!}{%
\begin{tabular}{@{}llcccccccl@{}}
\toprule
\textbf{Continual Drift Stream ($K$ blocks)} & \textbf{Arm} & \textbf{Final Mean} & \textbf{$\Delta_K$} & \textbf{$\bar{\Delta}$} & \textbf{Plast.\ $P$} & \textbf{Interf.\ $I$} & \textbf{Forget.\ $F$} & \textbf{Stab.\ $S$} & \textbf{Final Group Vector $A[K, \cdot]$ (\%)} \\
\midrule
\textbf{Workflow Drift} ($K{=}2$) & \texttt{No-Evolve} ($W_0$) & 65.00\% & $0.00\text{pp}$ & $0.00\text{pp}$ & $0.00\text{pp}$ & $0.00\text{pp}$ & $0.00\text{pp}$ & $0.00\text{pp}$ & $[53.33,\; 76.67]$ \\
\texttt{BFCL-Miss} $\to$ \texttt{BFCL-Base} & \textsc{Hermes+} / \textsc{MemToolAgent} & 65.00\% & $0.00\text{pp}$ & $+2.50\text{pp}$ & $+8.34\text{pp}$ & $-3.34\text{pp}$ & $-13.34\text{pp}$ & $-8.34\text{pp}$ & $[53.33,\; 76.67]$ \\
 & \textsc{RRSI} \cite{xia2026rrsi} & 60.00\% & $-5.00\text{pp}$ & $-1.67\text{pp}$ & $\mathbf{+11.67\text{pp}}$ & $-13.34\text{pp}$ & $-20.00\text{pp}$ & $-16.67\text{pp}$ & $[50.00,\; 70.00]$ \\
 & \textsc{GEPA} \cite{agrawal2026gepa} & \textbf{70.00\%} & $\mathbf{+5.00\text{pp}}$ & $\mathbf{+5.00\text{pp}}$ & $+3.33\text{pp}$ & $\mathbf{0.00\text{pp}}$ & $\mathbf{+3.34\text{pp}}$ & $\mathbf{+1.67\text{pp}}$ & $[66.67,\; 73.33]$ \\
 & \textsc{Meta-Harness} \cite{lee2026metaharness} & 63.33\% & $-1.67\text{pp}$ & $-5.84\text{pp}$ & $+8.33\text{pp}$ & $-20.00\text{pp}$ & $0.00\text{pp}$ & $-10.00\text{pp}$ & $[53.33,\; 73.33]$ \\
 & \cellcolor{LightBlue}\textbf{\texttt{GUARD} (Ours)} & \cellcolor{LightBlue}\textbf{70.00\%} & \cellcolor{LightBlue}$\mathbf{+5.00\text{pp}}$ & \cellcolor{LightBlue}$\mathbf{+5.00\text{pp}}$ & \cellcolor{LightBlue}$+8.34\text{pp}$ & \cellcolor{LightBlue}$-6.67\text{pp}$ & \cellcolor{LightBlue}$0.00\text{pp}$ & \cellcolor{LightBlue}$-3.34\text{pp}$ & \cellcolor{LightBlue}$[\mathbf{70.00},\; 70.00]$ \\
\midrule
\textbf{Topic Drift} ($K{=}3$) & \texttt{No-Evolve} ($W_0$) & 67.78\% & $0.00\text{pp}$ & $0.00\text{pp}$ & $0.00\text{pp}$ & $0.00\text{pp}$ & $0.00\text{pp}$ & $0.00\text{pp}$ & $[76.67,\; 76.67,\; 50.00]$ \\
\texttt{BIRD-F1} $\to$ \texttt{BIRD-Code} & \textsc{Hermes+} / \textsc{MemToolAgent} & \textbf{70.00\%} & $\mathbf{+2.22\text{pp}}$ & $+0.74\text{pp}$ & $-1.11\text{pp}$ & $+1.66\text{pp}$ & $\mathbf{+1.66\text{pp}}$ & $+1.66\text{pp}$ & $[76.67,\; \mathbf{80.00},\; 53.33]$ \\
$\to$ \texttt{BIRD-TP} & \textsc{RRSI} \cite{xia2026rrsi} & 66.67\% & $-1.11\text{pp}$ & $+1.85\text{pp}$ & $-2.22\text{pp}$ & $+4.17\text{pp}$ & $-3.34\text{pp}$ & $+0.42\text{pp}$ & $[76.67,\; 73.33,\; 50.00]$ \\
 & \textsc{GEPA} \cite{agrawal2026gepa} & 67.78\% & $0.00\text{pp}$ & $+1.48\text{pp}$ & $-2.22\text{pp}$ & $\mathbf{+5.00\text{pp}}$ & $-1.66\text{pp}$ & $\mathbf{+1.67\text{pp}}$ & $[76.67,\; 70.00,\; \mathbf{56.67}]$ \\
 & \textsc{Meta-Harness} \cite{lee2026metaharness} & 68.89\% & $+1.11\text{pp}$ & $\mathbf{+3.70\text{pp}}$ & $+1.11\text{pp}$ & $\mathbf{+5.00\text{pp}}$ & $-4.99\text{pp}$ & $0.00\text{pp}$ & $[76.67,\; 76.67,\; 53.33]$ \\
 & \cellcolor{LightBlue}\textbf{\texttt{GUARD} (Ours)} & \cellcolor{LightBlue}\textbf{70.00\%} & \cellcolor{LightBlue}$\mathbf{+2.22\text{pp}}$ & \cellcolor{LightBlue}$+1.11\text{pp}$ & \cellcolor{LightBlue}$\mathbf{+2.22\text{pp}}$ & \cellcolor{LightBlue}$0.00\text{pp}$ & \cellcolor{LightBlue}$0.00\text{pp}$ & \cellcolor{LightBlue}$0.00\text{pp}$ & \cellcolor{LightBlue}$[76.67,\; \mathbf{80.00},\; 53.33]$ \\
\midrule
\textbf{Domain Drift} ($K{=}3$) & \texttt{No-Evolve} ($W_0$) & 76.05\% & $0.00\text{pp}$ & $0.00\text{pp}$ & $0.00\text{pp}$ & $0.00\text{pp}$ & $0.00\text{pp}$ & $0.00\text{pp}$ & $[74.80,\; 76.67,\; 76.67]$ \\
\texttt{ScienceWorld} $\to$ \texttt{BFCL-Base} & \textsc{Hermes+} / \textsc{MemToolAgent} & 76.26\% & $+0.21\text{pp}$ & $+0.31\text{pp}$ & $+0.96\text{pp}$ & $\mathbf{+1.66\text{pp}}$ & $-1.78\text{pp}$ & $-0.06\text{pp}$ & $[75.43,\; 76.67,\; 76.67]$ \\
$\to$ \texttt{BIRD-F1} & \textsc{RRSI} \cite{xia2026rrsi} & 79.17\% & $+3.12\text{pp}$ & $+0.42\text{pp}$ & $\mathbf{+9.32\text{pp}}$ & $-4.17\text{pp}$ & $-5.76\text{pp}$ & $-4.96\text{pp}$ & $[77.50,\; 76.67,\; \mathbf{83.33}]$ \\
 & \textsc{GEPA} \cite{agrawal2026gepa} & \textbf{79.31\%} & $\mathbf{+3.26\text{pp}}$ & $+2.03\text{pp}$ & $+8.41\text{pp}$ & $-4.17\text{pp}$ & $-2.72\text{pp}$ & $-3.45\text{pp}$ & $[77.93,\; \mathbf{83.33},\; 76.67]$ \\
 & \textsc{Meta-Harness} \cite{lee2026metaharness} & 79.29\% & $+3.24\text{pp}$ & $+2.10\text{pp}$ & $+4.53\text{pp}$ & $-3.34\text{pp}$ & $\mathbf{+1.40\text{pp}}$ & $-0.97\text{pp}$ & $[81.20,\; 80.00,\; 76.67]$ \\
 & \cellcolor{LightBlue}\textbf{\texttt{GUARD} (Ours)} & \cellcolor{LightBlue}78.69\% & \cellcolor{LightBlue}$+2.64\text{pp}$ & \cellcolor{LightBlue}$\mathbf{+2.64\text{pp}}$ & \cellcolor{LightBlue}$+2.64\text{pp}$ & \cellcolor{LightBlue}$0.00\text{pp}$ & \cellcolor{LightBlue}$0.00\text{pp}$ & \cellcolor{LightBlue}$\mathbf{0.00\text{pp}}$ & \cellcolor{LightBlue}$[\mathbf{82.73},\; 76.67,\; 76.67]$ \\
\midrule
\textbf{Compound Drift} ($K{=}4$) & \texttt{No-Evolve} ($W_0$) & 70.84\% & $0.00\text{pp}$ & $0.00\text{pp}$ & $0.00\text{pp}$ & $0.00\text{pp}$ & $0.00\text{pp}$ & $0.00\text{pp}$ & $[76.67,\; 76.67,\; 53.33,\; 76.67]$ \\
\texttt{BIRD-Code} $\to$ \texttt{BIRD-F1} & \textsc{Hermes+} / \textsc{MemToolAgent} & 70.00\% & $-0.84\text{pp}$ & $+0.83\text{pp}$ & $+2.50\text{pp}$ & $\mathbf{+1.11\text{pp}}$ & $-8.89\text{pp}$ & $-3.89\text{pp}$ & $[\mathbf{80.00},\; 70.00,\; 50.00,\; \mathbf{80.00}]$ \\
$\to$ \texttt{BFCL-Miss} $\to$ \texttt{BFCL-Base} & \textsc{RRSI} \cite{xia2026rrsi} & 65.84\% & $-5.00\text{pp}$ & $-2.50\text{pp}$ & $-0.83\text{pp}$ & $-2.41\text{pp}$ & $-3.33\text{pp}$ & $-2.87\text{pp}$ & $[76.67,\; 76.67,\; 40.00,\; 70.00]$ \\
 & \textsc{GEPA} \cite{agrawal2026gepa} & 72.50\% & $+1.66\text{pp}$ & $+0.21\text{pp}$ & $+1.67\text{pp}$ & $+0.74\text{pp}$ & $-3.33\text{pp}$ & $-1.30\text{pp}$ & $[73.33,\; \mathbf{83.33},\; 56.67,\; 76.67]$ \\
 & \textsc{Meta-Harness} \cite{lee2026metaharness} & 73.33\% & $+2.50\text{pp}$ & $+0.62\text{pp}$ & $-0.84\text{pp}$ & $-1.48\text{pp}$ & $\mathbf{+3.33\text{pp}}$ & $\mathbf{+0.93\text{pp}}$ & $[73.33,\; 76.67,\; 63.33,\; \mathbf{80.00}]$ \\
 & \cellcolor{LightBlue}\textbf{\texttt{GUARD} (Ours)} & \cellcolor{LightBlue}\textbf{74.17\%} & \cellcolor{LightBlue}$\mathbf{+3.33\text{pp}}$ & \cellcolor{LightBlue}$\mathbf{+2.08\text{pp}}$ & \cellcolor{LightBlue}$\mathbf{+5.00\text{pp}}$ & \cellcolor{LightBlue}$-0.74\text{pp}$ & \cellcolor{LightBlue}$0.00\text{pp}$ & \cellcolor{LightBlue}$-0.37\text{pp}$ & \cellcolor{LightBlue}$[\mathbf{80.00},\; 76.67,\; \mathbf{70.00},\; 70.00]$ \\
\bottomrule
\end{tabular}%
}
\end{table}

\begin{promptlisting}[label={box:drift_example}]{An admitted scoped rule and a rejected overwrite on \textsc{Drift Gauntlet}}
\textbf{(a) Admitted scoped rule} (\emph{compound drift}, Block~1, section keyed to signature \texttt{bird:codebase\_community}; validation net lift $+5.21\text{pp}$ with no hurt task; held-out \texttt{BIRD-Code} $76.67\% \to 80.00\%$). One of the promoted Tier-2 rules, verbatim:\par
\texttt{[iff the prompt or Evidence expresses year or date predicates such as YEAR(col) = YYYY in a SQLite environment]} Replace unsupported \texttt{YEAR()} calls with valid SQLite date extraction (\texttt{STRFTIME('\%Y', col) = 'YYYY'} or \texttt{col LIKE 'YYYY\%'}) and enforce the year filter on every date column specified in the Evidence.

\smallskip
\textbf{(b) Rejected overwrite} (\emph{workflow drift}, Block~2, shared signature \texttt{bfcl}). After \texttt{BFCL-Base} arrived, the pool winner rewrote the shared section. On the $24$-task validation buffer it lifted the mean score by $6.25\text{pp}$ but regressed four tasks (three \texttt{BFCL-Miss} reservoir replay tasks and one \texttt{BFCL-Base} task) with a hurt penalty of $6.25\text{pp}$, so Soft-HURT kept $W_1$ and logged the candidate to the rejection ledger. Held-out \texttt{BFCL-Miss} stayed at $70.00\%$, whereas the ungated \textsc{RRSI} rewrite lost $20.00\text{pp}$ on that group.
\end{promptlisting}

\subsection{Five-Domain \textsc{EvoHarnessBench} Breakdown, Component Ablations (\texttt{Analyze}, \texttt{Propose}, \texttt{Gate}), and Full Calibration Audit}
\label{app:seam_ablation}
Table~\ref{tab:evoharness_domains} reports the per-domain continual evolution breakdown across all five local-gym \textsc{EvoHarnessBench} domains alongside the hosted \texttt{Calendar} track, and Table~\ref{tab:calibration_gate_fidelity_full} gives the complete nine-row offline self-reviewer calibration and Soft-HURT gate audit summarized in Table~\ref{tab:calibration_gate_fidelity} (\S\ref{sec:calibration_ablation}). We evaluate component-level ablations across both model regimes introduced in \S\ref{sec:experiments}: Table~\ref{tab:evoharness_seam_ablation} tests leave-one-out removal of \texttt{Analyze} and \texttt{Gate} across all five local-gym \textsc{EvoHarnessBench} domains (\texttt{Calendar}, \texttt{Drive}, \texttt{CSM}, \texttt{HR}, and \texttt{ITSM}; $16$ continual stages, $97$ visible adaptation tasks, and $236$ held-out test tasks) in the teacher--student distillation setting (\texttt{gemini-3.8-flash} solver paired with \texttt{claude-opus-5-5} self-reviewer, skill distiller, and prompt mutator, $M = 2$, seed $s_0$), while Table~\ref{tab:propose_seam_ablation} isolates the \texttt{Propose} stage under our primary single-model \texttt{Gemini 4} configuration across both \textsc{EvoHarnessBench} \texttt{ITSM} evolution axes (\texttt{evolving\_tools}, $T{=}3$ stages, and \texttt{evolving\_skills}, $T{=}4$ stages; pooled $N{=}132$ held-out test tasks $\times N_{\mathrm{rep}}{=}3$ replicates across four controlled configurations, strictly isolated rollouts with zero cache sharing and train-only self-review, distillation, and Soft-HURT gating at $\delta_{\mathrm{noise}} = 15$):
\begin{itemize}[leftmargin=1.4em,itemsep=2pt,topsep=3pt]
    \item \textbf{Necessity of the \texttt{Analyze} stage (paired regret ranking, rule pinning, and contrasts).} Jointly ablating $\mathrm{Priority}_t(c) = \Delta_{\mathrm{hurt},t}(c) + \Delta_{\mathrm{gap},t}(c)$, rule pinning, and paired $(\tau^{+}, \tau^{-})$ trace contrasts in favor of single-round failure frequency $1 - \mathrm{obey}(c; W_{t-1})$ (\texttt{GUARD w/o Analyze}), while keeping the identical $M = 2$ candidate pool and Soft-HURT promotion gate, reduces five-domain macro accuracy by $-7.33\text{pp}$ ($66.46\% \to 59.13\%$, $-6.68\text{pp}$ below static $W_0$), strict solve rate by $-5.39\text{pp}$ ($38.31\% \to 32.92\%$), and mean forward transfer to $-7.57\text{pp}$ (vs.\ $+1.00\text{pp}$ under \texttt{GUARD}). The Soft-HURT gate alone cannot rescue an unscoped proposer: it still promotes four stages on visible source tasks whose frequency-distilled rules overfit early stages and regress both multi-stage skill domains on held-out cohorts (\texttt{CSM} drops $61.40\% \to 38.94\%$, $-22.46\text{pp}$; \texttt{ITSM} drops $60.28\% \to 46.12\%$, $-14.16\text{pp}$).
    \item \textbf{Necessity of the Soft-HURT promotion gate (\texttt{Gate} stage).} Retaining paired regret priority, rule pinning, and contrastive candidate generation while unconditionally promoting the within-group standardized-fitness winner across all $16$ stages drops five-domain macro accuracy by $-10.35\text{pp}$ ($66.46\% \to 56.11\%$), strict solve rate by $-10.04\text{pp}$ ($38.31\% \to 28.27\%$), mean forward transfer to $-5.62\text{pp}$, and weighted backward transfer to $-6.90\text{pp}$. Ungated promotion causes negative transfer vs.\ $W_0$ on $3 / 5$ domains (\texttt{Calendar} $-1.89\text{pp}$, \texttt{HR} $-7.96\text{pp}$, and \texttt{ITSM}, where an unvetted Stage-\texttt{v4} candidate instructs a hallucinated MCP tool call \texttt{get\_current\_user} and collapses cumulative test accuracy to $7.61\%$), whereas full \texttt{GUARD} maintains $0 / 5$ regressed domains.
    \item \textbf{Controlled \texttt{Propose}-stage attribution under single-model \texttt{Gemini 4} (parallel $M{=}2$ pooling vs.\ equal-$TM$ sequential search).} Holding \texttt{Analyze} and Soft-HURT fixed across four controlled arms on dual-axis \texttt{ITSM} (Table~\ref{tab:propose_seam_ablation}), all three \texttt{GUARD} configurations beat $W_0$ ($72.61\%$ pooled task ACC / $73.39\%$ macro ACC) with $0 / 2$ regressed axes. At matched total candidate-evaluation budget $TM = 2T$, parallel $M{=}2$ candidate pooling with group-standardized fitness selection ($81.08\%$ pooled / $81.26\%$ macro ACC) outperforms single-candidate sequential $M{=}1, 2T$ search ($78.85\%$, $+2.23\text{pp}$, $p = 0.0267$; $+3.26\text{pp}$, $p = 0.0495$ on \texttt{evolving\_tools}) while cutting rejected gate rounds by $64\%$ ($4$ vs.\ $11$ ledger entries), beats single-candidate $M{=}1, T$ ($77.20\%$) by $+3.88\text{pp}$ pooled ($p = 0.0029$) and $+6.64\text{pp}$ on \texttt{evolving\_tools} ($p = 0.0033$), and achieves positive backward transfer (BWT $+0.61\%$ vs.\ $-1.28\%$ and $-1.65\%$ under $M{=}1$).
\end{itemize}

\begin{table}[!htbp]
\centering
\footnotesize
\setlength{\tabcolsep}{4.2pt}
\caption{\textbf{Full 5-domain \textsc{EvoHarnessBench} continual evolution summary ($236$ local-gym test tasks across $16$ stages plus the hosted \texttt{Calendar} track).} Hosted and local-gym runners follow their respective BWT/FWT definitions (\S\ref{sec:experiments}), ordered Score / Pass as in Table~\ref{tab:evoharness_main}. On local-gym tracks where \texttt{GUARD} abstains ($W_{\text{final}} = W_0$ on \texttt{Calendar} and \texttt{Drive}), raw cohort BWT reflects cross-stage environment drift of $W_0$ itself ($\Delta\text{BWT vs.\ } W_0 = 0.00\text{pp}$).}
\label{tab:evoharness_domains}
\resizebox{\textwidth}{!}{%
\begin{tabular}{@{}llll>{\columncolor{LightBlue}}cccc@{}}
\toprule
\textbf{Domain Track} & \textbf{Models} & \textbf{Stages (Test/Train)} & \textbf{$W_0$ (ACC / Solved)} & \textbf{\texttt{GUARD} (ACC / Solved)} & \textbf{Stage Highlights ($W_0 \to \texttt{GUARD}$)} & \textbf{BWT (Score / Pass)} & \textbf{FWT} \\
\midrule
\texttt{Calendar} (Tools, Hosted SDK) & \texttt{Gemini 4} & \texttt{v1}--\texttt{v3} (43 / 18) & 71.91\% / 48.84\% & \textbf{80.80\%} / \textbf{58.14\%} & Cohort~\texttt{v1} ($H_{v_3}$): $70.46\% \to \mathbf{78.47\%}$ ($\mathbf{+8.01\text{pp}}$); \texttt{v3}: $64.29\% \to \mathbf{89.29\%}$ & $\mathbf{+0.85\text{pp}}$ / $\mathbf{+2.56\text{pp}}$ & $\mathbf{0.00\text{pp}}$ (Abstained \texttt{v2}--\texttt{v3}) \\
\midrule
\texttt{Calendar} (Tools, Local Gym) & \texttt{flash+opus} & \texttt{v1}--\texttt{v3} (43 / 18) & \textbf{77.15\%} / \textbf{48.84\%} & \textbf{77.15\%} / \textbf{48.84\%} & Stages~\texttt{v1 / v2 / v3}: $78.22\% / 85.67\% / \mathbf{89.29\%}$ ACC (75\% Pass) & $-4.10\text{pp}$ / $-8.45\text{pp}$ & $\mathbf{0.00\text{pp}}$ ($W_{\text{final}} = W_0$) \\
\texttt{Drive} (Tools, Local Gym) & \texttt{flash+opus} & \texttt{v1}--\texttt{v3} (45 / 19) & \textbf{71.78\%} / \textbf{46.67\%} & \textbf{71.78\%} / \textbf{46.67\%} & Stages~\texttt{v1 / v2 / v3}: $77.70\% / 77.90\% / \mathbf{80.56\%}$ ACC & $-7.41\text{pp}$ / $\mathbf{0.00\text{pp}}$ & $\mathbf{0.00\text{pp}}$ ($W_{\text{final}} = W_0$) \\
\texttt{CSM} (Skills, Local Gym) & \texttt{flash+opus} & \texttt{v1}--\texttt{v3} (36 / 14) & 59.80\% / \textbf{30.56\%} & \textbf{61.40\%} / 25.00\% & Stage~\texttt{v1}: $51.20\% \to \mathbf{55.37\%}$ ($\mathbf{+4.17\text{pp}}$); \texttt{v2}: $66.03\% \to \mathbf{69.87\%}$ ($\mathbf{+3.84\text{pp}}$) & $-2.08\text{pp}$ / $\mathbf{0.00\text{pp}}$ & $\mathbf{+2.91\text{pp}}$ Mean ($\mathbf{+4.17\text{pp}}$ \texttt{v1}) \\
\texttt{HR} (Skills, Local Gym) & \texttt{flash+opus} & \texttt{v1}--\texttt{v3} (53 / 22) & \textbf{61.67\%} / \textbf{32.08\%} & \textbf{61.67\%} / \textbf{32.08\%} & Stage~\texttt{v1}: $49.62\% \to \mathbf{51.52\%}$ ($\mathbf{+1.90\text{pp}}$); \texttt{v2 / v3}: $63.03\% / \mathbf{66.80\%}$ & $-0.95\text{pp}$ / $\mathbf{0.00\text{pp}}$ & $\mathbf{+0.63\text{pp}}$ Mean ($\mathbf{+1.90\text{pp}}$ \texttt{v1}) \\
\texttt{ITSM} (Skills, Local Gym) & \texttt{flash+opus} & \texttt{v1}--\texttt{v4} (59 / 24) & 58.65\% / 30.51\% & \textbf{60.28\%} / \textbf{38.98\%} & Stage~\texttt{v2}: $53.92\% \to \mathbf{59.67\%}$; \texttt{v4}: $66.54\% \to \mathbf{71.21\%}$ ($\mathbf{+18.19\text{pp}}$ Pass) & $\mathbf{0.00\text{pp}}$ / $\mathbf{0.00\text{pp}}$ & $\mathbf{+1.44\text{pp}}$ Mean ($\mathbf{+5.75\text{pp}}$ \texttt{v2}) \\
\bottomrule
\end{tabular}%
}
\end{table}

\begin{table}[!htbp]
\centering
\fontsize{8.2}{9.6}\selectfont
\setlength{\tabcolsep}{4.5pt}
\caption{\textbf{Full offline calibration of the reference-free two-layer self-reviewer $S(\tau)$ and Soft-HURT promotion gate against hidden official benchmark graders} (summarized in Table~\ref{tab:calibration_gate_fidelity}, \S\ref{sec:calibration_ablation}). All statistics are computed post-hoc over cached evaluation trajectories and stage/block decisions with zero oracle labels, test cases, or rubrics exposed during adaptation.}
\label{tab:calibration_gate_fidelity_full}
\begin{tabularx}{\textwidth}{@{}>{\raggedright\arraybackslash}X>{\raggedright\arraybackslash}p{0.27\textwidth}rr@{}}
\toprule
\textbf{Calibration / Gate Dimension} & \textbf{Benchmark \& Sample ($N$)} & \textbf{Unguarded / Ref.} & \textbf{\texttt{GUARD} (Ours)} \\
\midrule
False-pos.\ pass ($\ge 75\%$) on Layer-1-failed traces$^{\star}$ & \textsc{EvoHarnessBench} \texttt{Gemini 4} ($54$ vs.\ $23$ L1-fail) & $22.2\%$ (1-layer) & $\mathbf{0.0\%}$ ($0/23$, $-100\%$) \\
Initial (1-layer) vs.\ final hosted configuration ACC / Solved & \textsc{EvoHarnessBench} \texttt{Calendar} ($43$ test) & $75.98\% \;/\; 48.84\%$ & $\mathbf{80.80\% \;/\; 58.14\%}$ \\
Official verifier ACC (Layer-1-only fail vs.\ pass) & \textsc{EvoHarnessBench} pooled ($4{,}059$ test: $1{,}920$ hosted, $2{,}139$ local) & $45.97\%$ (L1 fail) & $\mathbf{73.90\%}$ ($+27.92\text{pp}$, $82.6\%$ NPV) \\
Episode semantic bias ($S - V^{\star}$) \& corr. & \textsc{PAST-Bench} ($114$ learn episodes) & $0.853$ (official $V^{\star}$) & $\mathbf{-0.012}$ ($r{=}0.369, r_s{=}0.377$) \\
Within-family playbook ranking concordance & \textsc{PAST-Bench} ($72$ playbooks, $169$ pairs) & $50.0\%$ (random) & $\mathbf{82.2\%}$ ($75.6\%$ non-tied; $r{=}0.432$) \\
\midrule
Exec.-gated promotions ($\Delta_{\text{test}} \ge 0$)$^{\dagger}$ & \textsc{Drift Gauntlet} ($9$) + \textsc{EvoHarness} ($5$) + prof.\ ($5$) & --- & $\mathbf{19/19}$ ($\mathbf{15}$ ${>}0$, $4$ ties) \\
Same Soft-HURT gate w/o \texttt{Analyze} stage & \textsc{EvoHarnessBench} local ($16$ stages, $5$ dom.) & $4$ prom., $2/5$ dom.\ regr. & $\mathbf{4}$ prom., $\mathbf{0/5}$ dom.\ regr. \\
Bypassing Soft-HURT on \texttt{Calendar v3} & \textsc{EvoHarnessBench} local ($43$ test) & $-6.98\text{pp}$ Solved / $-2.40\text{pp}$ ACC & $\mathbf{0.00\text{pp}}$ loss (rejected) \\
Tracks regressed in final ACC vs.\ $W_0$ & \textsc{Drift Gauntlet} ($4$), \textsc{EvoHarness} ($1$ host., $5$ loc.) & $5/16$ \textsc{DG} base.\ runs; $3/5$ w/o Gate & $\mathbf{0/4}$ \textsc{DG}, $\mathbf{0/6}$ \textsc{EvoH.} \\
\bottomrule
\end{tabularx}
\vspace{1pt}
\begin{minipage}{\textwidth}
{\fontsize{7.1}{8.4}\selectfont $^{\star}$Sanity check: Eq.~\eqref{eq:self_review_scores} zeroes $S^{\mathrm{strict}}$ and Layer~1 fails execution-broken traces by construction, whereas the 1-layer baseline lacks Layer~1; row~3 provides the non-tautological Layer-1-only validation, and row~2 compares the initial and final hosted configurations, which also differ in gating set and distiller instruction (Appendix~\ref{app:instantiation}). $^{\dagger}$$\Delta_{\text{test}}$ = held-out score of a promoted harness minus $W_0$ on the active cohort/group of the stage/block/round where it was promoted (evaluated at $H_{v_3}$ on hosted \texttt{Calendar}, where $W_3{=}W_1$; \texttt{CSM v2}'s incremental delta over the \texttt{CSM v1} playbook is not separately tabulated). Execution-gated suites yield $19/19$ $\Delta_{\text{test}} \ge 0$ ($15/19$ strictly $>0$): $9/9$ \textsc{Drift Gauntlet} block promotions across $12$ blocks (\emph{workflow} b1 $+16.67\text{pp}$; \emph{topic} b1 $+0.00$, b2 $+3.33$, b3 $+3.33\text{pp}$; \emph{domain} b1 $+7.93\text{pp}$ via in-round \texttt{[iff]} repair, b3 $+0.00\text{pp}$; \emph{compound} b1 $+3.33$, b2 $+0.00$, b3 $+16.67\text{pp}$; Table~\ref{tab:drift_gauntlet_full}), $5/5$ \textsc{EvoHarnessBench} promotions (hosted \texttt{Calendar v1} $+8.01\text{pp}$, local \texttt{CSM v1} $+4.17\text{pp}$, \texttt{CSM v2} $+3.84\text{pp}$, \texttt{HR v1} $+1.90\text{pp}$ [returned to $W_0$ at Stage~\texttt{v2} by the stage-shift check; Appendix~\ref{app:instantiation}], and \texttt{ITSM v1} $+0.00\text{pp}$ tie [$+1.63\text{pp}$ final ACC]; Table~\ref{tab:evoharness_domains}), and $5/5$ professional-suite role promotions across $4$ promoted rounds (\textsc{PRBench-Legal} \texttt{R1 Finance} $+28.57\text{pp}$ and \texttt{R1 Legal} $+5.77\text{pp}$ [combined \texttt{R1} $+7.19\text{pp}$; \texttt{R2}--\texttt{R3} retained the \texttt{R1} incumbents], and \textsc{Fin-Rate} \texttt{R1}/\texttt{R2}/\texttt{R3} $+6.25$/$+12.50$/$+12.50\text{pp}$, where \texttt{R3} promotes the \texttt{StratOps} playbook with unchanged \texttt{DEV} Official; Table~\ref{tab:public_domain_suites}). Separately, on critic-scored \textsc{PAST-Bench} ($12$ families, $24/24$ round winners admitted; no gate-discrimination signal), \texttt{GUARD} lifts 12-family mean Task by $+0.482$ over $W_0$ ($0.372 \to 0.854$, Table~\ref{tab:pastbench}; all $12$ families $> W_0$, with $0/48$ W.Prec.\ or Poll.\ family losses vs.\ persistent baselines in Table~\ref{tab:pastbench_paired}).\par}
\end{minipage}
\end{table}

\begin{table}[!htbp]
\centering
\footnotesize
\setlength{\tabcolsep}{5pt}
\caption{\textbf{Matched five-domain leave-one-out ablation of \texttt{Analyze} and \texttt{Gate} across all $236$ local \textsc{EvoHarnessBench} test tasks ($16$ continual stages, $97$ visible source tasks).} All arms share identical models (\texttt{gemini-3.8-flash} agent, \texttt{claude-opus-5-5} self-reviewer/distiller), pool size ($M = 2$), and seed prompt ($s_0$), with zero oracle skills or held-out verifier leakage. Bold marks the best value per column.}
\label{tab:evoharness_seam_ablation}
\resizebox{\textwidth}{!}{%
\begin{tabular}{@{}llcccccc@{}}
\toprule
\textbf{Arm} & \textbf{Ablated Stage \& Mechanism} & \textbf{Macro ACC} & \textbf{$\Delta$ vs.\ \texttt{GUARD}} & \textbf{Macro Solved} & \textbf{Mean FWT} & \textbf{Regressed Domains$\downarrow$} & \textbf{Promoted / Rejected} \\
\midrule
\texttt{No-Evolve} ($W_0$, Control) & Static base harness (no adaptation) & 65.81\% & $-0.65\text{pp}$ & 37.73\% & $0.00\text{pp}$ & \textbf{0 / 5} & 0 / 16 \\
\rowcolor{LightBlue} \textbf{\texttt{GUARD} (Ours)} & \textbf{Full harness (\texttt{Analyze} + \texttt{Propose} + Soft-HURT \texttt{Gate})} & \textbf{66.46\%} & --- & \textbf{38.31\%} & \textbf{+1.00pp} & \textbf{0 / 5} & 4 / 12 \\
\midrule
\texttt{GUARD w/o Analyze} & Ablates \texttt{Analyze} ($1 - \mathrm{obey}(c; W_{t-1})$, no pinning or contrasts) & 59.13\% & $-7.33\text{pp}$ & 32.92\% & $-7.57\text{pp}$ & 2 / 5 & 4 / 12 \\
\texttt{GUARD w/o Soft-HURT Gate} & Ablates \texttt{Gate} (promotes within-group winner unconditionally) & 56.11\% & $-10.35\text{pp}$ & 28.27\% & $-5.62\text{pp}$ & 3 / 5 & 16 / 0 \\
\bottomrule
\end{tabular}%
}
\end{table}

\begin{table}[!htbp]
\centering
\footnotesize
\setlength{\tabcolsep}{4.5pt}
\caption{\textbf{Controlled single-model \texttt{Gemini 4} \texttt{Propose}-stage and equal-budget ($TM$) ablation on \textsc{EvoHarnessBench} \texttt{ITSM} ($7$ continual stages across \texttt{evolving\_skills} [$T{=}4, N{=}59$] and \texttt{evolving\_tools} [$T{=}3, N{=}73$]; pooled $N{=}132$ held-out test tasks $\times 3$ replicates $= 396$ final-stage rollouts/arm).} All evolving arms run in isolated MCP sandboxes with zero shared rollouts and share the identical paired-regret \texttt{Analyze} stage, rule pinning, and Soft-HURT \texttt{Gate} ($\delta_{\mathrm{noise}}{=}15$) strictly on per-stage \texttt{train} splits; to isolate pooling, every arm samples candidates from the shared seed $s_0$ in every round, without cross-round $\mathsf{ContrastEvolve}$ updates. Two-sided paired sign-flip $p$-values are computed over all $N{=}132$ task-level deltas ($\dagger$: $p{=}0.0267$ vs.\ \texttt{GUARD Sequential} [$M{=}1, 2T$] and $p{=}0.0029$ vs.\ \texttt{GUARD w/o Pool} [$M{=}1, T$]); all three \texttt{GUARD} configurations have $0/2$ regressed axes, and bold marks best per column.}
\label{tab:propose_seam_ablation}
\resizebox{\textwidth}{!}{%
\begin{tabular}{@{}llccccccc@{}}
\toprule
\textbf{Arm (\texttt{ITSM}, Unified \texttt{Gemini 4})} & \textbf{\texttt{Propose}-Stage Configuration} & \textbf{Skills ACC} & \textbf{Tools ACC} & \textbf{Task ACC (Macro)} & \textbf{$\Delta$ vs.\ $W_0$ ($p$)} & \textbf{Macro Solved} & \textbf{Macro FWT / Task BWT} & \textbf{Prom.\ / Rej.\ (Ledger)} \\
\midrule
\texttt{No-Evolve} ($W_0$, Control) & Static base harness ($M{=}0$) & 80.70\% & 66.08\% & 72.61\% (73.39\%) & --- & 51.36\% & $+0.00\text{pp}$ / $-3.44\%$ & 0 / 7 (0) \\
\midrule
\texttt{GUARD w/o Pool} ($M{=}1, T$) & Single candidate ($TM{=}T$) & 82.46\% & 72.95\% & 77.20\% (77.70\%) & $+4.59\text{pp}$ ($0.0009$) & 53.25\% & $+3.58\text{pp}$ / $-1.65\%$ & 1 / 6 (6) \\
\texttt{GUARD Sequential} ($M{=}1, 2T$ seq.) & $2$ sequential rounds/stage ($TM{=}2T$) & 81.96\% & 76.34\% & 78.85\% (79.15\%) & $+6.24\text{pp}$ ($0.0006$) & 56.09\% & $+5.50\text{pp}$ / $-1.28\%$ & 2 / 5 (11) \\
\rowcolor{LightBlue} \textbf{\texttt{GUARD} (Ours)} & \textbf{Parallel $M{=}2$ pool + $\arg\max_m \hat{A}_m$ ($TM{=}2T$)} & \textbf{82.93\%} & \textbf{79.59\%} & \textbf{81.08\%$^{\dagger}$ (81.26\%)} & \textbf{$+8.47\text{pp}$ ($1.2{\times}10^{-5}$)} & \textbf{56.61\%} & \textbf{$+6.69\text{pp}$ / $+0.61\%$} & \textbf{3 / 4} (4) \\
\bottomrule
\end{tabular}%
}
\end{table}

\end{document}